\pdfoutput=1

\documentclass[11pt]{article}

\usepackage[hyperref]{EACL2023}
\usepackage{graphicx, caption, subcaption, tabularx, ragged2e,booktabs}
\usepackage{xcolor}
\usepackage{natbib}
\usepackage{amsmath}

\newcommand{\snli}{\textsc{snli}}
\newcommand{\anli}{\textsc{anli}}
\newcommand{\wanli}{\textsc{wanli}}
\newcommand{\mnli}{\textsc{mnli}}
\newcommand{\mcme}{\textsc{mcme}}
\newcommand{\bald}{\textsc{bald}}
\newcommand{\dal}{\textsc{dal}}
\newcommand{\random}{\textsc{random}}

\usepackage{times}
\usepackage{latexsym}
\usepackage{algorithm}
\usepackage{algpseudocode}

\usepackage[T1]{fontenc}

\usepackage[utf8]{inputenc}

\usepackage{microtype}

\usepackage{inconsolata}

\title{Investigating Multi-source Active Learning for Natural Language Inference}

\author{
Ard Snijders$^1$\quad Douwe Kiela$^{2}$ \quad Katerina Margatina$^3$\\
$^1$University of Amsterdam\quad$^2$Stanford University\quad$^3$University of Sheffield\\
\texttt{ardsnijders@gmail.com}\\
}
\begin{document}
\maketitle
\begin{abstract}
In recent years, active learning has been successfully applied to an array of NLP tasks.
However, prior work often assumes that training and test data are drawn from the same distribution. 
This is problematic, as in real-life settings data may stem from several sources of varying relevance and quality.
We show that four popular active learning schemes fail to outperform random selection when applied to unlabelled pools comprised of \textit{multiple} data sources on the task of natural language inference. 
We reveal that uncertainty-based strategies perform poorly due to the acquisition of collective outliers, i.e., hard-to-learn instances that hamper learning and generalization. 
When outliers are removed, strategies are found to recover and outperform random baselines.
In further analysis, we find that collective outliers vary in form between sources, and show that hard-to-learn data is not always categorically harmful. 
Lastly, we leverage dataset cartography to introduce difficulty-stratified testing and find that different strategies are affected differently by example learnability and difficulty. 
\end{abstract}

\section{Introduction}


In recent years, active learning (AL) \cite{Cohn_1996} has emerged as a promising avenue for data-efficient supervised learning \cite{Zhang2022}.
AL has been successfully applied to a variety of NLP tasks, such as text classification~\citep{Zhang2016,
Siddhant2018, 
Prabhu, 
Ein-Dor, 
Margatina_tapt}, entity recognition~\citep{Shen2017, Siddhant2018, Lowell},
part-of-speech tagging~\cite{chaudhary-etal-2021-reducing} 
and neural machine translation~\citep{Peris2018, Liu2018, Zhao2020_nmt}. 

However, these works share a major limitation: they often implicitly assume that unlabelled training data comes from a single \textit{source}\footnote{Throughout the paper, we use the term ``\textit{source}'' to describe the varying domains of the textual data that we use in our experiments. More broadly, ``different sources'' refers to having data drawn from different distributions.} \citep{Houlsby2011, Sener, Huang2016, 
Gissin2019,
Margatina_cal}. We refer to this setting as \textit{single-source AL}.
\begin{figure}[t]
    \centering
    \includegraphics[width=0.9\columnwidth]{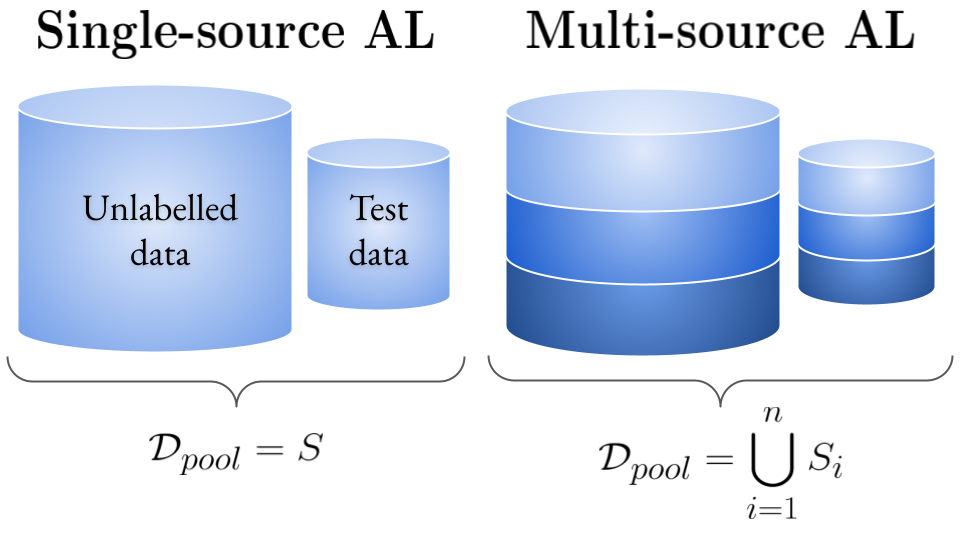}
    \caption{The pool of unlabelled data consists of a source $S$, or multiple sources $\bigcup_iS_i$, for the \textit{single-source} and \textit{multi-source} AL setting, respectively. The in-domain test set follows the same distribution of sources.}
    \label{fig:intro}
\end{figure}
The single-source assumption is problematic for various reasons \cite{Kirsch2021}. In real-life settings we have no guarantees that all unlabelled data at our disposal will necessarily stem from the same distribution, nor will we have assurances that all examples are of consistent quality, or that they bear sufficient relevancy to our task. For instance, quality issues may arise when unlabelled data is collected through noisy processes with limited room for monitoring individual samples, such as web-crawling~\cite{kreutzer-etal-2022-quality}. Alternatively, one may have access to \textit{several} sources of unlabelled data of decent quality, but incomplete knowledge of their relevance to the task-at-hand. For instance, medical data may be collected from various different physical sites (hospitals, clinics, general practitioners) which may differ statistically from the target distribution due to e.g., differences in patient demographics. Ideally, AL methods should be robust towards these conditions in order to achieve adequate solutions.



In this work, we study whether existing AL methods can adequately select relevant data points in a \textit{multi-source scenario} for NLP. We examine robustness by evaluating AL performance in both in-domain (ID) and out-of-domain (OOD) settings, while we also conduct an extensive analysis to interpret our findings. A primary phenomenon of interest here concerns \textit{collective outliers}: data points which models struggle to learn due to high ambiguity, the requirement of specialist skills or labelling errors~\cite{han2000dm}. 
While such outliers were previously found to disrupt several AL methods for visual question answering~\cite{Karamcheti2021}, their impact on text-based tasks remains under-explored. Given the wide body of work on the noise and biases that pervade the task of natural language inference (NLI)~\citep{Bowman, Williams2017, Gururangan2018, Poliak2018, Tsuchiya2019, Geva2019, Liu2022}, we may reasonably assume popular NLI datasets to suffer from collective outliers as well. 

Concretely, our contributions can be summarised as follows: \textbf{(1)} We apply several popular AL methods in the under-explored multi-source, pool-based setting on the task of NLI, using RoBERTa-large~\citep{Liu2020} as our acquisition model, and find that no strategy consistently outperforms random selection (\S\ref{section:main_results}). 
\textbf{(2)} We seek to explain our findings by creating datamaps to explore the actively acquired data (\S\ref{section:cartography_method}) and show that uncertainty-based acquisition functions perform poorly due to the acquisition of collective outliers (\S\ref{par:strat_maps}).
\textbf{(3)} We examine the effect of training data difficulty on downstream performance (\S\ref{section:training_difficulty}) and after thorough experiments we find that uncertainty-based AL methods 
recover and even surpass random selection when hard-to-learn data points are removed from the pool (\S\ref{section:ablation}).
\textbf{(4)} Finally, we introduce difficulty-stratified testing and show that the \textit{learnability} of acquired training data affects different strategies differently  at test-time (\S\ref{section:stratified_testing}).
Our code is publicly available at \url{https://github.com/asnijders/multi_source_AL}.

\section{Related Work}
\paragraph{Multi-source AL for NLP}
While AL has been studied for a variety of tasks in NLP~\cite{Siddhant2018, Lowell,Ein-Dor, Shelmanov, Margatina_cal,yuan-etal-2022-adapting, schroder-etal-2022-revisiting, Margatina_tapt, kirk-etal-2022-data,Zhang2022}, the majority of work remains limited to settings where training data is assumed to stem from a single source. 
Some recent works have sought to address the issues that arise when relaxing the single-source assumption~\citep{Ghorbani, Kirsch2021, Kirsch2021a}, though results remain primarily limited to image classification. Moreover, these works study how AL fares under the presence of \textit{corrupted} training data, such as duplicating images or adding Gaussian noise, and they do not consider settings where sampling from multiple sources may be \textit{beneficial} due to complementary source attributes. \citet{He2021} examine a multi-domain AL setting, but they focus on leveraging common knowledge between domains to learn a \textit{set} of models for a set of domains, which contrasts with our single-model pool-based setup. Closest to our work, \citet{Longpre} explore pool-based AL over multiple domains 
and find that some strategies consistently outperform random on question answering and sentiment analysis. However, the authors crucially omit a series of measurements, as they instead perform \textit{a single AL iteration}, limiting the effectiveness of AL, while complicating comparison with our results.

\paragraph{Dataset Cartography}
\citet{Karamcheti2021} employ dataset cartography \cite{Swayamdipta2020} and show that a series of AL algorithms fail to outperform random selection in visual question answering due to the presence of collective outliers. \citet{Zhang} apply datamaps to AL and introduce the cartography active learning strategy, identifying that examples with poor learnability often suffer from label errors. Our work contrasts with both of these works in that we show that hard-to-learn data is not always unequivocally harmful to learning. Moreover, both works only examine learnability of \textit{training} examples, whereas we also consider how learnability of acquired data affects model performance at test-time.

\section{Single \& Multi-source Active Learning}\label{sec:al_settings}

We assume a warm-start, pool-based AL scenario~\cite{Settles} with access to a pool of \textit{unlabelled} training data,  $\mathcal{D}_{pool}$, and a seed dataset of labelled examples, $\mathcal{D}_{train}$. During each iteration $i$ of AL, we first train a model $\mathcal{M}$ with $\mathcal{D}_{train}$ and then use it 
in conjunction with some acquisition function $\mathcal{A}$ to select a new batch of unlabelled examples $\mathcal{D}_{batch}$ from $\mathcal{D}_{pool}$ for labelling. Upon labelling, these examples are removed from $\mathcal{D}_{pool}$ and added to $\mathcal{D}_{train}$, after which a new round of AL begins.
\begin{gather*}
    \mathcal{D}_{train}^{i+1} = \mathcal{D}_{train}^{i} \cup \mathcal{D}_{batch}^{i+1}\\
    \mathcal{D}_{pool}^{i+1} = \mathcal{D}_{pool}^{i} \setminus \mathcal{D}_{batch}^{i+1}
\end{gather*}

\paragraph{Single-source AL}\label{sec:single_source} For single-source AL, we assume that $\mathcal{D}_{pool}$ is a set of unlabelled data that stems from a single source $S$, i.e.\ $\mathcal{D}_{pool} = S$.
\paragraph{Multi-source AL} For multi-source AL, we assume that $\mathcal{D}_{pool}$ comprises a union of distinct sources $S_1, S_2, ..., S_n$ such that $\mathcal{D}_{pool} = \bigcup\limits_{i=1}^{n} S_i$.


\subsection{Data Acquisition}
An acquisition function $\mathcal{A}$ is responsible for selecting the most \textit{informative} unlabelled data from the pool, aiming to improve over random sampling.
We use a set of acquisition functions which we deem representative for the wider AL toolkit: Monte Carlo Dropout Max-Entropy, \citep[\textbf{\mcme{}};][]{Gal2017} 
is an uncertainty-based acquisition strategy where we take the mean label distribution over $T$ Monte-Carlo dropout \citep{Gal2015} network samples and select the $k$ data points with the highest predictive entropy. Bayesian Active Learning by Disagreement, \citep[\textbf{\bald{}};][]{Houlsby2011} is an uncertainty-based acquisition strategy which employs Bayesian uncertainty to identify data points for which many models disagree about.
Discriminative Active Learning  \citep[\textbf{\dal{}};][]{Gissin2019} is a diversity-based acquisition function designed to acquire a training set that is indistinguishable from the unlabelled set. 




\subsection{Analysis of Acquired Data}
At each data acquisition step, we seek to examine what kind of data each acquisition function has selected for annotation.
Following standard practice in active learning literature \citep{Zhdanov2019,Yuan2020, Ein-Dor, Margatina_cal} we profile datasets acquired by strategies via acquisition metrics. Concretely, we consider the \textit{input diversity} and \textit{output uncertainty} metrics. We provide more details in Appendix~\ref{sec:app_analysis_metrics}.
\paragraph{Input Diversity}
To evaluate the diversity of acquired sets in the input space, we follow \citet{Yuan2020} and measure input diversity as the Jaccard similarity 
between the set of tokens from the acquired training set $\mathcal{D}_{train}$ and the set of tokens from the \textit{remainder} of the unlabelled pool $\mathcal{D}_{pool}$.
This function assigns \textit{high} diversity to strategies acquiring samples with \textit{high} token overlap with the unlabelled pool, and vice versa.
\paragraph{Output Uncertainty}
To approximate the output uncertainty of an acquired training set $\mathcal{D}_{train}$ for a given strategy, we follow \citet{Yuan2020} and use a model trained on the \textit{entire dataset} to compute predictive entropy of all the examples in the dataset that we want to examine. The model is trained on all training data as this grants more accurate uncertainty measurements.

\section{Experimental Setup}\label{sec:datasets}
\paragraph{Data}
We perform experiments on Natural Language Inference (NLI), a popular
classification task to gauge a model's natural language understanding~\citep{Bowman}. We construct the unlabelled pool from three distinct datasets: \textbf{\snli{}} \citep{Bowman}, \textbf{\anli{}} \citep{Nie2019} and \textbf{\wanli{}} \citep{Liu2022}.
We consider 
\textbf{\mnli{}}
\citep{Williams2017} 
as an out-of-domain set to evaluate the transferability of actively acquired training sets. For more details see Appendix~\ref{sec:app_datasets}.

\paragraph{Experiments}
We apply AL with two distinct end-goals: \textbf{in-domain} (ID) generalization, where the same source(s) are used for both the unlabelled pool $\mathcal{D}_{pool}$ and the test set $\mathcal{D}_{test}$, and \textbf{out-of-domain} (OOD) generalization, where we evaluate on the test set of an external source of which no training data was present in the unlabelled pool.
\paragraph{Constructing multi-source pools}\label{section:multi_source_desc}
For multi-source AL, 
the unlabelled pool comprises the \textit{union} of the \snli{}, \anli{} and \wanli{} training sets. During model selection, we similarly assume the union of \snli{}, \anli{} and \wanli{} validation sets. For both training and validation data we down-sample sources to the minority source size to obtain pools with even shares per source. We sub-sample \textit{training data} for each source to reduce experiment runtimes. This yields a pool of $60$K unlabelled training examples comprised of $20$K shares of each source. 

\paragraph{AL Parameters}
We assume an initial seed training set 
of size $|\mathcal{D}_{train}^{i=0}|=500$ 
and an acquisition size $k = 500$, such that $k$
examples are acquired per round of AL.
We perform $7$ rounds of AL: the final actively acquired labelled dataset $\mathcal{D}_{train}^{i=7}$ comprises $4$K examples.
We run each experiment $5$ times with different random seeds to account for stochasticity in parameter initializations. 
\begin{figure*}[!t]

\minipage{0.30\textwidth}
\includegraphics[width=\linewidth]{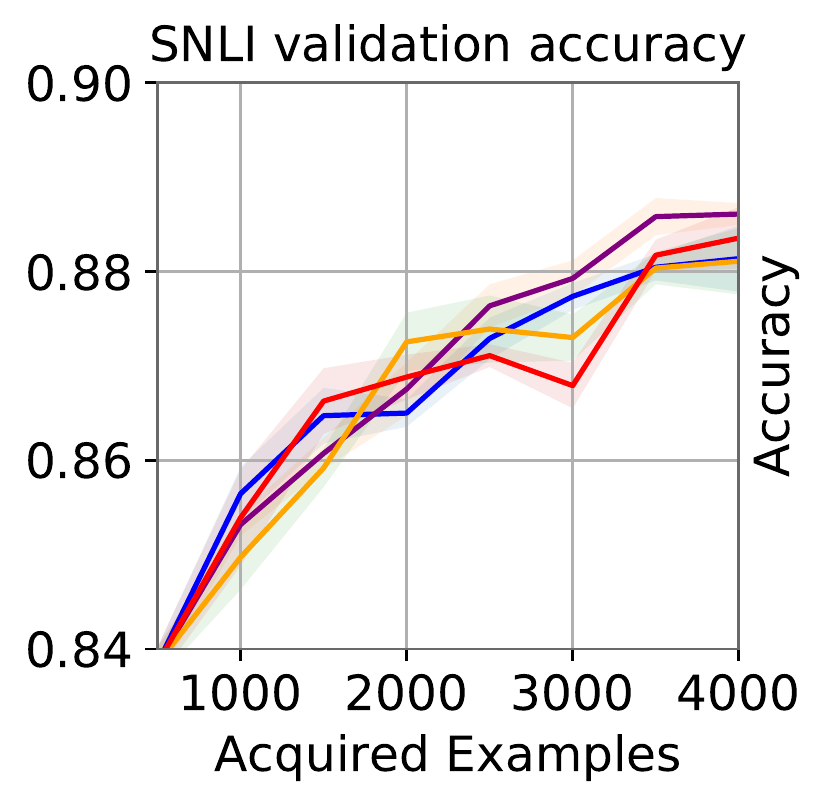}
\endminipage
\minipage{0.30\textwidth}
\includegraphics[width=\linewidth]{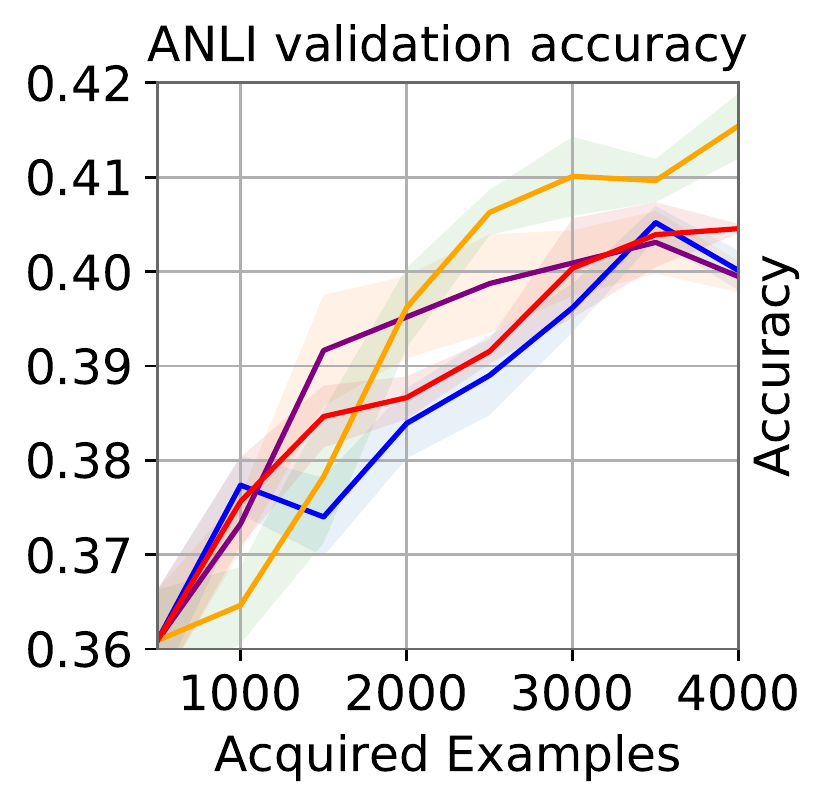}
\endminipage
\centering
\minipage{0.30\textwidth}
\includegraphics[width=\linewidth]{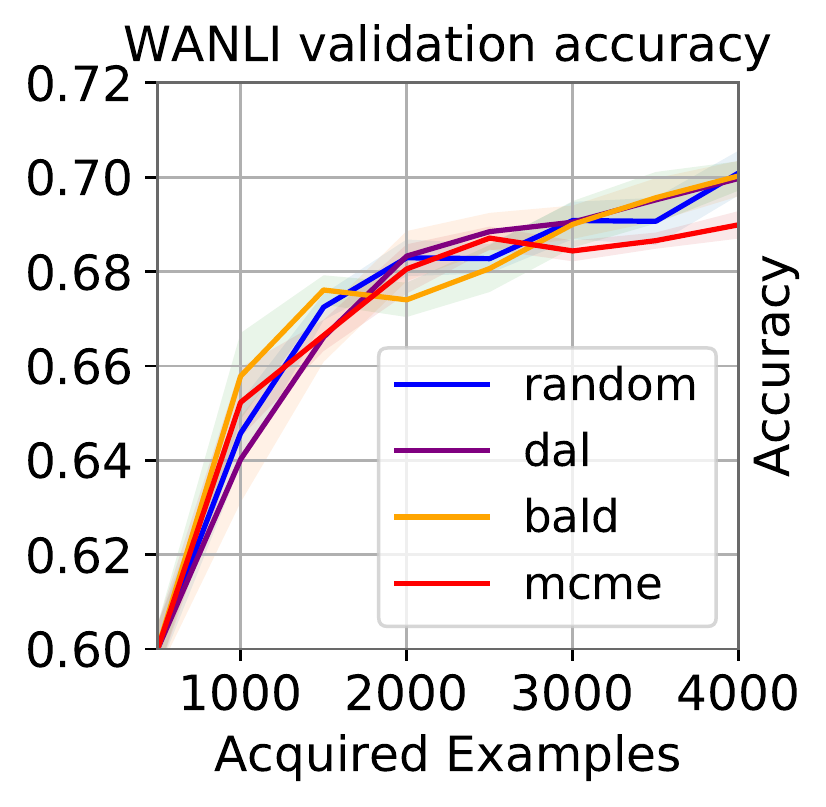}
\endminipage
\caption{AL \textbf{single-source} in-domain learning curves. 
}
\label{fig:baseline_learning_curves}
\end{figure*}
\section{Results}\label{section:main_results}
\subsection{AL over single sources}\label{section:ss_results}
We first provide single-source AL
results in Figure~\ref{fig:baseline_learning_curves}. AL has mixed performance across sources; it fails to consistently outperform random selection on \snli{} and \wanli{}, though most strategies tend to outperform random on \anli{}. However, none of the strategies consistently outperform others across sources. Moreover, the extent of improvement over the acquisition phase varies between tasks; e.g., for \snli{}, strategy curves are close and the accuracy improvement from $500$ to $4$K examples is small. This is also reflected in test outcomes~(Table~\ref{table:single_source_test}), with models performing similarly across sources.

Observing Table \ref{table:single_source_basic_metrics},\footnote{See Table~\ref{table:single_source_basic_metrics_full} for standard deviations in the Appendix.} we find that uncertainty-based methods (\bald{}, \mcme{}) tend to acquire the most uncertain data points, as expected. Still, differences between strategies are small for \wanli{} and \anli{}. Conversely, \random{} and \dal{} tend to acquire data with greater diversity in the input space, but again variance across methods is low. Here, observed homogeneity in the input space may explain why test outcomes lie closely together: if the acquired training sets per strategy are very similar, we should expect to see similarly small differences in test outcomes for models trained on them. Relating these outcomes to the per-source learning curves in Figure \ref{fig:baseline_learning_curves}, however, we observe no clear relation between metric results and performance outcomes: \textit{acquiring uncertain or diverse data does not seem to be predictive of AL success throughout the acquisition process}. 

\begin{table}[t]
\centering
\resizebox{\columnwidth}{!}{%
\begin{tabular}{|c|c|c|c|}
\hline
      & \snli{} & \anli{} & \wanli{} \\ \hline \hline
\textbf{\random{}}  & $ 87.09 \pm 0.30 $  
& $ 39.90 \pm 1.02 $ 
& $ \textbf{70.24} \pm 0.70 $ \\\hline
\textbf{\dal{}}  & $ 87.49 \pm 0.20 $  & $ 39.84 \pm 1.15 $  & $ 69.80 \pm 1.00 $ \\\hline
\textbf{\bald{}}  & $ 87.37 \pm 0.40 $ 
& $ \textbf{40.16} \pm 0.69 $
& $ 69.54 \pm 0.70 $ \\\hline
\textbf{\mcme{}} & $ \textbf{87.58} \pm 0.80 $ &  $ 40.15 \pm 0.60 $ & $ 69.07 \pm 1.20 $ \\ \hline
\end{tabular}
}
\caption{\textbf{Single-source AL} test results. At test-time, the labeled set $\mathcal{D}_{train}$ comprises $4$K examples.}
\label{table:single_source_test}
\end{table}
\begin{table}[t!]
\centering
\resizebox{\columnwidth}{!}{
\begin{tabular}{|c|ll|c|c||c|c|c|}
\hline
\multicolumn{1}{|l|}{\textbf{Task}} & \multicolumn{2}{l|}{\textbf{Strategy}}& \textbf{I-Div.} & \textbf{Unc.} & \textbf{N} & \textbf{E} & \textbf{C} \\ 
\hline \hline

                      & \multicolumn{2}{l|}{\random{}}  & $0.259$     & $0.051$  & 0.34  & 0.33  & 0.33 \\ 
    \textbf{\snli{}}      & \multicolumn{2}{l|}{\dal{}}    & $\textbf{0.267}$     & $0.067$  & 0.38  & 0.29  & 0.33  \\ 
                      & \multicolumn{2}{l|}{\bald{}}    & $0.249$     & $0.072$  & 0.32   & 0.31   & 0.37  \\ 
                      & \multicolumn{2}{l|}{\mcme{}}    & $0.261$    & $\textbf{0.093}$  & 0.32   & 0.34   & 0.34  \\ \hline \hline

                      & \multicolumn{2}{l|}{\random{}}  & $\textbf{0.324}$     & $0.208$  & 0.47  & 0.38  & 0.15 \\ 
    \textbf{\wanli{}}     & \multicolumn{2}{l|}{\dal{}}    & $0.321$     & $0.223$  & 0.49  & 0.38  & 0.13  \\ 
                      & \multicolumn{2}{l|}{\bald{}}    & $0.317$     & $0.213$  & 0.40   & 0.39 & 0.21  \\ 
                      & \multicolumn{2}{l|}{\mcme{}}    & $0.318$    & $\textbf{0.229}$  & 0.38   & 0.35  & 0.27  \\ \hline \hline

                      & \multicolumn{2}{l|}{\random{}}  & $\textbf{0.492}$     & $0.082 $  & 0.43  & 0.32  & 0.26 \\ 
    \textbf{\anli{}}     & \multicolumn{2}{l|}{\dal{}}    & $0.467 $     & $0.084$  & 0.47  & 0.3  & 0.23  \\ 
                      & \multicolumn{2}{l|}{\bald{}}    & $0.478 $     & $0.094 $  & 0.39   & 0.30   & 0.31  \\ 
                      & \multicolumn{2}{l|}{\mcme{}}    & $0.491 $    & $\textbf{0.110} $  & 0.37   & 0.31   & 0.31  \\ \hline \hline

                          & \multicolumn{2}{l|}{\random{}}  & $0.276 $              & $0.405 $  & 0.41  & 0.34  & 0.25 \\ 
    \textbf{Multi}    & \multicolumn{2}{l|}{\dal{}}     & $0.276$              & $0.463 $  & 0.47  & 0.34  & 0.19  \\ 
                      & \multicolumn{2}{l|}{\bald{}}    & $0.220 $              & $0.445 $  & 0.34   & 0.32   & 0.35  \\ 
                      & \multicolumn{2}{l|}{\mcme{}}    & $\textbf{0.323} $     & $\textbf{0.556} $  & 0.40   & 0.30   & 0.30  \\ \hline
\end{tabular}
}
\caption{Profiling strategy acquisitions for \textbf{single-source} and \textbf{multi-source} AL in terms of input diversity (\textbf{I-Div.}),  uncertainty (\textbf{Unc.}) and class distributions (\textbf{N}: \textit{neutral}, \textbf{E}: \textit{entailment}, \textbf{C}: \textit{contradiction}) across seeds. Each strategy is profiled \textit{after} all rounds of AL.}
\label{table:single_source_basic_metrics}
\end{table}

\begin{figure}[t!]
\centering
\minipage{0.5\columnwidth}
\includegraphics[width=\linewidth]{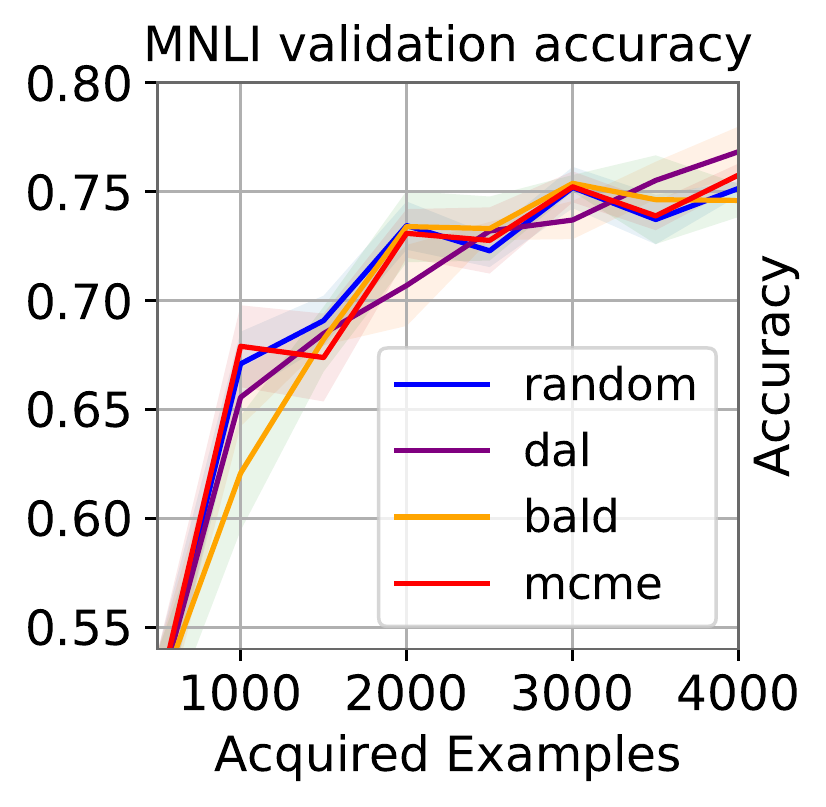}
\endminipage
\minipage{0.5\columnwidth}
\includegraphics[width=\linewidth]{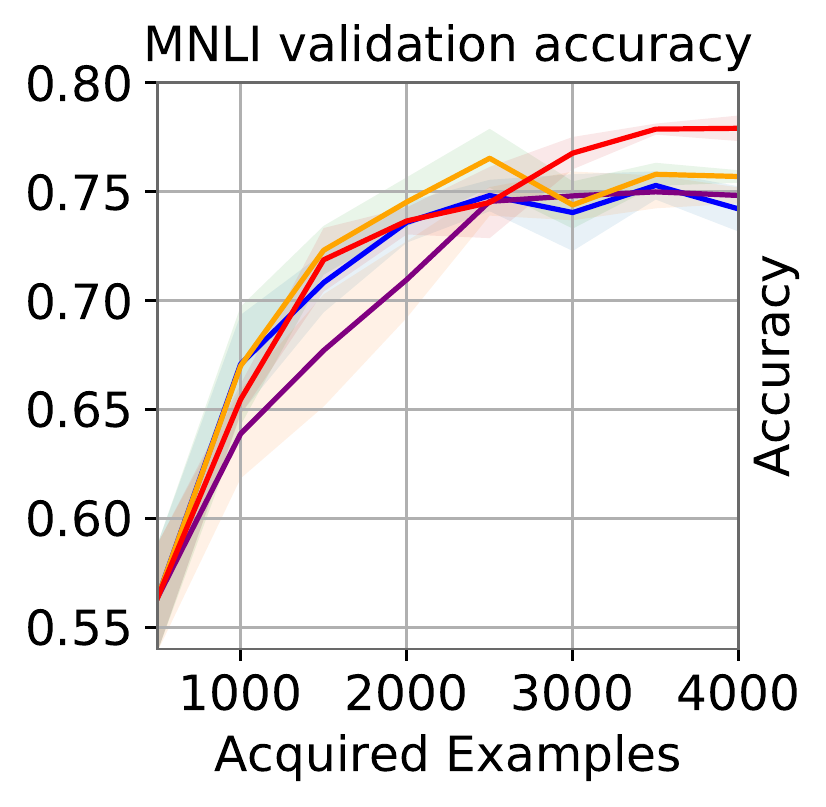}
\endminipage
\vfill
\minipage{0.5\columnwidth}
\includegraphics[width=\linewidth]{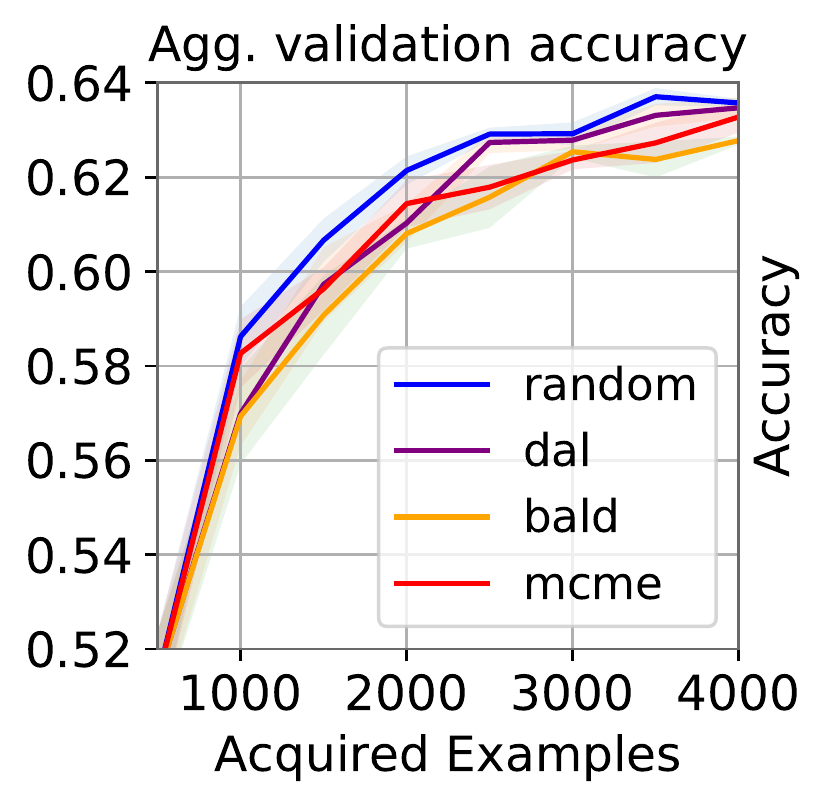}
\subcaption{Original training data}
\endminipage
\minipage{0.5\columnwidth}
\includegraphics[width=\linewidth]{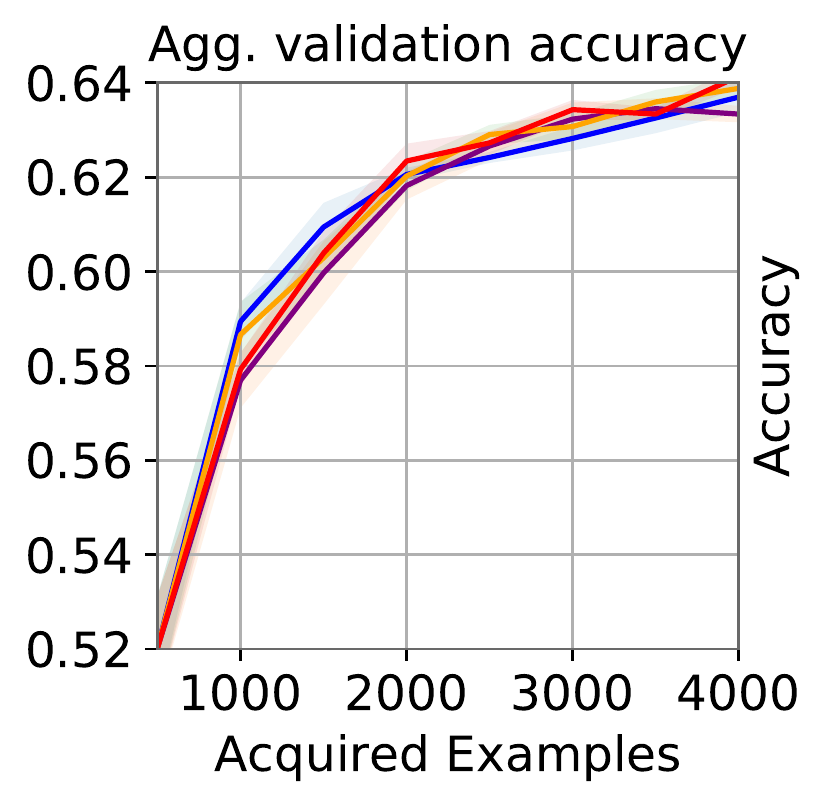}
\subcaption{Hard-to-learn removed}
\endminipage
\caption{\textbf{Multi-source} AL OOD (\mnli{}) and ID (\snli{}, \anli{}, \wanli{}) learning curves. On the left, the unlabelled pool has been left untreated; on the right, hard-to-learn data was removed from the pool.
}
\label{fig:ablating_AL_agg_OD}
\end{figure}
\begin{table*}[t]
\centering
\resizebox{\textwidth}{!}{\begin{tabular}{|c|c|c|c|c|}
\hline
                    & \textbf{\snli{}}                 & \textbf{\anli{}}                 & \textbf{\wanli{}}                & \textbf{\mnli{}}         \\ \hline \hline
\textbf{\random{}}  & $85.9 \pm 0.9 \mid 85.3 \pm 0.3$ & $37.8 \pm 1.2 \mid 37.9 \pm 0.8$ & $67.7 \pm 1.6 \mid 67.2 \pm 0.9$ & $74.2 \pm 2.3 \mid 74.4 \pm 1.5$ \\ \hline
\textbf{\mcme{}}    & $85.9 \pm 0.7 \mid 84.3 \pm 0.6$ & $\underline{\textbf{38.6}} \pm 1.0 \mid \underline{38.4} \pm 0.9$ & $67.4 \pm 0.4 \mid 66.3 \pm 0.7$ & $\underline{\textbf{77.8}} \pm 1.3 \mid 75.6 \pm 2.2$ \\ \hline
\textbf{\bald{}}    & $\underline{\textbf{86.3}} \pm 1.0 \mid \underline{85.8} \pm 0.2$ & $36.9 \pm 1.3 \mid 36.1 \pm 0.5$ & $\underline{\textbf{68.4}} \pm 1.1 \mid 65.9 \pm 1.1$ & $75.8 \pm 0.9 \mid 74.3 \pm 1.6$ \\ \hline
\textbf{\dal{}}     & $84.3 \pm 1.2 \mid 85.3 \pm 0.9$ & $37.8 \pm 1.2 \mid 37.3 \pm 1.3$ & $68.0 \pm 0.6 \mid \underline{68.1} \pm 1.0$ & $74.6 \pm 1.2 \mid \underline{76.6} \pm 2.6$ \\ \hline
\end{tabular}}
\caption{\textbf{Multi-source} AL test outcomes with and without hard-to-learn instances removed. Cell scheme to be read as \textit{ablated} $\mid$ \textit{original}. \textbf{Bold} denotes best score for each test set overall and \underline{underline} best score per setting. At test-time, the labeled set $\mathcal{D}_{train}$ comprises $4$K examples.
}
\label{fig:ablating_AL_test}
\end{table*}

\subsection{AL over multiple sources}\label{section:ms_results}
Next, we provide the results of our multi-source setting, as described in \S\ref{sec:al_settings}, and observe that again AL fails to consistently outperform random sampling on the aggregate ID test set of \snli{}, \anli{} and \wanli{} (Figure \ref{fig:ablating_AL_agg_OD}, bottom left). We also evaluate on the OOD \mnli{} that was not present in the unlabelled training data (Figure \ref{fig:ablating_AL_agg_OD}, top left) and still find that \textit{AL fails to outperform \random{}}. 

Analyzing the acquired data in terms of input diversity, uncertainty and label distributions (Table~\ref{table:single_source_basic_metrics})
There do not appear to be clear relations between metric outcomes and strategy \textit{performance} otherwise, i.e.,\ acquiring diverse or uncertain batches does not seem to lead to higher performance. 

\section{Dataset Cartography for AL Diagnosis}\label{section:cartography}

\begin{figure*}[t]
\centering
\includegraphics[width=0.48\linewidth]{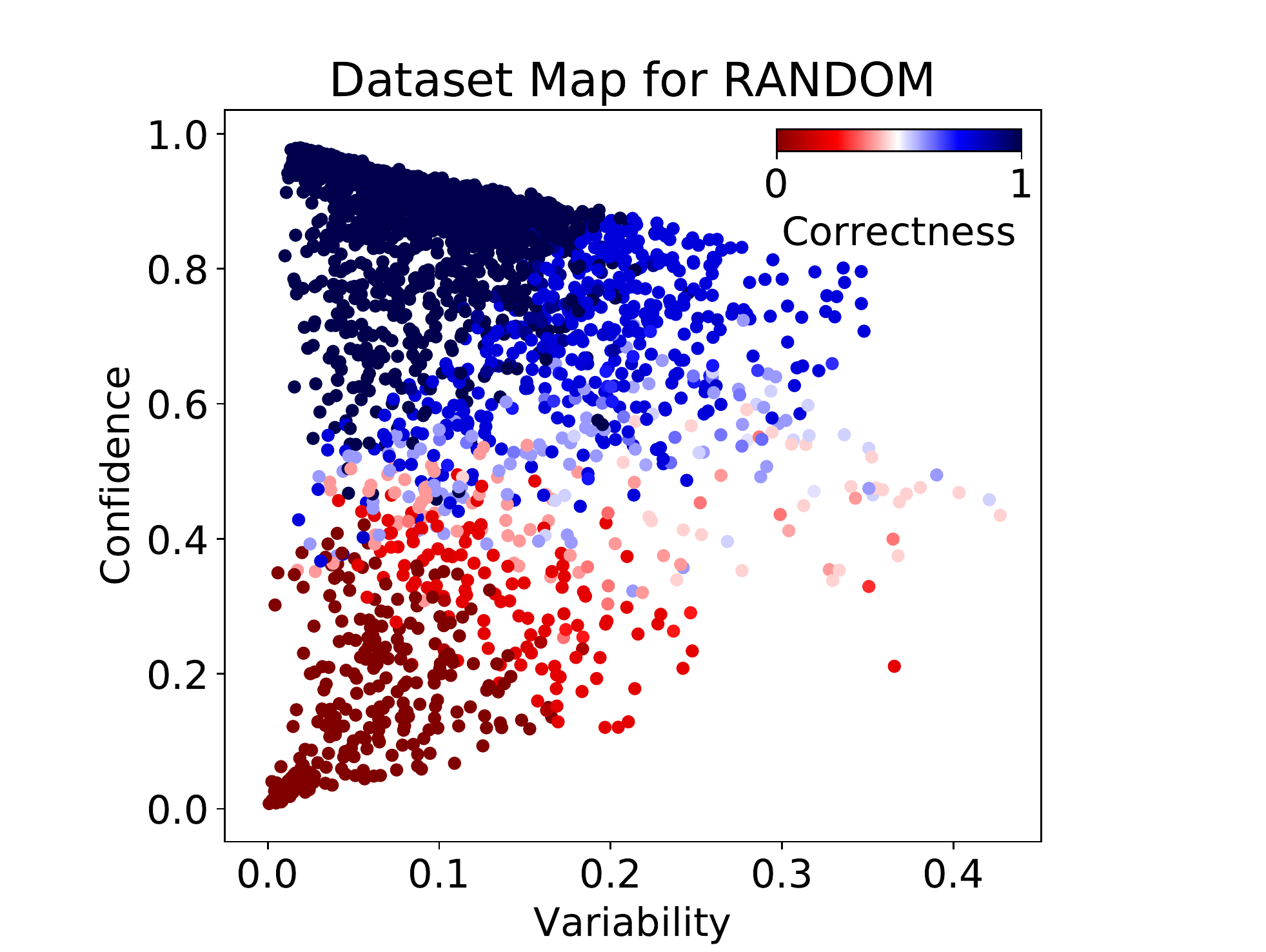}
\includegraphics[width=0.48\linewidth]{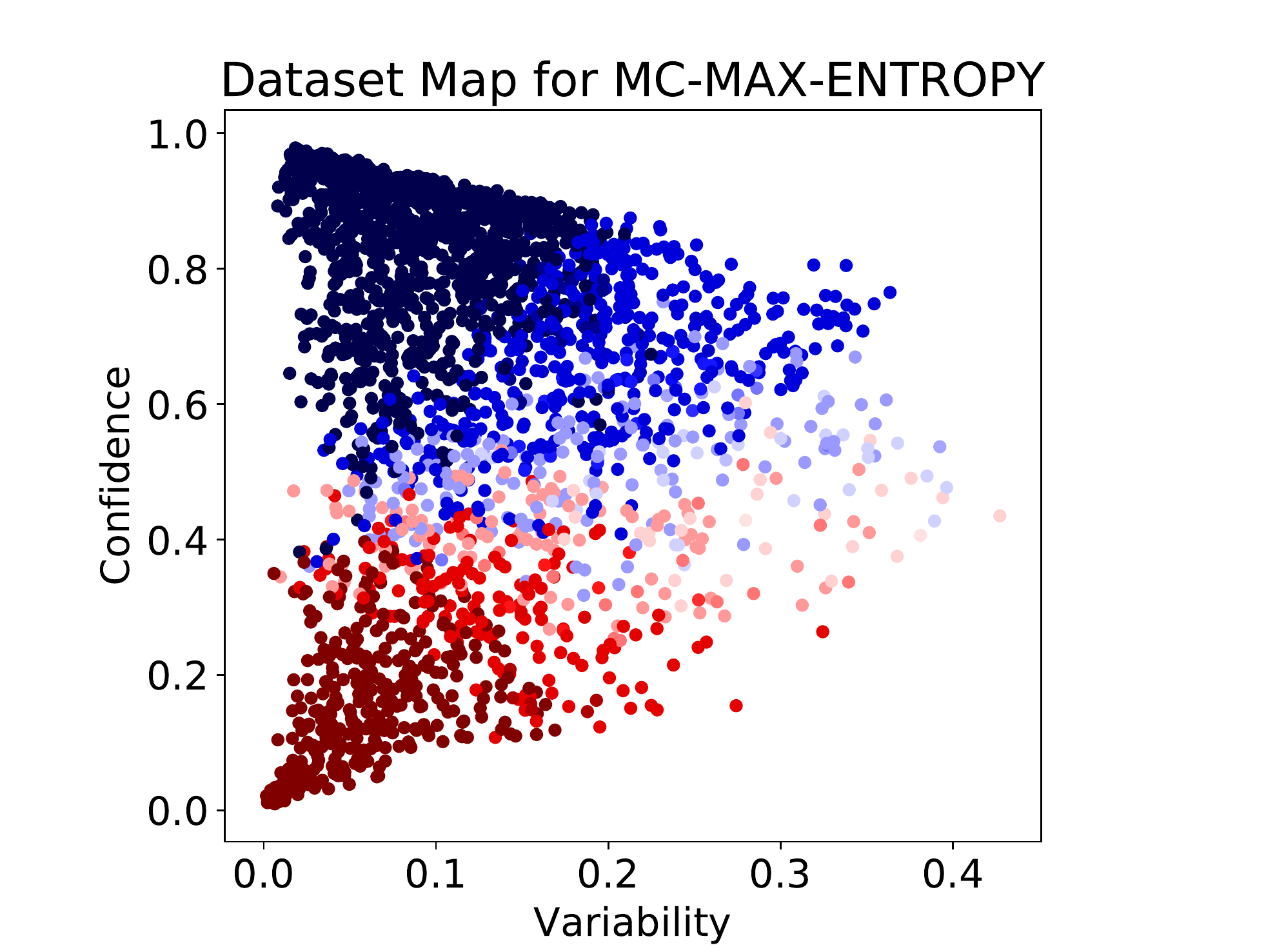}
\vfill
\includegraphics[width=0.22\linewidth]{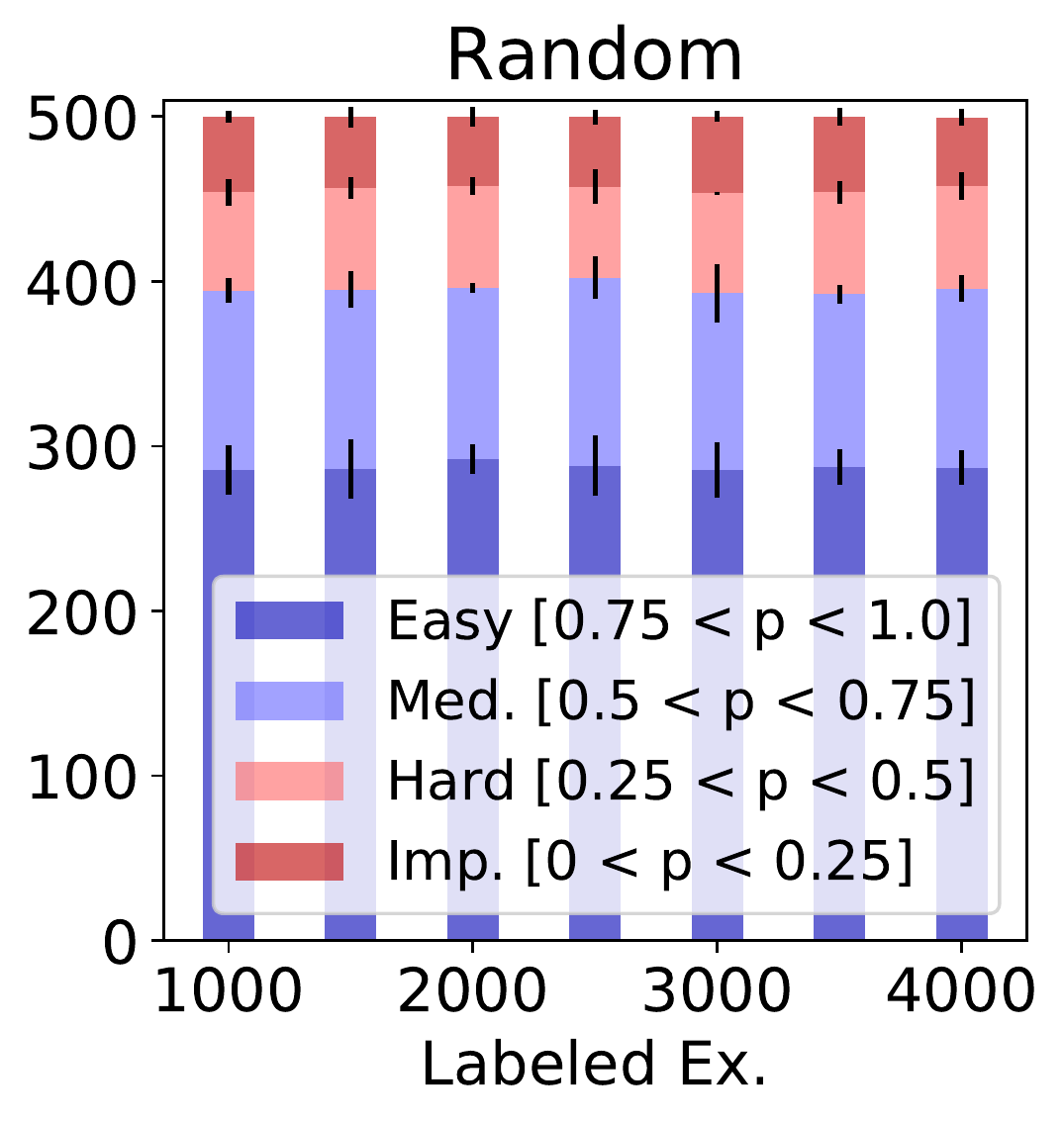}
\includegraphics[width=0.22\linewidth]{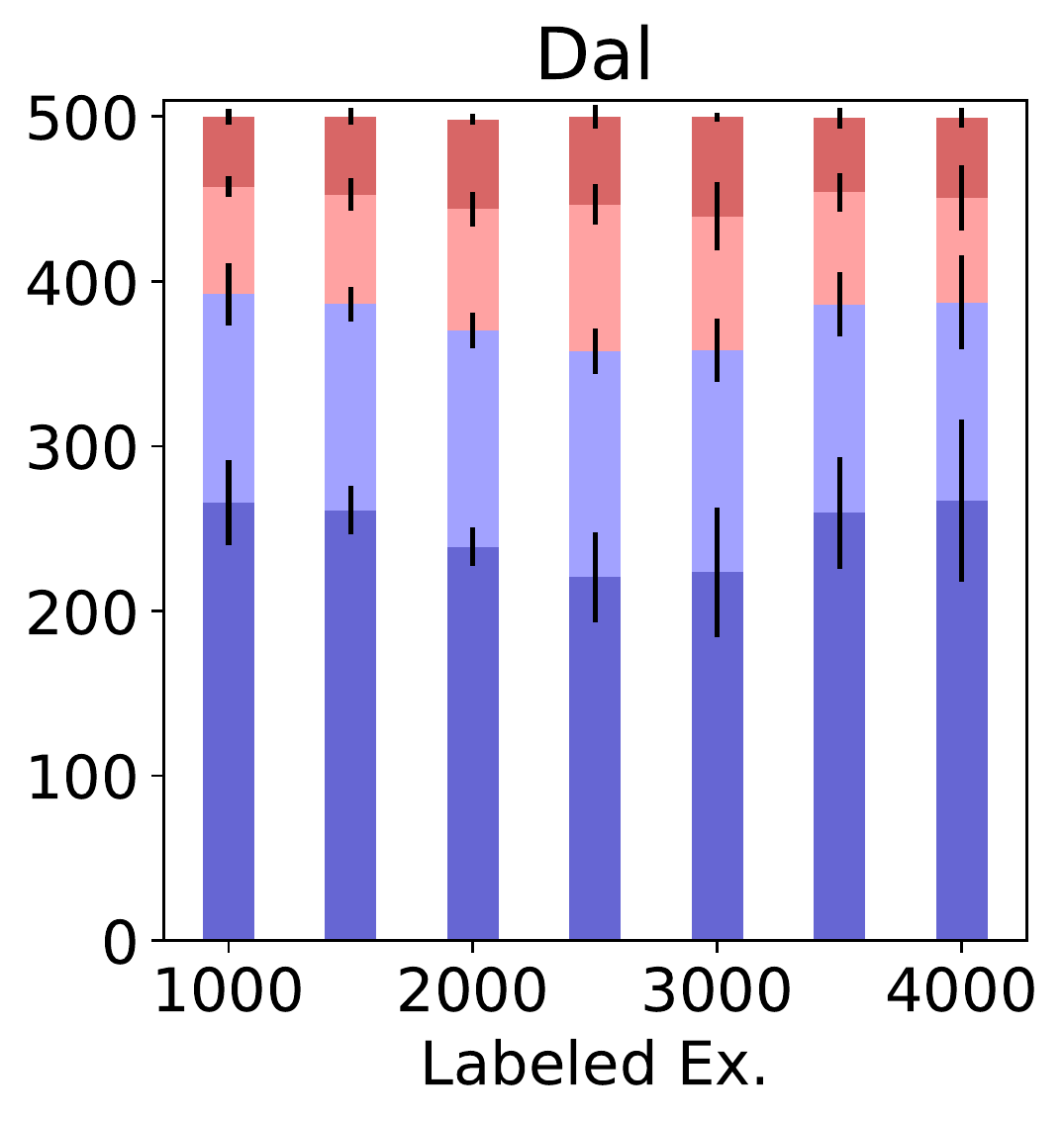}
\includegraphics[width=0.22\linewidth]{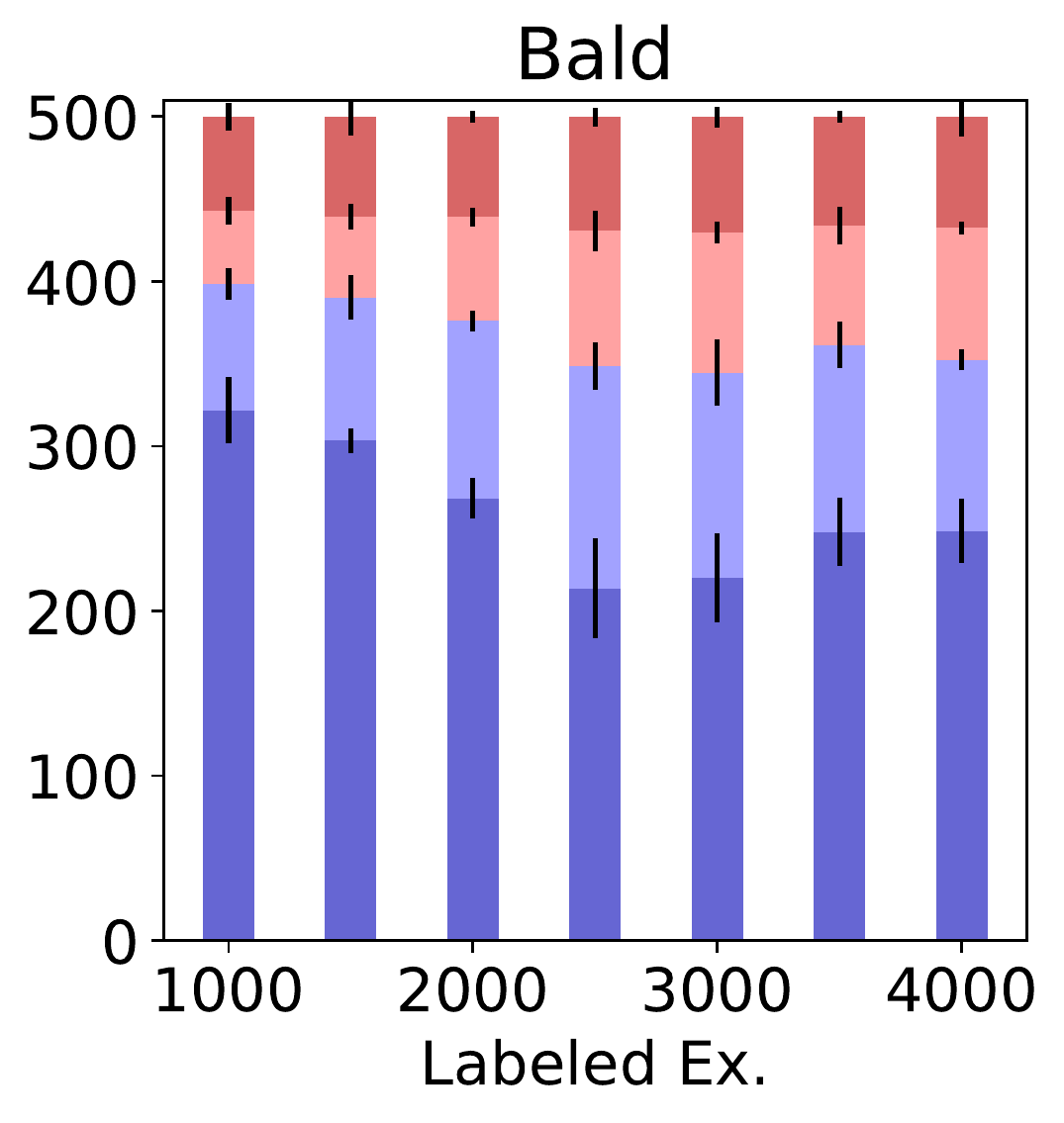}
\includegraphics[width=0.22\linewidth]{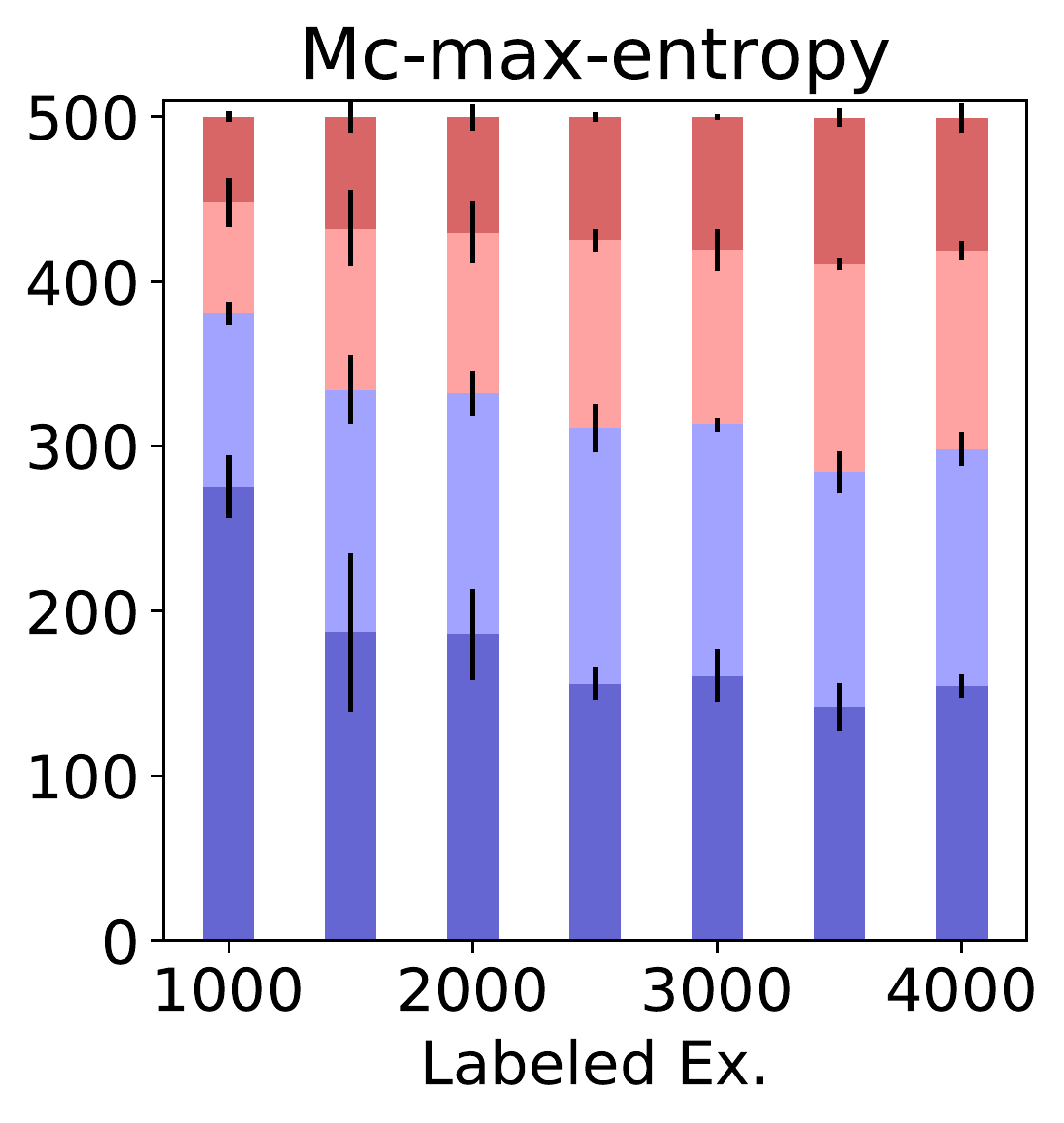}
\vfill
\caption{Top row: Strategy maps for \random{} and \mcme{} for \textbf{multi-source} AL. \textit{Correctness} denotes how often the model predicts the correct label. Bottom row: Acquisition by difficulty over time for \random{}, \dal{}, \bald{} and \mcme{}. We provide additional datamap plots in the appendix (Section \ref{sec:datamaps_appendix}).}
\label{fig:cartography_strategies_main}
\end{figure*}

Our findings in both single and multi-source AL settings, for both in-domain and out-of-domain evaluations, showed poor performance of all algorithms (\S\ref{section:main_results}). We therefore aim to investigate if the answer for the observed AL failure may lie in the presence of so-called \textit{collective outliers}: examples that models find hard to learn as a result of factors, such as high ambiguity, underspecification, requirement of specialist skills or labelling errors \cite{han2000dm}. Collective outliers can be identified through dataset cartography \citep{Swayamdipta2020}, a post-hoc model-based diagnostic tool which plots training examples along a so-called \textit{learnability spectrum}. 

\subsection{Creating Datamaps}\label{section:cartography_method}
Dataset Cartography assumes that the learnability of examples \textit{relative} to some model can be quantified by leveraging \textit{training dynamics}, where for each example we measure at fixed intervals (1) the mean model \textit{confidence} for the gold-truth label throughout training and (2) the \textit{variability} of this statistic. After gathering these statistics we can plot datasets on a \textbf{datamap}: a 2D graph with mean confidence on the Y-axis and confidence variability on the X-axis. The resulting figure enables us to identify how data is distributed along a \textit{learnability spectrum} (see an example in Figure~\ref{fig:cartography_strategies_main}).

We construct datamaps by training 
a model on the entire pool, 
i.e.\ $60$K examples in total. Every $\frac{1}{2}$ epoch we perform inference on the full training set to get per-example confidence statistics, where the prediction logit corresponding to the gold-truth label serves as a proxy for model confidence. Variability is computed as the standard deviation over the set of confidence measurements. 
Following \citet{Karamcheti2021}, we classify examples along four difficulties via a threshold on the mean confidence value $p$ (Figure \ref{fig:cartography_strategies_main}).
We provide more details in the Appendix~\ref{sec:datamaps_appendix}.

\subsection{Strategy Maps for Multi-source AL}\label{par:strat_maps}
We show the datamaps for random sampling and the \mcme{} acquisition functions on the multi-source AL setting in the top row of Figure \ref{fig:cartography_strategies_main}
(see Appendix~\ref{sec:datamaps_appendix} for all datamaps).
Examining the first datamap, it appears that \textit{randomly sampling from all sources in equal proportions tends to yield predominantly easy-to-learn instances}, with diminishing amounts of ambiguous and hard-to-learn instances. Conversely, we find that \mcme{} acquires considerably more examples with moderate to low confidence, suggesting a \textit{tendency to acquire hard and impossible examples}. Note that these outcomes mirror the findings by \citet{Karamcheti2021} for VQA. Observing the per-difficulty acquisition over time (Figure \ref{fig:cartography_strategies_main} bottom row), we observe that \bald{} and \mcme{} \textit{initially favour easy examples}, but as more examples are acquired, we observe a \textit{shift towards medium, hard and impossible examples}. One explanation for these trends is that in early phases of AL, models have only been trained on small amounts of data. At this stage, model confidence is still poorly calibrated and consequently confidence values may be noisy proxies for uncertainty. As the training pool grows, model confidence will more accurately reflect example difficulty enabling uncertainty-based strategies to identify difficult data points.

\begin{table*}[h!]
\small
\begin{tabularx}{\linewidth}{
    >{\hsize=0.18\hsize}X
    >{\hsize=2.3\hsize}X
    >{\hsize=1.2\hsize}X
    >{\hsize=0.06\hsize}X
    >{\hsize=0.06\hsize}X
    >{\hsize=0.18\hsize}X
  }
\toprule
\textbf{Src.} & \textbf{Premise} & \textbf{Hypothesis} & \textbf{GT}  & \textbf{P} & \textbf{Conf.} \\
\midrule
\snli{}
&
\texttt{A skier in electric green on the edge of a ramp made of metal bars.}	
&
\texttt{The skier was on the edge of the ramp.}	
&
N
&
E
&
$0.976$
 \vspace*{1.2mm} \\
\snli{}
&
\texttt{Man sitting in a beached canoe by a lake.}	
&
\texttt{a man is sitting outside}
&
N
&
E
&
$0.980$
 \vspace*{1.2mm} \\
\wanli{}
&
\texttt{The first principle of art is that art is not
a way of life, but a means of life.}
&
\texttt{Art is a way of life.}
&
E
&
C
&
$0.944$
 \vspace*{1.2mm} \\
\wanli{}
&
\texttt{Some students believe that to achieve their goals they must take the lead.}
&
\texttt{Some students believe that to achieve their goals they must follow the lead.}
&
E
&
C
&
$0.630$
 \vspace*{1.2mm} \\
\anli{}
&
\texttt{Marwin Javier González (born March 14, 1989) is a Venezuelan professional baseball infielder with the Houston Astros of Major League Baseball (MLB). Primarily a shortstop, González has appeared at every position except for pitcher and catcher for the Astros.}
&
\texttt{He is in his forties.}	
&
C	
&
N
&
$0.769$
 \vspace*{1.2mm} \\
\anli{}
&
\texttt{The Whitechapel murders were committed in or near the impoverished Whitechapel district in the East End of London between 3 April 1888 and 13 February 1891. At various points some or all of these eleven unsolved murders of women have been ascribed to the notorious unidentified serial killer known as Jack the Ripper.}
&
\texttt{The women killed in the Whitechapel murders were impoverished.}	
&
N	
&
E
&
$0.832$
\\
\bottomrule
\end{tabularx} 
\caption{Impossible NLI training examples.
\textbf{GT} denotes the gold truth label; \textbf{P} denotes the model prediction and \textbf{Conf.} denotes the prediction confidence. We provide more examples per source in the appendix (Section \ref{section:impossible_data}).
}\label{tab3}
\label{table:imp_snli}
\end{table*}

\section{Stratified Analysis on Data Difficulty}

We begin our analysis of collective outliers by exploring impossible data points in the three NLI datasets we use; \snli{}, \anli{} and \wanli{}. We denote as impossible data points those that yield a confidence value in the range $0 \leq p \leq 0.25$ (\S\ref{section:cartography_method}). We provide some examples in Table \ref{table:imp_snli} and in the Appendix \ref{section:impossible_data}. We find that impossible examples from \snli{} and \wanli{} are more prone to suffer from label errors and/or often lack a clear correct answer, which may explain their poor learnability. Conversely, we find impossible \anli{} examples to exhibit fewer of these issues - we hypothesize that their difficulty follows rather from requirement of advanced inference types, e.g.\ identifying relevant information from long passages and numerical reasoning about dates of birth and events.

\subsection{Examining the effect of training data difficulty}\label{section:training_difficulty}
Now that we are able to classify training examples as easy, medium, hard and impossible (\S\ref{par:strat_maps}), we proceed to explore the effect of data difficulty on learning and per-source outcomes. In this set of experiments, we aim to answer the research question: \textit{What data would the most beneficial training set consist of, in terms of data difficulty per example?} We conventionally (i.e., non-AL experiment) train RoBERTa-large on training sets of various difficulties. Each training set comprises $4$K examples, i.e.\ the same amount of examples that would be acquired after all rounds of AL. 
We consider the following combinations of data: \textbf{EM}, \textbf{EMH}, \textbf{MH}, \textbf{HI} and \textbf{EMHI}, where \textbf{E} denotes \textit{easy}, \textbf{M} \textit{medium}, \textbf{H} \textit{hard} and \textbf{I} \textit{impossible} examples.\footnote{For a given run, examples are sampled from each difficulty in equal proportion. For instance, when training on the \textit{easy-medium} (\textbf{EM}) split the training set comprises $2$K easy-to-learn instances and $2$K medium-to-learn instances. }
\paragraph{Results}
We provide test results for all combinations of training data difficulty in Table \ref{fig:ablating_AL_test_difficulty}.
We first observe that models trained only on \textbf{HI} data consistently perform the worst, resulting in a drop of up to \textasciitilde$50\%$(!) points in accuracy compared to the \textbf{EM} split for \snli{} and \textasciitilde$40\%$ for \mnli{}.
Surprisingly, we find that 
models trained on \textbf{MH} perform worse than splits that include \textit{easy} examples, except for \anli{}.
Intuitively, this makes sense: of all datasets, \anli{} features the most difficult examples, and thus it is plausible that hard-to-learn instances translate to more learning gains than easy ones. The \textbf{EMHI} split slightly underperforms relative to the \textbf{EM} and \textbf{EMH} splits. Otherwise, we do not observe great differences between the latter two splits. In the second part of Table \ref{fig:ablating_AL_test_difficulty}, we compute the average performance for easy test sets (\snli{} $\cup$ \mnli{}), hard (\anli{} $\cup$ \wanli{}) and all test sets combined. We can now observe very clear patterns. The more difficult data points we include in the training set, the more the performance on an easy test set drops (\textsc{avg-easy}$\downarrow$). Similarly, when testing on a harder test set, the more difficult the training data the better (\textsc{avg-hard}$\downarrow$), but without including data points that are characterized as \textit{impossible}.
\textit{Overall, we observe a strong correlation between training and test data difficulty, and we conclude that we should train models on data of the same difficulty as the test set, while always removing data points that are impossible to be learned from the model (i.e. collective outliers)}.

\begin{table*}\centering
\resizebox{0.9\textwidth}{!}{
\begin{tabular}{|c|c|c|c|c||c|c|c|c|}
\hline
             & \textbf{\snli{}}                    & \textbf{\anli{}}                         & \textbf{\wanli{}}                    & \textbf{\mnli{}} & \textbf{\textsc{avg-easy}} & \textbf{\textsc{avg-hard}} & \textbf{\textsc{avg-all}} \\ \hline \hline
\textbf{EM}  & $\textbf{85.95} \pm \textbf{0.90}$  & $36.85\pm 1.05$                          & $\textbf{68.34} \pm \textbf{0.68}$   & $\textbf{76.16} \pm \textbf{2.32}$  & $\textbf{81.06}$ & $52.60$ & $\textbf{66.83}$\\ \hline
\textbf{EMH} & $84.98 \pm 0.56$                    & $37.84\pm 0.79$                          & $68.16\pm 0.66$                      & $75.50\pm 2.00$ &$80.24$ &$53.00$ &$66.62$\\ \hline
\textbf{MH}  & $76.32 \pm 1.04$                    & $\textbf{39.11} \pm \textbf{0.79}$       & $63.56\pm 1.38$                      & $69.82\pm 6.10$ & $73.07$ & $\textbf{54.47}$ & $62.20$ \\ \hline
\textbf{HI}  & $33.57 \pm 1.60$                    & $34.00 \pm 0.69$                         & $45.89 \pm 3.85$                     & $32.69 \pm 1.93$  & $33.13$ & $39.95$ & $36.54$\\ \hline
\textbf{EMHI}& $79.12 \pm 0.80$                    & $38.41\pm 0.57$                          & $64.52\pm 1.38$                      & $73.52 \pm 0.07$  & $76.32$ & $51.47$ & $63.89$\\ \hline
\end{tabular}
}
\caption{\textbf{Multi-source} test outcomes for training data of \textit{varying difficulty}. Row headers denote difficulty splits, with \textbf{E} denoting \textit{easy}, \textbf{M} \textit{medium}, \textbf{H} \textit{hard} and \textbf{I} \textit{impossible} examples. Column headers denote the test set. We also compute the average over \snli{}+\mnli{} (easy), \anli{}+\wanli{} (hard) and all $4$ datasets, respectively.
}
\label{fig:ablating_AL_test_difficulty}
\end{table*}
\subsection{Ablating Outliers for AL}\label{section:ablation}
Having uncovered the effects of data difficulty on learning outcomes, we now examine \textit{how the presence or absence of hard and impossible instances affects AL strategy success}. Specifically, we repeat our multi-source AL experiment (\S\ref{section:ms_results}) whilst excluding hard and impossible (\textbf{HI}) examples from $\mathcal{D}_{pool}$. 
Employing datamap statistics, we compute the product of an example's mean confidence and its variability for the entire unlabelled pool, after which we exclude the $25\%$ of examples with the smallest products, following \citet{Karamcheti2021}. Examples are filtered out for each source dataset separately to preserve equivalence in source sizes.
\paragraph{Results}
Examining the learning curves (Figure \ref{fig:ablating_AL_agg_OD}, right) and the final test results (Table~\ref{fig:ablating_AL_test}, left), it appears that \textit{excluding \textbf{HI} examples particularly affects the performance of uncertainty-based acquisition strategies} (\mcme{} and \bald{}): across sources, both strategies do consistently \textit{better}. Moreover, for \anli{}, \mcme{} clearly outperforms random selection and even achieves a higher accuracy compared to the non-ablated run. Similar results are obtained for the OOD \mnli{} dataset: \textit{previously, none of the strategies consistently outperformed random, but after ablating \textbf{HI} instances both \bald{} and \mcme{} are either on par with random or outperform it}. 
%
Intuitively, under normal circumstances \mcme{} and \bald{} tend to acquire more difficult examples than \random{} or \dal{}, which typically leads to \textit{poorer} models (\S\ref{section:training_difficulty}). Therefore, we should expect to see \textit{improvements} when these instances are ablated (i.e., removed).

\subsection{Stratified Testing}\label{section:stratified_testing}
Having established that some strategies acquire more hard and examples than others (\S\ref{par:strat_maps}), training on such examples can both hurt or help generalization (\S\ref{section:training_difficulty}), and removal of this data can help strategies improve (\S\ref{section:ablation}), a question arises: Do strategies that predominantly acquire data of a certain difficulty also perform better on \textit{test data} of that difficulty? To investigate this we introduce \textbf{stratified testing}: stratifying test outcomes \textit{in terms of difficulty}. Here, we follow the approach as outlined in Section \ref{section:cartography_method}, but this time training a cartography model on the \textit{test set} to obtain learnability measurements for each \textit{test} example. This enables us to observe how strategies perform across test examples of varying difficulties.

\paragraph{Results} Table \ref{tab:stratified_test_merge} shows the AL results for stratified testing for both \textit{ablated} (i.e., \textbf{HI} removed) and the \textit{original} setting.
We observe in the pre-ablation experiments (right side) that \random{} and \dal{} tend to do better on easy and medium data relative to \mcme{} and \bald{}, while conversely they underperform when tested on hard and impossible examples.
The same pattern also occurs also in the OOD setting (\mnli{}).
%
In the ablation setting (left side), 
we find that across all sources, \textbf{HI} examples tends to yield a performance drop on hard and impossible examples for all strategies. 
Corroborating our previous analysis (\S\ref{section:training_difficulty}), our findings here suggest that \textit{it is essential to have data of the same difficulty in both unlabelled pool and in test set}.
Next, we find that under ablation, \random{} does moderately better on easy and medium, while \dal{} shows marginally lower test outcomes on medium and hard examples. For the uncertainty-based strategies, ablation tends to yield improvements across all difficulties, with \mcme{} outperforming \bald{} on hard examples. We hypothesize that with \textbf{HI} instances absent, uncertainty-based methods select the ``next-best'' available data: medium examples which offer greater learning gains than easy ones.
Overall, we find that post-ablation, \bald{} and \mcme{} outperform both \random{} and \dal{} across most difficulty splits.

\begin{table}[t]
\centering
\resizebox{\columnwidth}{!}{\begin{tabular}{|c|ll|c|c|c|c|}
\hline
\multicolumn{1}{|l|}{Task} & \multicolumn{2}{c|}{D}& \textbf{\random{}} & \textbf{\mcme{}} & \textbf{\bald{}} & \textbf{\dal{}} \\ 
\hline \hline
                      & \multicolumn{2}{c|}{\textbf{E}}                 &  $95.4 \mid 95.0 $              & $95.0 \mid 93.3 $               & $95.5 \mid 94.8 $           & $93.7 \mid 94.6 $ \\ 
    \textbf{\snli{}}     & \multicolumn{2}{l|}{\textbf{M}}                 &  $69.0 \mid 67.7 $              & $70.2 \mid 68.0 $               & $70.5 \mid 69.5 $           & $67.8 \mid 68.2 $ \\ 
                      & \multicolumn{2}{c|}{\textbf{H}}                 &  $46.3 \mid 44.6 $              & $48.5 \mid 46.0 $               & $47.1 \mid 49.2 $           & $44.7 \mid 45.8 $ \\ 
                      & \multicolumn{2}{c|}{\textbf{I}}                 &  $17.7 \mid 17.2 $              & $20.3 \mid 22.1 $               & $20.7 \mid 22.2 $           & $18.4 \mid 17.5 $ \\ \hline \hline
                       
                      & \multicolumn{2}{c|}{\textbf{E}}                 &  $81.4 \mid 81.7 $              & $77.0 \mid 79.2 $               & $82.1 \mid 73.8 $           & $81.8 \mid 79.9 $ \\ 
    \textbf{\anli{}}     & \multicolumn{2}{c|}{\textbf{M}}                 &  $62.2 \mid 62.1 $              & $61.7 \mid 57.8 $               & $61.1 \mid 57.1 $           & $63.1 \mid 59.8 $ \\ 
                      & \multicolumn{2}{c|}{\textbf{H}}                 &  $39.3 \mid 40.2 $              & $42.9 \mid 41.6 $               & $39.2 \mid 39.7 $           & $39.0 \mid 38.5 $ \\ 
                      & \multicolumn{2}{c|}{\textbf{I}}                 &  $14.4 \mid 14.2 $              & $15.7 \mid 17.0 $               & $12.6 \mid 14.5 $           & $14.2 \mid 15.0 $ \\ \hline \hline
                       
                      & \multicolumn{2}{c|}{\textbf{E}}                 &  $94.3 \mid 92.9 $              & $92.7 \mid 91.2 $               & $95.2 \mid 90.9 $           & $94.1 \mid 93.1 $ \\ 
    \textbf{\wanli{}}    & \multicolumn{2}{c|}{\textbf{M}}                 &  $72.7 \mid 69.6 $              & $71.4 \mid 69.9 $               & $72.7 \mid 67.9 $           & $72.3 \mid 73.8 $ \\ 
                      & \multicolumn{2}{c|}{\textbf{H}}                 &  $41.9 \mid 40.4 $              & $43.1 \mid 44.2 $               & $42.4 \mid 40.7 $           & $40.6 \mid 41.0 $ \\ 
                      & \multicolumn{2}{c|}{\textbf{I}}                 &  $12.7 \mid 14.4 $              & $14.0 \mid 16.6 $               & $11.9 \mid 14.7 $           & $13.9 \mid 14.6 $ \\ \hline

    \hline
    \hline

                      & \multicolumn{2}{c|}{\textbf{E}}                 &  $93.3 \mid 93.6 $              & $94.3 \mid 92.8 $               & $94.4 \mid 92.4 $           & $93.4 \mid 93.8 $ \\ 
    \textbf{\mnli{}}     & \multicolumn{2}{c|}{\textbf{M}}                 &  $72.6 \mid 72.9 $              & $78.6 \mid 74.7 $               & $75.9 \mid 72.8 $           & $74.7 \mid 78.1 $ \\ 
                      & \multicolumn{2}{c|}{\textbf{H}}                 &  $41.8 \mid 44.5 $              & $51.9 \mid 49.3 $               & $45.7 \mid 46.7 $           & $46.1 \mid 48.3 $ \\ 
                      & \multicolumn{2}{c|}{\textbf{I}}                 &  $12.8 \mid 15.7 $              & $20.9 \mid 19.5 $               & $13.1 \mid 16.9 $           & $12.6 \mid 18.5 $ \\ \hline

    \hline \hline
                      
                      & \multicolumn{2}{c|}{\textbf{E}}                 &  $91.0 \mid 90.8 $              & $89.8 \mid 89.1 $               &  $91.8\mid 88.0 $           & $90.7 \mid 90.4 $ \\ 
    \textbf{All}      & \multicolumn{2}{c|}{\textbf{M}}                 &  $69.3 \mid 68.1 $              & $70.5 \mid 67.6 $               &  $70.0\mid 66.8 $           & $69.4 \mid 69.9 $ \\ 
                      & \multicolumn{2}{c|}{\textbf{H}}                 &  $42.3 \mid 42.4 $              & $46.5 \mid 45.3 $               &  $43.6\mid 44.1 $           & $42.6 \mid 43.4 $ \\ 
                      & \multicolumn{2}{c|}{\textbf{I}}                 &  $14.4 \mid 15.3 $              & $17.7 \mid 18.8 $               &  $14.6\mid 17.1 $           & $14.8 \mid 16.4 $ \\ \hline
                       
\end{tabular}}
\caption{\textbf{Multi-source} strategy comparison (accuracy) for difficulty-stratified test outcomes between original training set and training set with hard-to-learn examples excluded. Test examples (D) belong to Easy (\textbf{E}), Medium (\textbf{M}), Hard (\textbf{H}) and Impossible (\textbf{I}) sets. Cell scheme to be read as \textit{ablated} $\mid$ \textit{original}.}
\label{tab:stratified_test_merge}
\end{table}

\section{Conclusion and Future Work}
Our work highlights the challenge of successfully applying AL when the (training) pool comprises several sources which span various domains in the task of NLI.
Similar to \citet{Karamcheti2021}, we show that uncertainty-based strategies, such as \mcme{} and \bald{}, perform poorly (\S\ref{section:main_results}) due to acquisition of collective outliers which impede successful learning (\S\ref{par:strat_maps}). However, these strategies recover when outliers are removed (\S\ref{section:ablation}).
Practically, this suggests that uncertainty-based methods may fare well under more carefully curated datasets and labelling schemes,
while alternative strategies (e.g. diversity-based) may be preferable in cases with poorer quality guarantees (e.g.\, data collection with limited budget for annotation verification). 
Next, we find that performance outcomes between strategies differ for test data of various difficulties (\S\ref{section:stratified_testing}). On the one hand, this complicates strategy selection: it is unclear whether a strategy that performs well on hard data but poorer on easy data is preferable to a strategy with opposite properties. On the other hand, knowing which strategies work well for test data of a certain difficulty may be advantageous when the difficulty of the test set is known, in out-of-domain settings \cite{Lalor2018,genbench}.

Lastly, in contrast with \citet{Karamcheti2021} and \citet{Zhang}, we have shown that cases exist in which training examples in the hard-to-learn region do not hamper learning but are in fact pivotal for achieving good generalization (\S\ref{section:training_difficulty}). Consequently, there may be value in refining existing cartography-based methods such that they can discriminate between useful and harmful hard-to-learn data. More broadly, our findings underscore the potential of understanding these phenomena for other NLP tasks and datasets. 

\section*{Acknowledgements}
We thank Dieuwke Hupkes for helping during the initial stages of this work. The presentation of this paper was financially supported by the Amsterdam ELLIS Unit and Qualcomm. We also thank \href{www.surf.nl}{SURF} for the support in using the Lisa Compute Cluster. Katerina is supported by Amazon through the Alexa Fellowship scheme.
\section*{Limitations}
In our work, we have shown that standard AL algorithms struggle to outperform random selection in some datasets for the task of NLI, which is in fact rather surprising as a large body of work has shown positive AL results in a wide spectrum of NLP tasks. Still, we are not the first to show negative AL results in NLP, as \citet{Lowell,Karamcheti2021,Kees} have shown similar problematic behavior in the cases of text classification, visual question answering and argument mining, respectively.


More broadly, while AL outcomes have shown to not always reliably generalize across models and tasks \citep{Lowell}, we recognise that in our work several experimental conditions remain under-explored which warrant further attention. First, this work mostly examined point-wise acquisition functions; it remains unclear whether our outcomes hold for \textit{batch-wise} functions. Similarly, it is unknown how the chosen model-in-the-loop affects strategy outcomes. We use RoBERTa-large, a comparatively powerful large language model. Some authors hypothesize that as such models are already able to achieve great performance even with randomly labeled data, this could significantly raise the bar for AL algorithms to yield substantial performance improvements \textit{on top} of a random selection baseline \citep{Ein-Dor}. Combined with the relative homogeneity of acquired batches between sources in terms of input diversity, this would explain why strategies tend to do similar across the board at test-time both in the presence and absence of hard-to-learn examples. Another factor tying into this is that small differences between results may be connected to the so-called inverse cold-start problem. This problem states that there may be cases where the initial seed set is too large, leaving comparatively little room for substantial improvements in sample efficiency.  We leave further exploration of these variables to future work.

Another area within AL research which warrants further examination concerns the evaluation on out-of-domain datasets of which no data is present in the pool of unlabelled training data. Particularly, the majority of work typically assumes that acquisition is target-agnostic, i.e., target validation and test sets are assumed to be decoupled entirely from the acquisition process. This can be problematic as for different target sets, different subsets of the unlabelled training data may yield the best possible performance. Consequently, performance outcomes on \textit{some} out-of-domain test data may not necessarily pose as reliable signals for determining the best strategy, for if a different target set had been chosen, a previously 'poor' performing strategy may suddenly achieve the best result. Despite this being a clear shortcoming of the existing AL toolkit, it remains an understudied area within AL research. 

While this problem may be partially alleviated by evaluating strategies on a large and diverse array of target sets, the issue remains that current acquisition functions do not acquire data \textit{with respect to the target set}. This problem becomes even more apparent in a \textit{multi-source} setting, where depending on the target set at hand, acquiring data from the appropriate source(s) may be pivotal to achieve good performance. In such cases, we may want to explicitly regularise the acquisition process towards target-relevant training data. A small body of work has examined ways in which such target-aware acquisition could be formalized - most noticeably in the work of \citet{Kirsch2021}, who introduce several methods to perform test-distribution aware active learning. While examination of such methods lies outside the scope of this work, we recognize its potential for future work on multi-source AL.
\bibliography{anthology,eacl2023}
\bibliographystyle{acl_natbib}

\clearpage 
\section{Appendix}

\subsection{Analysis metrics}\label{sec:app_analysis_metrics}
Following standard practice in active learning literature \citep{Zhdanov2019,Yuan2020, Ein-Dor, Margatina_cal} we profile datasets acquired by strategies via acquisition metrics. Concretely, we consider the \textit{input diversity} and \textit{output uncertainty} metrics. 
\paragraph{Input Diversity}
Input diversity quantifies the diversity of acquired sets in the input space, meaning that it operates directly on the raw input passages. We follow \cite{Yuan2020} and measure input diversity as the Jaccard similarity $\mathcal{J}$ between the set of tokens from the acquired training set $\mathcal{D}_{train}$, $\mathcal{V}$, and the set of tokens from the \textit{remainder} of the unlabelled pool $\mathcal{D}_{pool}$, $\mathcal{V}\prime$, which yields:
$$\mathcal{J}(\mathcal{V}, \mathcal{V}\prime) = \frac{\mid \mathcal{V} \cap \mathcal{V}\prime \mid}{\mid \mathcal{V} \cup \mathcal{V}\prime\mid}$$
This function assigns \textit{high} diversity to strategies acquiring samples with \textit{high} token overlap with the unlabelled pool, and vice versa.
\paragraph{Output Uncertainty}
To approximate the output uncertainty of an acquired training set $\mathcal{D}_{train}$ for a given strategy, we train RoBERTa-large to convergence on the entire 60K training set. We then use the trained model to perform inference over $\mathcal{D}_{train}$. Following \citep{Yuan2020}, the output uncertainty of each strategy is computed as the mean predictive entropy over all examples $x$ in its acquired set $\mathcal{D}_{train}$:
 $$ -\frac{1}{|\mathcal{D}_{train}|} \sum_{x \in \mathcal{D}_{train}} \sum_{c=1}^{C} p(y=c \mid x) \log p(y=c \mid x)$$
 
\subsection{Datasets}\label{sec:app_datasets}
As mentioned in the paper (\S\ref{sec:datasets}), we perform experiments on Natural Language Inference (NLI), a popular text classification task to gauge a model's natural language understanding \citep{Bowman, Williams2017}. We recognize that NLI is somewhat artificial by nature - making it of lesser practical relevance for real-life active learning scenarios. However, recent work has sought to address shortcomings of existing NLI benchmarks such as \snli{} \citep{Bowman} and \mnli{} \citep{Williams2017}. This has lead to the emergence of novel approaches to dataset-creation such as Dynamic Adversarial Data Collection (\anli{}, \citep{Nie2019}) and worker-and-AI-collaboration (\wanli{}, \citep{Liu2022}). As we seek to investigate how characteristics of data gathered through such alternative protocols may affect acquisition performance in a multi-source active learning setting, using NLI for our experiments is a natural choice. 
We construct the unlabelled pool from three distinct datasets: \snli{}, \anli{} and \wanli{}.
Next, we consider the Multi Natural Language Inference (\mnli{}) corpus \citep{Williams2017} as an out-of-domain challenge set to evaluate the transferability of actively acquired training sets. We provide datasets statistics in Table \ref{fig:dataset_statistics}.

\begin{table*}[t]
\centering
\resizebox{\textwidth}{!}{\begin{tabular}{c | c c c c | c c c c | c c c c}
\multicolumn{13}{c}{\textbf{Label Distributions}}    \\ 
\cline{1-13}
\toprule
\textbf{Source}  &  \multicolumn{4}{c}{\textbf{Train}}         &  \multicolumn{4}{c}{\textbf{Val}} &  \multicolumn{4}{c}{\textbf{Test}}        \\ \cline{1-13}
{}               &    N    &    E    &   C   &   \textbf{Size} &    N   &    E   &   C   &   \textbf{Size}   &   N    &    E   &   C   &   \textbf{Size}  \\ \hline 
\textbf{SNLI}    &  183.4K   &  182.7K   &  183.1K &   549.5K  &  3.2K  &  3.3K  &  3.3K &   9.8K            &  3.2K   &  3.4K  &  3.2K  &   9.8K   \\ \hline
\textbf{ANLI}    &  61.7K    &  46.7K    &  37.4K  &   146K    &  0.7K  &  0.7K  &  0.7K &   2.2K            &  0.7K   &  0.7K  &  0.7K  &   2.2K   \\ \hline
\textbf{WANLI}   &  48.8K    &  39.1K    &  14.4K  &   103K    &  1.2K  &  0.9K  &  0.4K &   2.5K            &  1.2K   &  0.9K  &  0.4K  &   2.5K   \\ \hline
\textbf{MNLI-m}* &  n/a      &  n/a      &  n/a    &   n/a     &  n/a   &  n/a   &  n/a  &   n/a             &  1.6K   &  1.7K  &  1.6K  &   4.9K   \\ \hline
\end{tabular}}\caption{Statistics for the used datasets. N = Neutral, E = Entailment, C = Contradiction. For MNLI, only a held-out test set was used, and thus statistics for the training and dev set are omitted. Since WANLI only has a train and test set, we split its test set in two equally sized subsets and use one half as our validation set.}
\label{fig:dataset_statistics}
\end{table*}

\subsection{Training details \& Reproducibility}
We use RoBERTa-large \citep{Liu2020} from Huggingface \citep{Wolf2019} as our model-in-the-loop and optimize with AdamW \citep{loshchilov2018fixing}, with a learning rate of $2e-6$ and a batch size of 32. We use Dropout with $p=0.3$. Hyperparameters were chosen following a manual tuning process, evaluating models on classification accuracy.
For the BALD and MCME strategies we use 4 Monte Carlo Dropout samples. Our framework is implemented in PyTorch Lightning; transformers are implemented using the. All experiments were ran on a single NVIDIA Titan RTX GPU. Trialling all acquisition functions for 7 rounds of active learning (assuming experiments are ran in series), for a single seed, requires approximately 21 hours of compute. See Table \ref{fig:runtimes} for per-strategy runtimes.

\begin{table*}
\centering
\begin{tabular}{c|c c c c }
\hline
                    & \textbf{\random{}} & \textbf{\dal{}}     & \textbf{\bald{}} & \textbf{\mcme{}}                 \\ \hline
\textbf{Runtime}    &  $ 190 \pm 19 $ & $ 269 \pm 15 $   & $ 350 \pm  62  $ & $ 450 \pm 90 $ \\
\hline
\end{tabular}
\caption{Total runtime per strategy in minutes. Scheme follows \textit{mean} $\pm$ \textit{standard error} format.}
\label{fig:runtimes}
\end{table*}

\subsection{Detailed Results}
\begin{table*}[t]
\centering
\begin{tabular}{|c|ll|c|c||c|c|c|}
\hline
\multicolumn{1}{|l|}{\textbf{Task}} & \multicolumn{2}{l|}{\textbf{Strategy}}& \textbf{I-Div.} & \textbf{Unc.} & \textbf{N} & \textbf{E} & \textbf{C} \\ 
\hline \hline

                      & \multicolumn{2}{l|}{\random{}}  & $0.259 \pm 0.002$     & $0.051 \pm 0.002$  & 0.34  & 0.33  & 0.33 \\ 
    \textbf{\snli{}}      & \multicolumn{2}{l|}{\dal{}}    & $\textbf{0.267} \pm 0.002$     & $0.067 \pm 0.002$  & 0.38  & 0.29  & 0.33  \\ 
                      & \multicolumn{2}{l|}{\bald{}}    & $0.249 \pm 0.004$     & $0.072 \pm 0.005$  & 0.32   & 0.31   & 0.37  \\ 
                      & \multicolumn{2}{l|}{\mcme{}}    & $0.261 \pm 0.004$    & $\textbf{0.093} \pm 0.005$  & 0.32   & 0.34   & 0.34  \\ \hline \hline

                      & \multicolumn{2}{l|}{\random{}}  & $\textbf{0.324} \pm 0.005$     & $0.208 \pm 0.012$  & 0.47  & 0.38  & 0.15 \\ 
    \textbf{\wanli{}}     & \multicolumn{2}{l|}{\dal{}}    & $0.321 \pm 0.003$     & $0.223 \pm 0.008$  & 0.49  & 0.38  & 0.13  \\ 
                      & \multicolumn{2}{l|}{\bald{}}    & $0.317 \pm 0.001$     & $0.213 \pm 0.005$  & 0.40   & 0.39 & 0.21  \\ 
                      & \multicolumn{2}{l|}{\mcme{}}    & $0.318 \pm 0.002$    & $\textbf{0.229} \pm 0.009$  & 0.38   & 0.35  & 0.27  \\ \hline \hline

                      & \multicolumn{2}{l|}{\random{}}  & $\textbf{0.492} \pm 0.003$     & $0.082 \pm 0.004$  & 0.43  & 0.32  & 0.26 \\ 
    \textbf{\anli{}}     & \multicolumn{2}{l|}{\dal{}}    & $0.467 \pm 0.008$     & $0.084 \pm 0.004$  & 0.47  & 0.3  & 0.23  \\ 
                      & \multicolumn{2}{l|}{\bald{}}    & $0.478 \pm 0.007$     & $0.094 \pm 0.003$  & 0.39   & 0.30   & 0.31  \\ 
                      & \multicolumn{2}{l|}{\mcme{}}    & $0.491 \pm 0.007$    & $\textbf{0.110} \pm 0.004$  & 0.37   & 0.31   & 0.31  \\ \hline \hline

                            & \multicolumn{2}{l|}{\random{}}  & $0.276 \pm 0.003$              & $0.405 \pm 0.012$  & 0.41  & 0.34  & 0.25 \\ 
    \textbf{Multi}  & \multicolumn{2}{l|}{\dal{}}     & $0.276 \pm 0.004$              & $0.463 \pm 0.018$  & 0.47  & 0.34  & 0.19  \\ 
                            & \multicolumn{2}{l|}{\bald{}}    & $0.220 \pm 0.012$              & $0.445 \pm 0.025$  & 0.34   & 0.32   & 0.35 \\ 
                            & \multicolumn{2}{l|}{\mcme{}}    & $\textbf{0.323} \pm 0.004$     & $\textbf{0.556} \pm 0.011$  & 0.4   & 0.3   & 0.3  \\ \hline
\end{tabular}
\caption{Profiling strategy acquisitions for \textbf{single-source} and \textbf{multi-source} active learning in terms of mean input diversity (\textbf{I-Div.}),  mean uncertainty (\textbf{Unc.}) and class distributions (\textbf{N}: \textit{neutral}, \textbf{E}: \textit{entailment}, \textbf{C}: \textit{contradiction}) across seeds, following a \textit{mean} $\pm$ \textit{std} cell scheme. Each strategy is profiled \textit{after} all rounds of active learning: $|\mathcal{D}_{train}| = 4K$.}
\label{table:single_source_basic_metrics_full}
\end{table*}

\begin{figure}[t!]
\centering
\includegraphics[width=\linewidth]{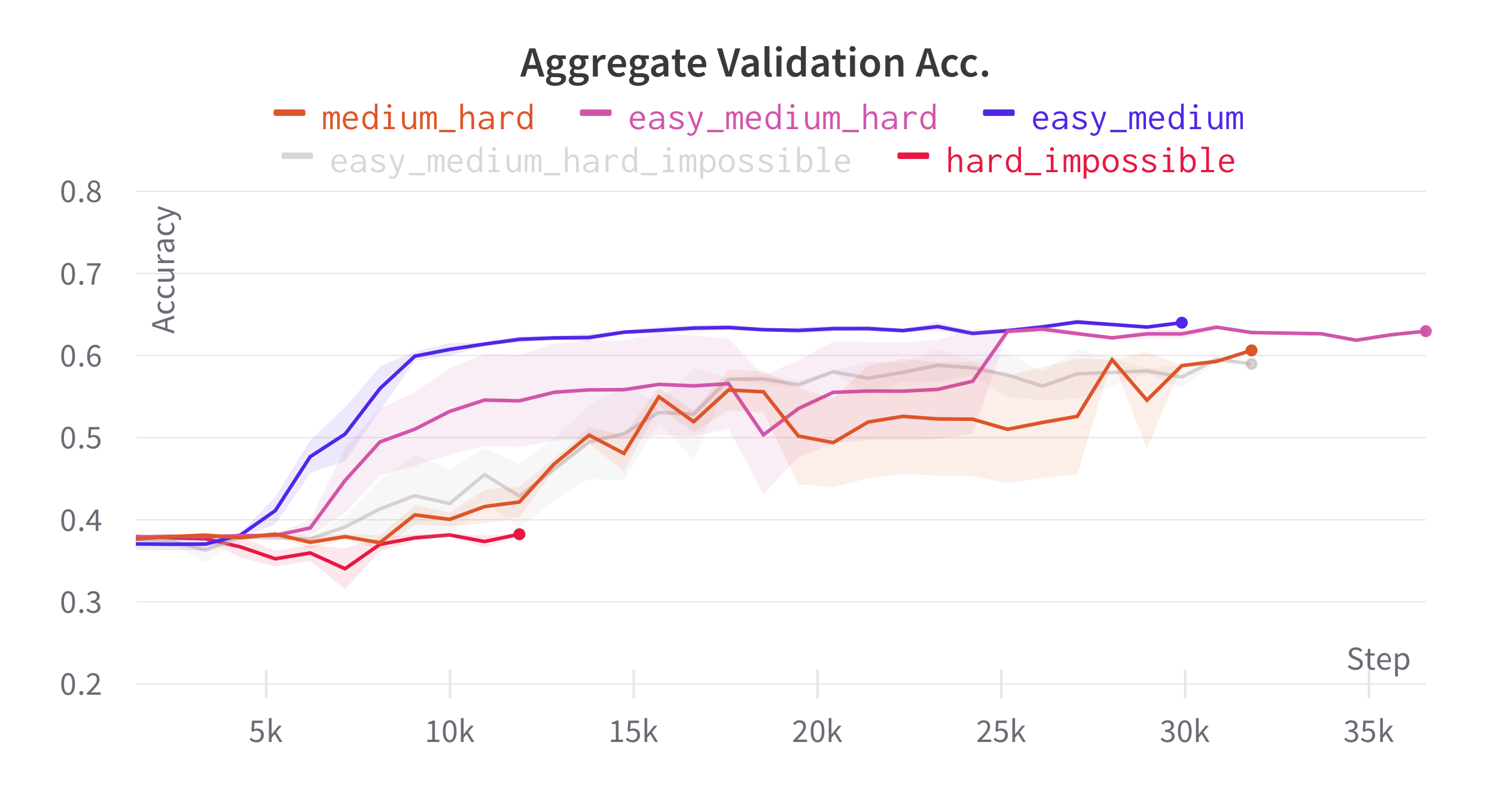}
\caption{Aggregate validation accuracy for models trained on data of various difficulties. Training data difficulty appears to affect convergence speed: models trained on easier training sets converge earlier, whilst inclusion of hard and impossible models requires more steps to converge. In our experiments, models trained only on hard/impossible examples never surpass at-chance performance and suffer from model failure more often.}
\label{fig:agg_difficulty_splits}
\end{figure}

\begin{figure*}[t!]
\minipage{0.33\textwidth}
\includegraphics[width=\linewidth]{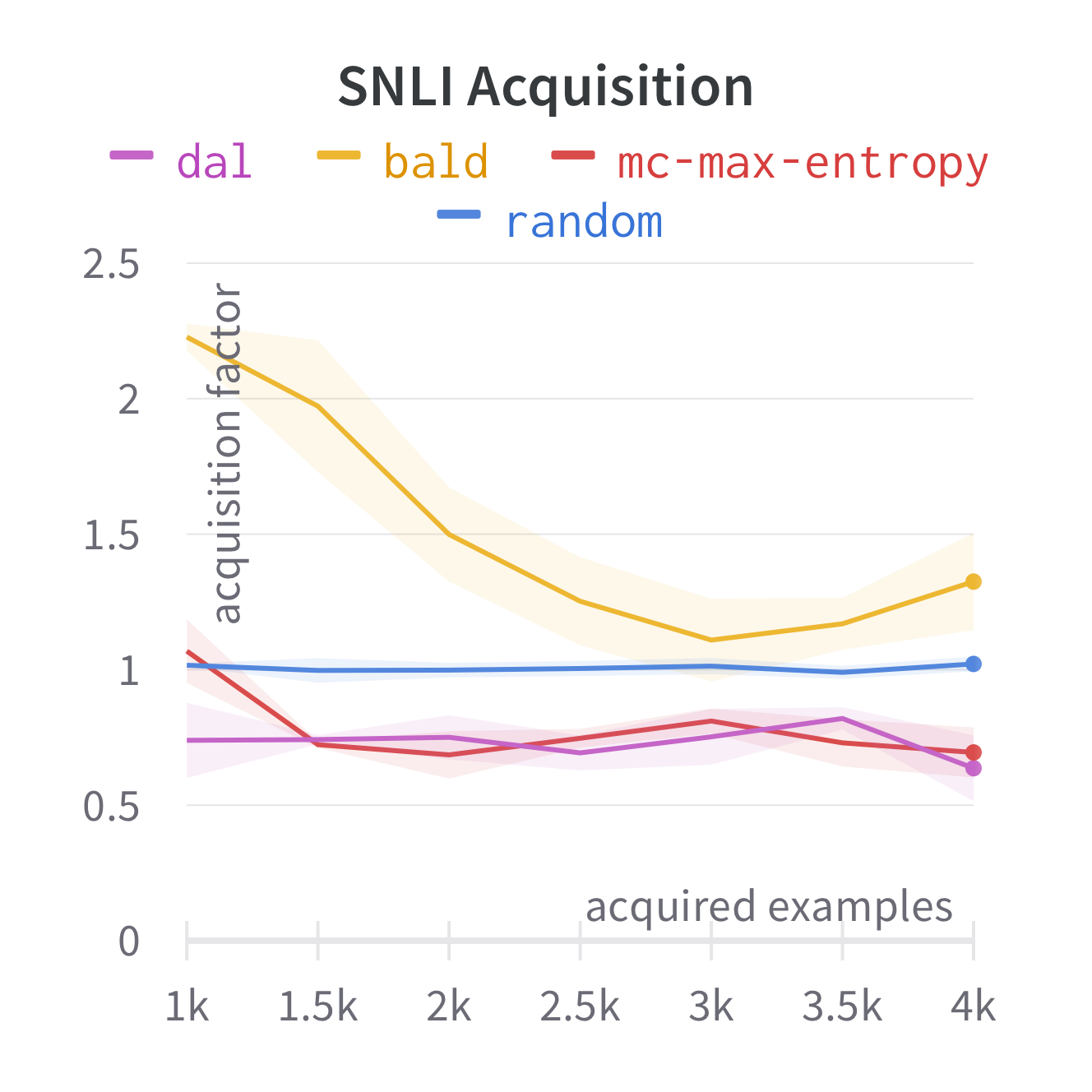}
\subcaption{SNLI Acquisition factors}
\endminipage
\minipage{0.33\textwidth}
\includegraphics[width=\linewidth]{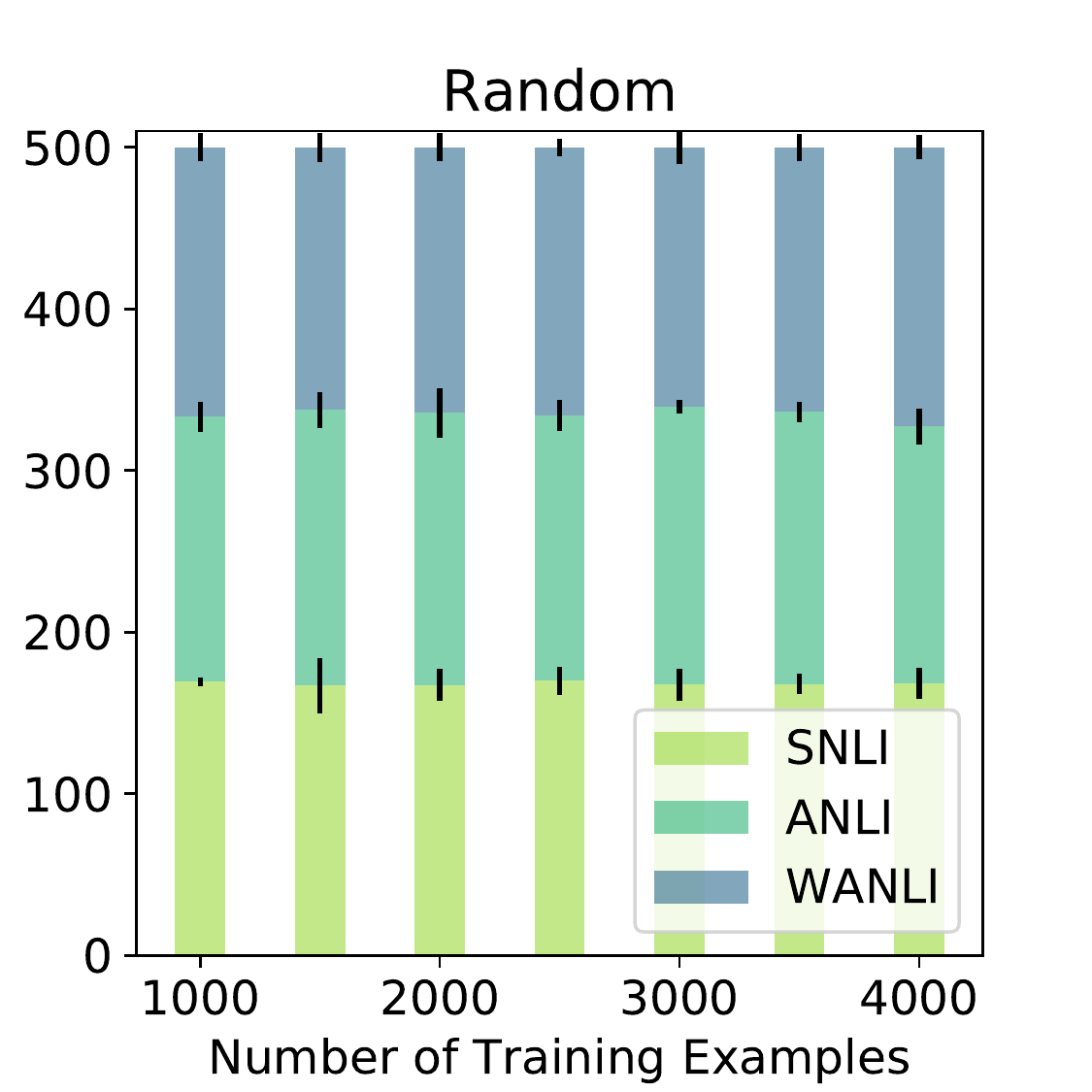}
\subcaption{Acquisitions by source}
\endminipage
\minipage{0.33\textwidth}
\includegraphics[width=\linewidth]{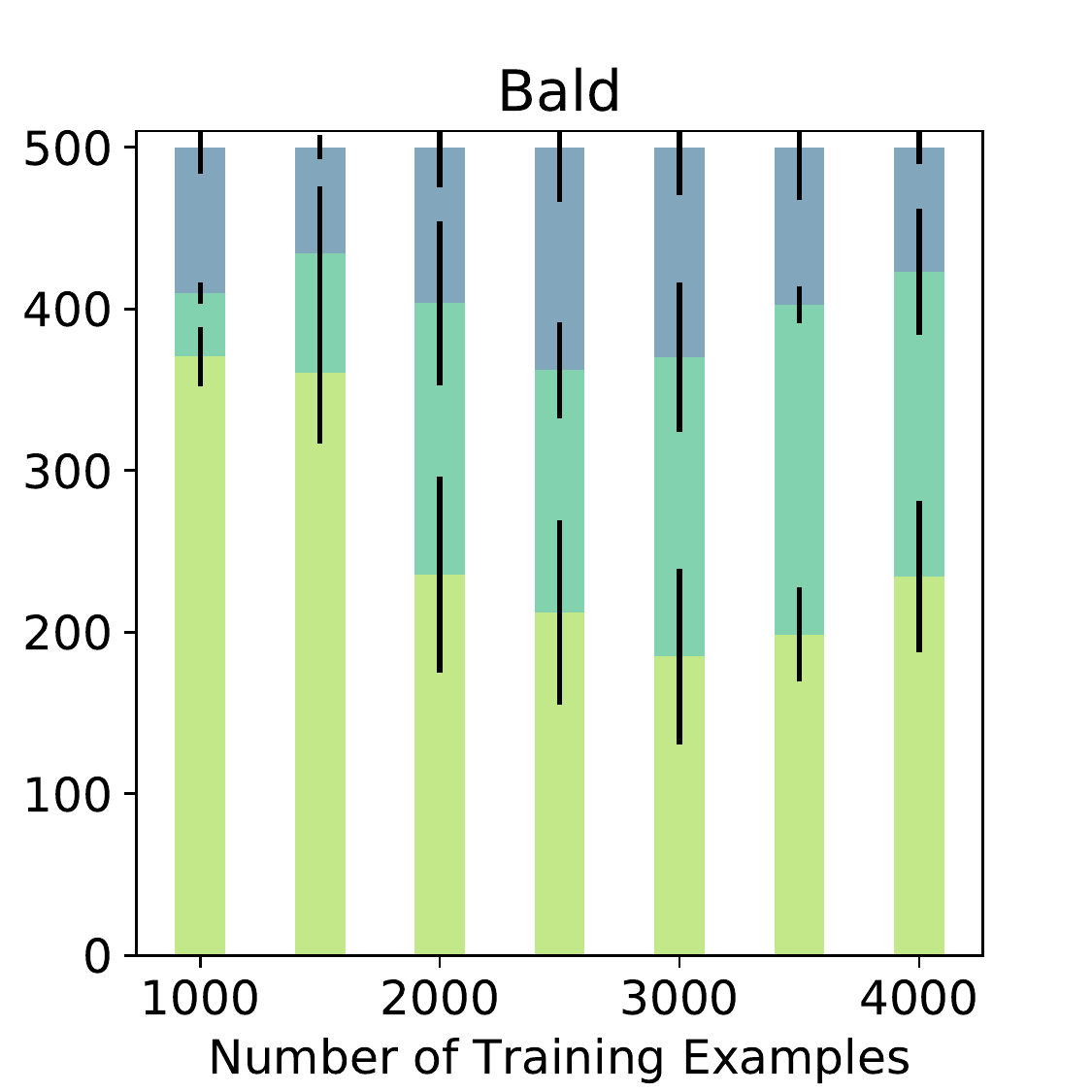}
\subcaption{Acquisitions by source}
\endminipage
\caption{Acquisition factor explained. By design, random selection acquires a random sample of the unlabelled pool. Since the pool is initialized with even amounts of each source, random consistently acquires a uniform sample over sources, as can be seen in \textbf{(b)}. Consequently, the SNLI acquisition factor for Random steadily hovers around 1 in \textbf{(a)}. Conversely, BALD initially acquires mostly SNLI examples, as reflected by its high initial acquisition factor. This corresponds to its absolute acquisition statistics in \textbf{(c)}, where we note that initially BALD acquires predominantly SNLI.}
\label{fig:acq_exp}
\end{figure*}

\paragraph{Acquisition Factor}\label{section:acq_exp}
For the multi-source experiments we also plot the \textit{acquisition factors}: a source-strategy-specific statistic which indicates how much data of a given source is acquired by some strategy, normalized by the share of that source in the unlabelled pool at the time of that acquisition round. This statistic is useful to interpret how much a strategy tends to acquisition of one source relative to others.
\\ \\
We normalize to correct for the effect that acquiring e.g.\ mostly SNLI data in an early round causes it to take up a relatively smaller share in the unlabelled pool in the next round, and by the simple consequence of there being fewer SNLI examples, they may have a lower likelihood to be acquired in those future rounds compared to examples from other sources.. As this could distort our impressions of the extent to which strategies acquire examples from different sources, we ideally want to correct for this.
\\ \\
We thus compute the acquisition factor for a given round as (1) the amount of source examples that were \textit{actually} acquired by the strategy for that round, \textit{divided} by (2) the amount of source examples that \textit{would} be acquired under \textit{random} sampling from the unlabelled pool at that point. For instance, if BALD has an acquisition factor of $ > 1 $ for SNLI, it means that it acquired more SNLI examples from the unlabeled pool than it \textit{would} have under random sampling. Conversely, the Random sampling baseline will always have consistent acquisition factors of 1, since the quantities in the numerator and denominator will be approximately the same: random selection always acquire as much from a source as it would under random selection. See Figure \ref{fig:acq_exp} for a graphical explanation.\footnote{Please note that these figures solely serve to support the textual explanation and should not be regarded as results otherwise.}
\\ \\

\begin{figure*}[t!]
\centering
\minipage{0.33\textwidth}
\includegraphics[width=\linewidth]{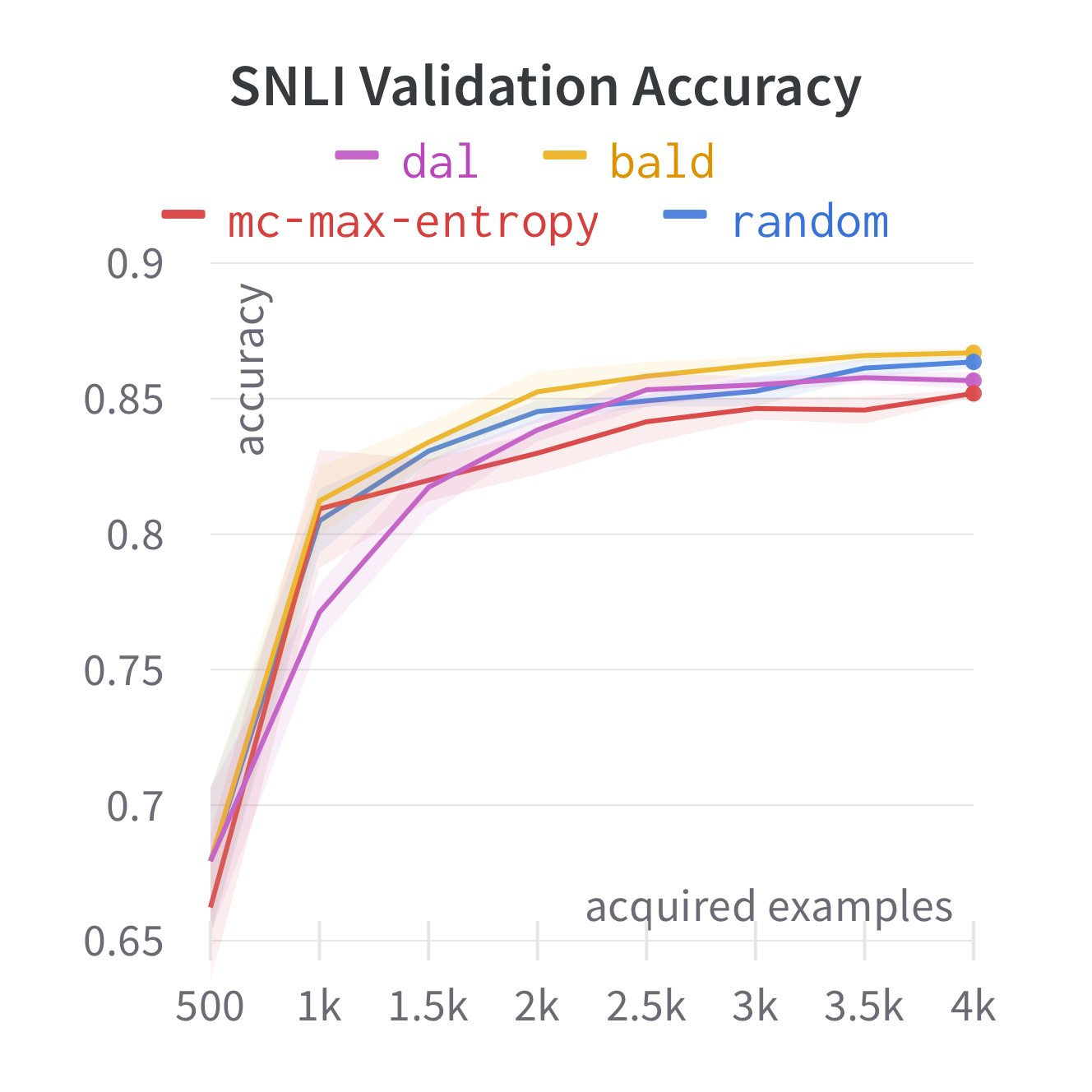}
\subcaption{SNLI Val. Acc.}
\endminipage
\minipage{0.33\textwidth}
\includegraphics[width=\linewidth]{src/imgs/new_main_exp/snli_acq.png}
\subcaption{SNLI Acquisition}
\endminipage
\minipage{0.33\textwidth}
\includegraphics[width=\linewidth]{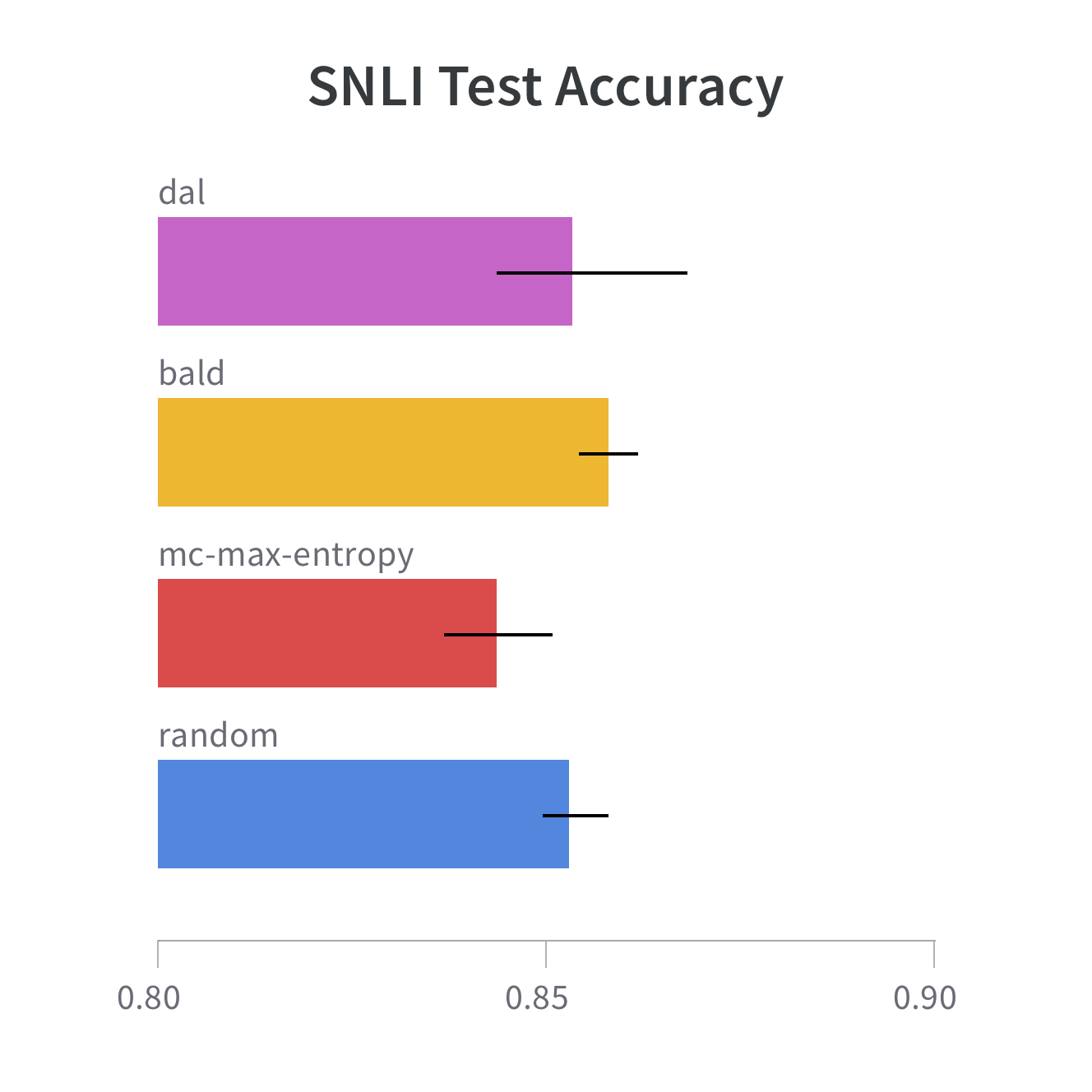}
\subcaption{SNLI Test Acc.}
\endminipage
\vfill
\minipage{0.33\textwidth}
\includegraphics[width=\linewidth]{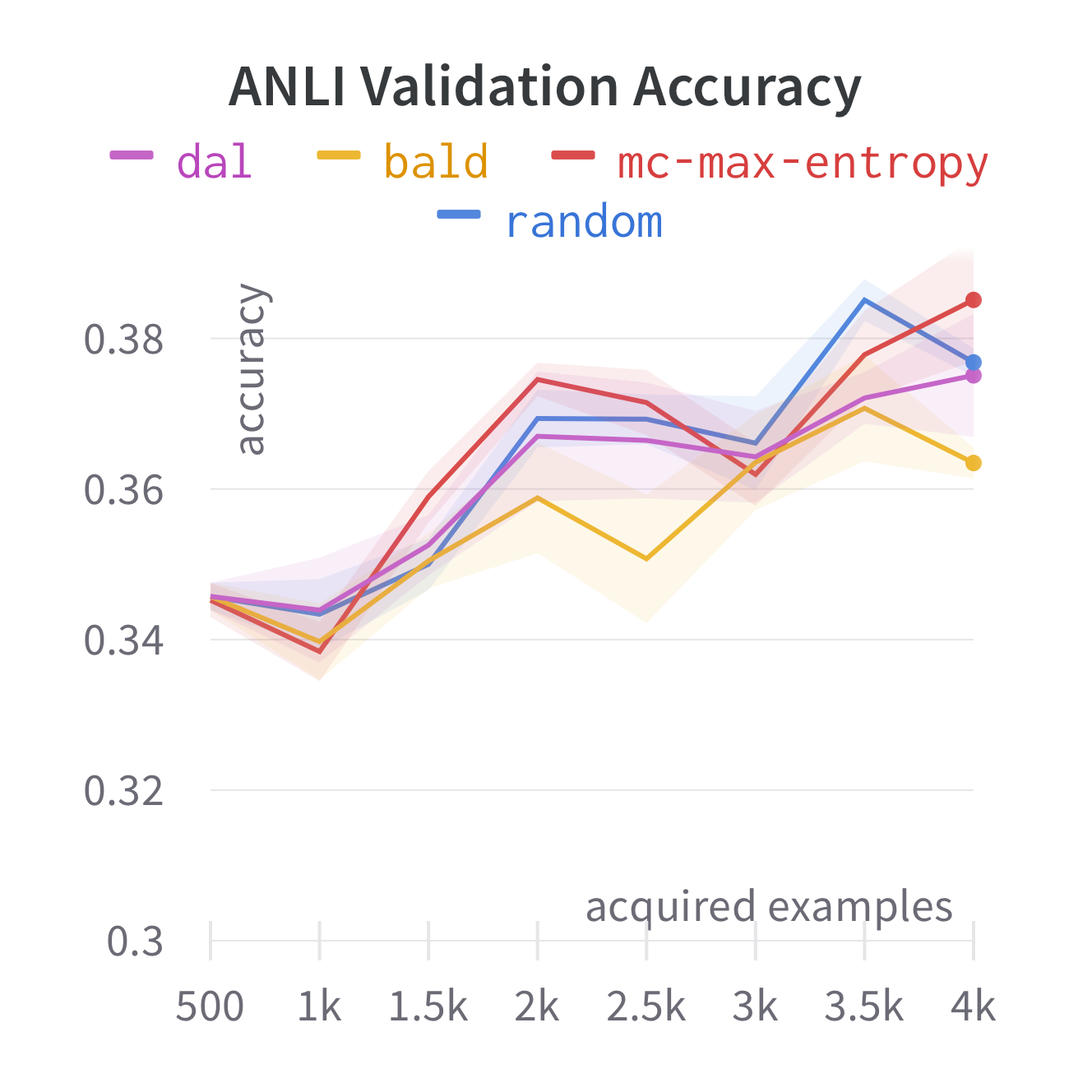}
\subcaption{ANLI Val. Acc.}\
\endminipage
\minipage{0.33\textwidth}
\includegraphics[width=\linewidth]{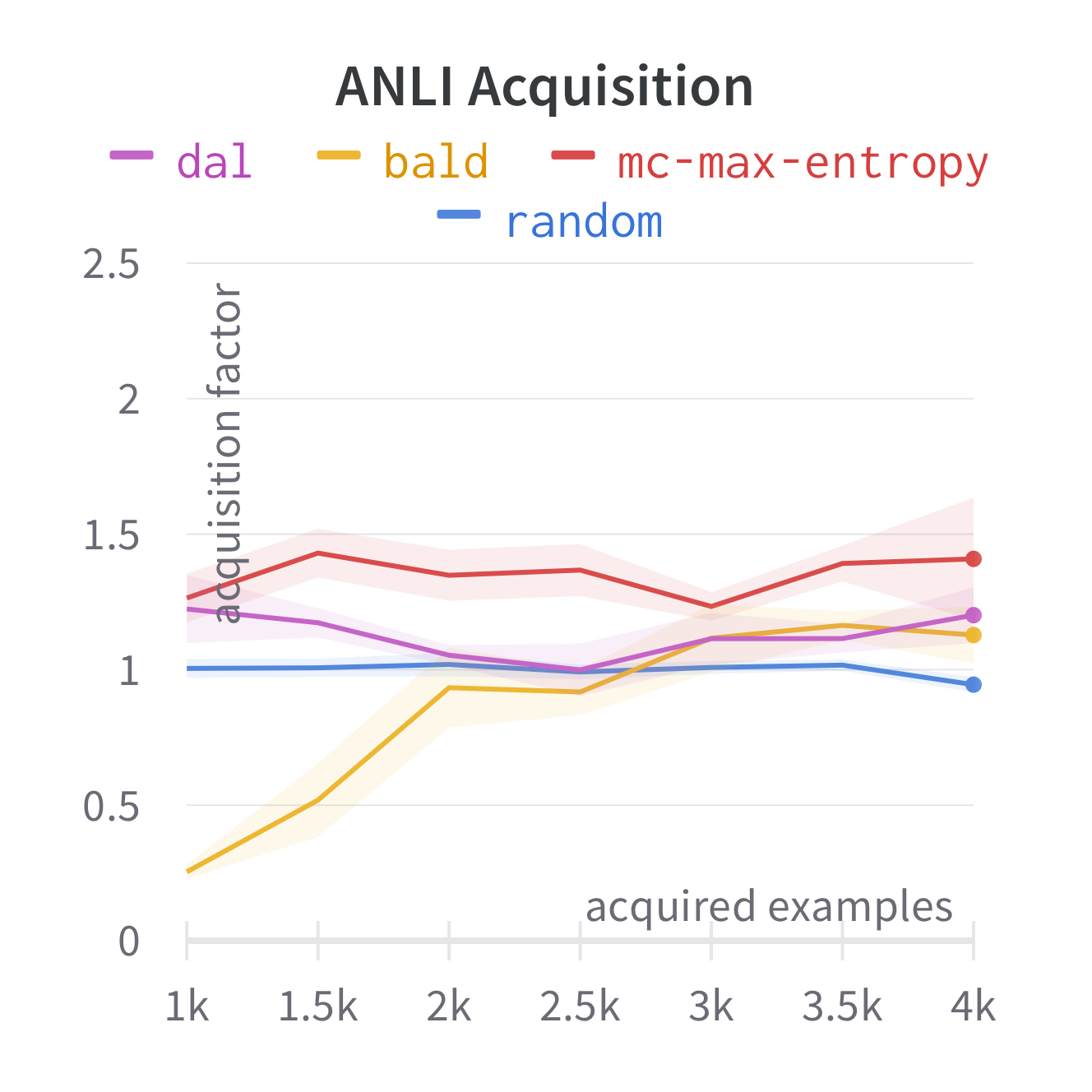}
\subcaption{ANLI Acquisition}
\endminipage
\minipage{0.33\textwidth}
\includegraphics[width=\linewidth]{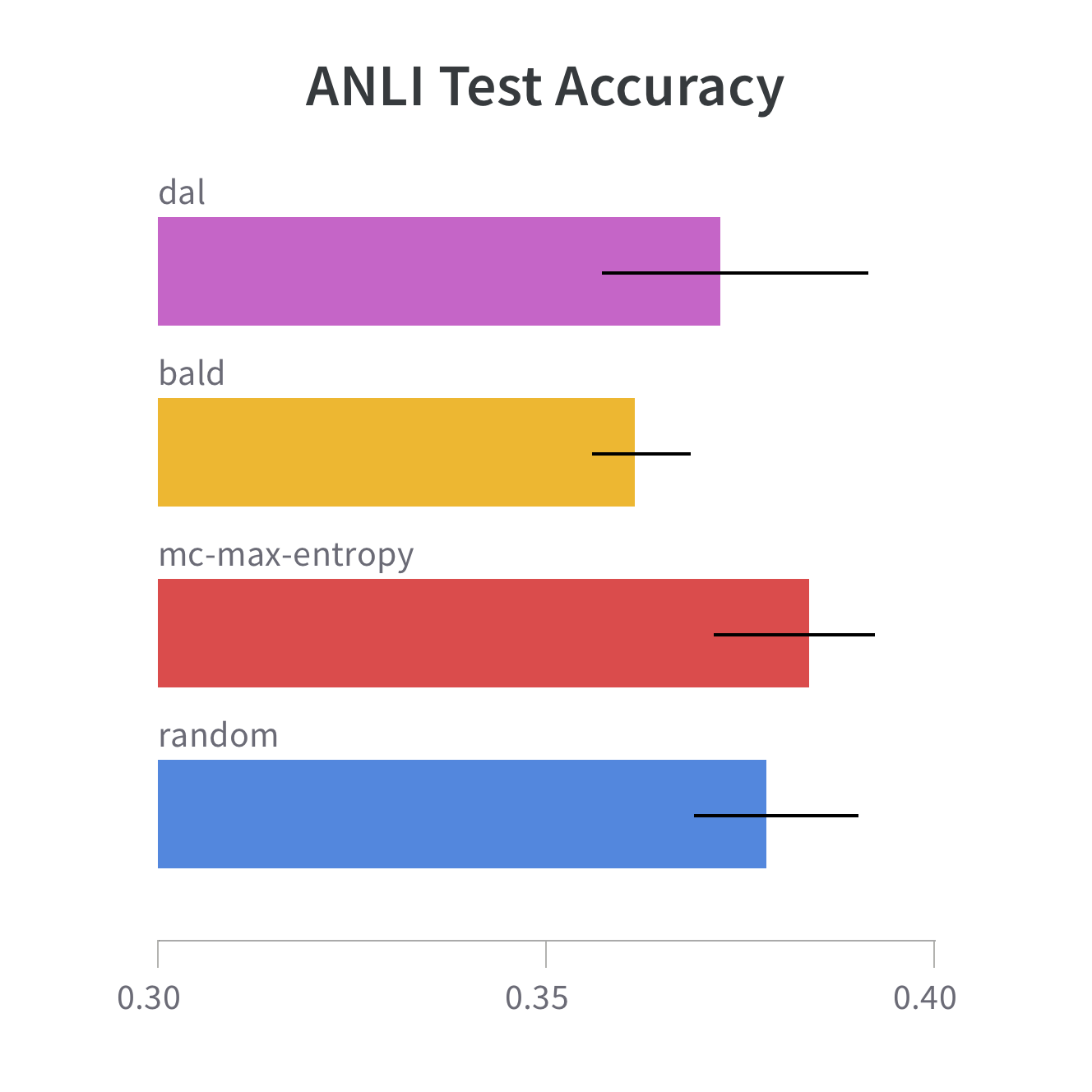}
\subcaption{ANLI Test Acc.}
\endminipage
\vfill
\minipage{0.33\textwidth}
\includegraphics[width=\linewidth]{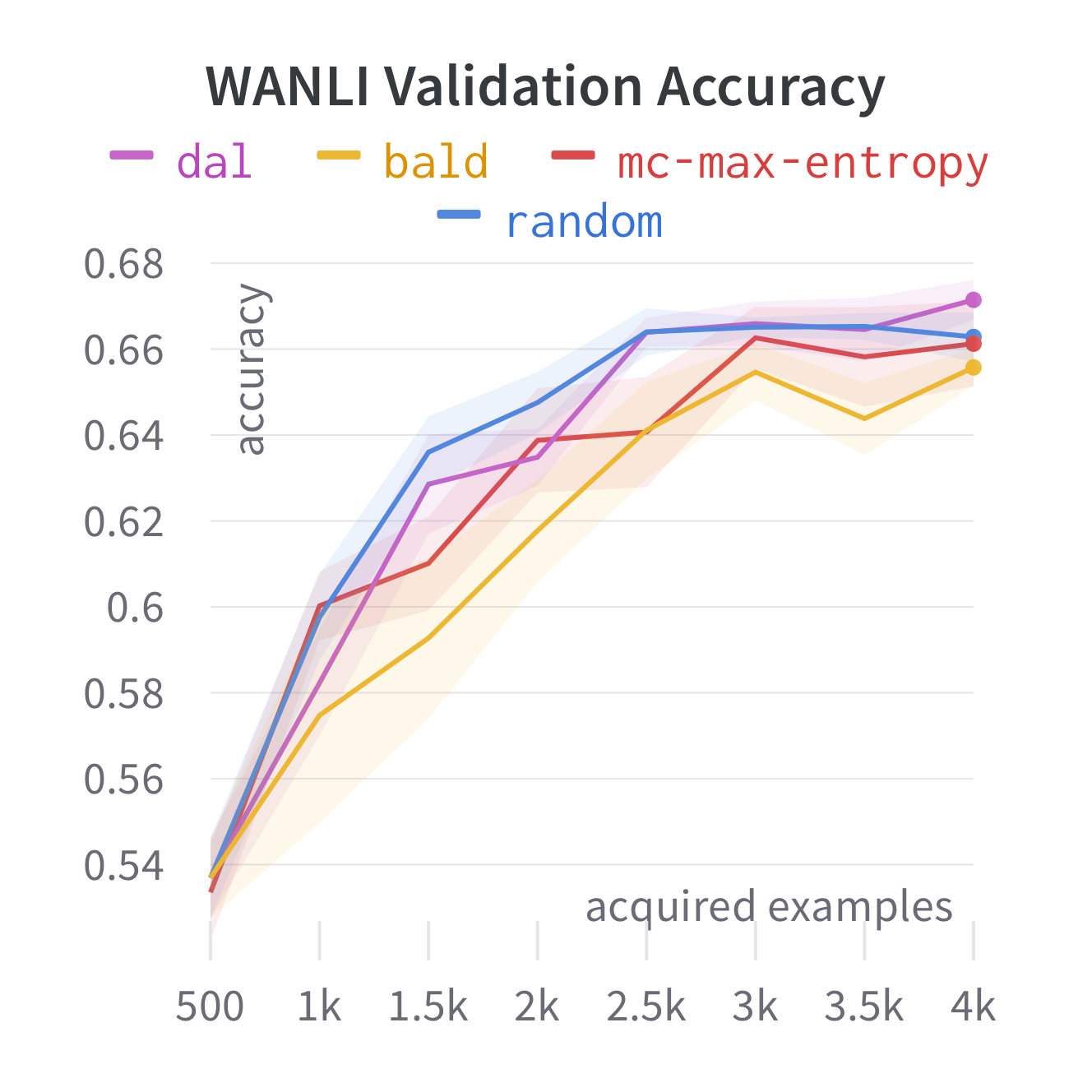}
\subcaption{WANLI Val. Acc.}
\endminipage
\minipage{0.33\textwidth}
\includegraphics[width=\linewidth]{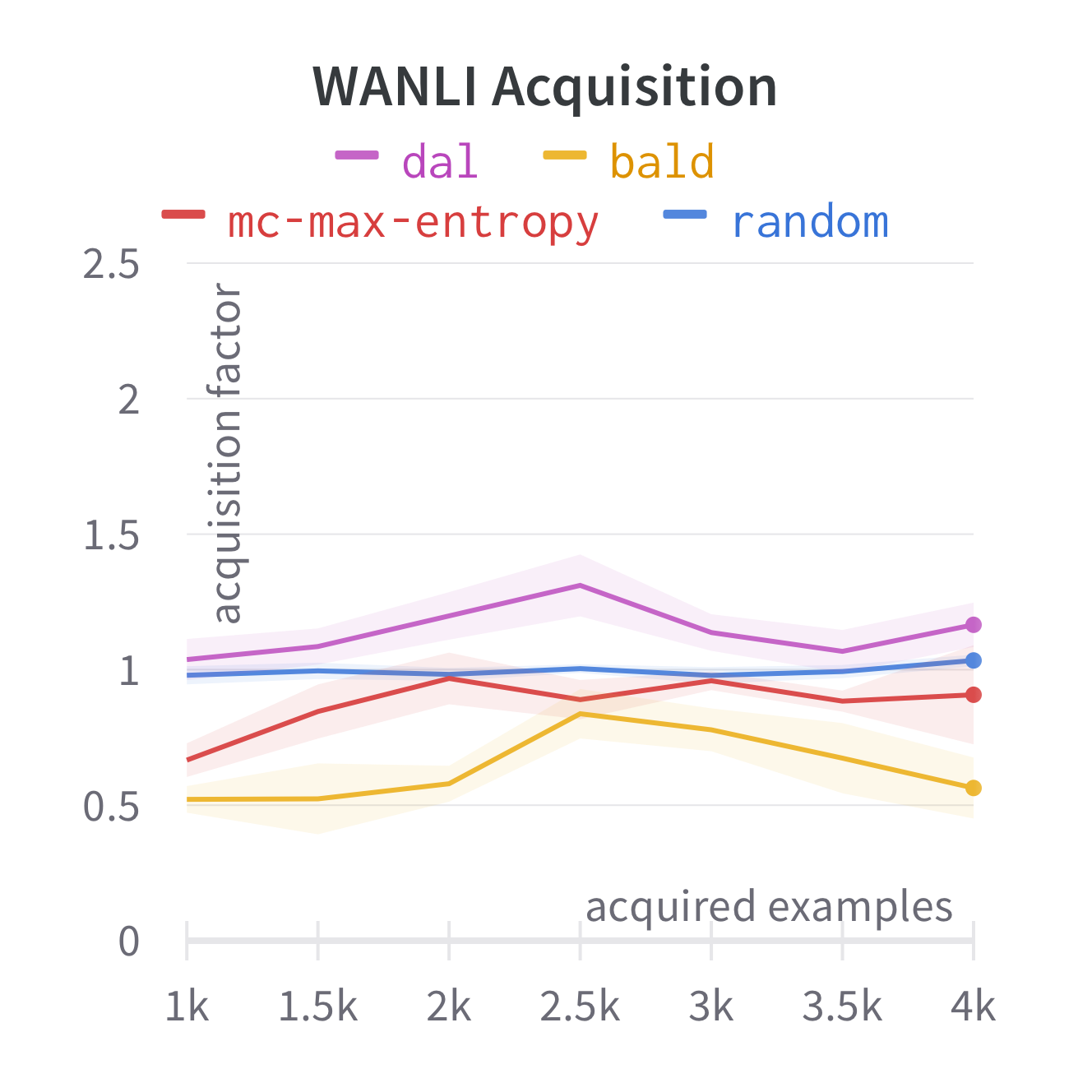}
\subcaption{WANLI Acquisition}
\endminipage
\minipage{0.33\textwidth}
\includegraphics[width=\linewidth]{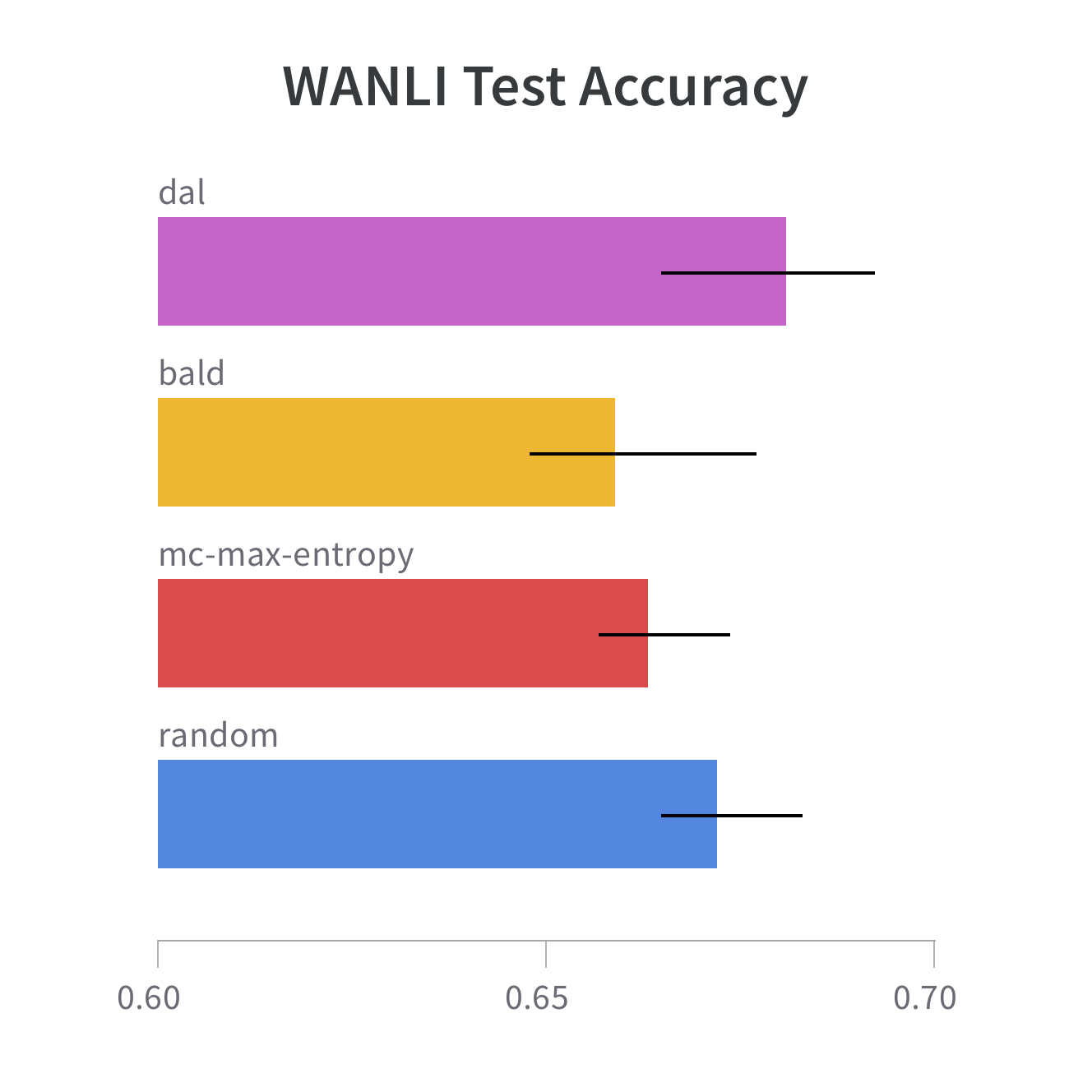}
\subcaption{WANLI Test Acc.}
\endminipage
\caption{In-domain multi-source Active Learning. The left-most graph represents the \textit{learning curve} per strategy, plotting the validation performance as more labelled data is acquired. The middle graph plots the \textit{acquisition factor} per source: how much a strategy acquires of a source, normalized by the share that source takes up in the wider data pool. Shaded regions indicate standard errors. Active learning fails to consistently beat random.}
\label{fig:main_ID}
\end{figure*}

\subsection{Dataset Cartography}\label{sec:datamaps_appendix}
As mentioned in Section \ref{section:cartography_method}, we train a RoBERTa-large model on a training set comprised of the entire unlabelled pool of training examples, 
i.e.\ $60$K examples in total.\footnote{The AL experiments are simulations, where we have the ground truth labels for the data of the pool, but we consider them unlabelled. So in this case, we use the original labelled dataset to train the cartography model.} Every $\frac{1}{2}$ epoch we perform inference on the full training set to get per-example confidence statistics, where the prediction logit corresponding to the gold-truth label serves as a proxy for model confidence. Variability is computed as the standard deviation over the set of confidence measurements. We stop training after $3$ epochs or after no improvement in validation accuracy on the aggregate of validation sets was observed. Following \citet{Karamcheti2021}, we classify examples along four difficulties via a threshold on the mean confidence value $p$.

\paragraph{Note on model discrepancy} When studying the strategy maps, it is instructive to note that there exists a discrepancy between models which may distort the truthfulness of learnability measurements. That is, the cartography model that was used to obtain confidence/variability measurements for datamap construction was trained on the entire pool of 60K examples, whereas the model used during AL is exposed to at most 4K examples after completing all rounds of acquisition. Consequently, examples which were found to be easy by the cartography model are likely to be substantially more difficult for the AL model. As datamaps are inherently model-based, the generated strategy maps should be interpreted as skewed estimates of the true learnability spectra. While we recognize the potential value of preserving scale equivalence when using dataset cartography for analysis, we choose to follow \citet{Karamcheti2021} and intend to employ datamaps as a \textit{post-hoc} diagnostic tool. In other words, we are interested in examining example learnability in an \textit{absolute} sense (i.e.\ with respect to the entire pool) rather than a \textit{relative} sense (with respect to the acquired data). Here, we also note that a model trained on a larger set of examples will likely more accurately reflect the true difficulty of examples. This is important if we want to draw new insights on the learnability of NLI datasets in a wider sense.

\begin{table*}
\resizebox{\textwidth}{!}{\begin{tabular}{|c|ll|c|c|c|c|}
\hline
\multicolumn{1}{|l|}{Task} & \multicolumn{2}{l|}{Diff.}& \textbf{Random} & \textbf{MCME} & \textbf{BALD} & \textbf{DAL} \\ 
\hline \hline
                      & \multicolumn{2}{l|}{\textbf{E}}                 &  $95.4 \pm 0.9 \mid 95.0 \pm 0.4$              & $95.0 \pm 0.5 \mid 93.3 \pm 0.7$               & $95.5 \pm 1.0 \mid 94.8 \pm 0.3$            & $93.7 \pm 1.2 \mid 94.6 \pm 0.8$ \\ 
    \textbf{SNLI}     & \multicolumn{2}{l|}{\textbf{M}}                 &  $69.0 \pm 2.6 \mid 67.7 \pm 1.2$              & $70.2 \pm 1.6 \mid 68.0 \pm 1.2$               & $70.5 \pm 2.6 \mid 69.5 \pm 1.2$            & $67.8 \pm 2.4 \mid 68.2 \pm 2.5$ \\ 
                      & \multicolumn{2}{l|}{\textbf{H}}                 &  $46.3 \pm 3.5 \mid 44.6 \pm 1.9$              & $48.5 \pm 1.6 \mid 46.0 \pm 2.2$               & $47.1 \pm 3.2 \mid 49.2 \pm 2.9$            & $44.7 \pm 2.2 \mid 45.8 \pm 3.9$ \\ 
                      & \multicolumn{2}{l|}{\textbf{I}}                 &  $17.7 \pm 1.7 \mid 17.2 \pm 1.7$              & $20.3 \pm 2.4 \mid 22.1 \pm 0.9$               & $20.7 \pm 1.7 \mid 22.2 \pm 0.4$            & $18.4 \pm 2.0 \mid 17.5 \pm 2.2$ \\ \hline \hline
                       
                      & \multicolumn{2}{l|}{\textbf{E}}                 &  $81.4 \pm 4.9 \mid 81.7 \pm 4.3$              & $77.0 \pm 4.1 \mid 79.2 \pm 1.4$               & $82.1 \pm 3.1  \mid 73.8 \pm 2.1$           & $81.8 \pm 3.5 \mid 79.9 \pm 3.4$ \\ 
    \textbf{ANLI}     & \multicolumn{2}{l|}{\textbf{M}}                 &  $62.2 \pm 2.7 \mid 62.1 \pm 0.4$              & $61.7 \pm 3.5 \mid 57.8 \pm 0.9$               & $61.1 \pm 1.6  \mid 57.1 \pm 4.0$           & $63.1 \pm 2.3 \mid 59.8 \pm 3.1$ \\ 
                      & \multicolumn{2}{l|}{\textbf{H}}                 &  $39.3 \pm 3.3 \mid 40.2 \pm 0.7$              & $42.9 \pm 3.2 \mid 41.6 \pm 2.4$               & $39.2 \pm 2.6  \mid 39.7 \pm 2.6$           & $39.0 \pm 2.0 \mid 38.5 \pm 1.5$ \\ 
                      & \multicolumn{2}{l|}{\textbf{I}}                 &  $14.4 \pm 2.4 \mid 14.2 \pm 2.2$              & $15.7 \pm 2.0 \mid 17.0 \pm 0.8$               & $12.6 \pm 1.2  \mid 14.5 \pm 1.8$           & $14.2 \pm 1.7 \mid 15.0 \pm 1.9$ \\ \hline \hline
                       
                      & \multicolumn{2}{l|}{\textbf{E}}                 &  $94.3 \pm 2.0 \mid 92.9 \pm 1.4$              & $92.7 \pm 0.8 \mid 91.2 \pm 0.7$               & $95.2 \pm 0.9  \mid 90.9 \pm 1.1$           & $94.1 \pm 0.4 \mid 93.1 \pm 1.3$ \\ 
    \textbf{WANLI}    & \multicolumn{2}{l|}{\textbf{M}}                 &  $72.7 \pm 3.2 \mid 69.6 \pm 3.5$              & $71.4 \pm 1.2 \mid 69.9 \pm 2.2$               & $72.7 \pm 1.7  \mid 67.9 \pm 3.0$           & $72.3 \pm 3.1 \mid 73.8 \pm 2.1$ \\ 
                      & \multicolumn{2}{l|}{\textbf{H}}                 &  $41.9 \pm 2.7 \mid 40.4 \pm 1.6$              & $43.1 \pm 2.6 \mid 44.2 \pm 1.2$               & $42.4 \pm 4.6  \mid 40.7 \pm 2.6$           & $40.6 \pm 1.9 \mid 41.0 \pm 3.4$ \\ 
                      & \multicolumn{2}{l|}{\textbf{I}}                 &  $12.7 \pm 1.5 \mid 14.4 \pm 1.4$              & $14.0 \pm 2.0 \mid 16.6 \pm 4.7$               & $11.9 \pm 2.6  \mid 14.7 \pm 4.4$           & $13.9 \pm 2.8 \mid 14.6 \pm 3.6$ \\ \hline

    \hline
    \hline

                          & \multicolumn{2}{l|}{\textbf{E}}                 &  $93.3 \pm 2.0 \mid 93.6 \pm 0.6$              & $94.3 \pm 0.9 \mid 92.8 \pm 1.2$               & $94.4 \pm 0.8  \mid 92.4 \pm 0.8$           & $93.4 \pm 1.3 \mid 93.8 \pm 1.5$ \\ 
    \textbf{MNLI}     & \multicolumn{2}{l|}{\textbf{M}}                 &  $72.6 \pm 3.8 \mid 72.9 \pm 1.5$              & $78.6 \pm 2.0 \mid 74.7 \pm 4.6$               & $75.9 \pm 1.6  \mid 72.8 \pm 4.4$           & $74.7 \pm 3.4 \mid 78.1 \pm 4.6$ \\ 
                      & \multicolumn{2}{l|}{\textbf{H}}                 &  $41.8 \pm 3.8 \mid 44.5 \pm 1.0$              & $51.9 \pm 4.9 \mid 49.3 \pm 3.3$               & $45.7 \pm 4.1  \mid 46.7 \pm 3.3$           & $46.1 \pm 3.9 \mid 48.3 \pm 4.7$ \\ 
                      & \multicolumn{2}{l|}{\textbf{I}}                 &  $12.8 \pm 3.8 \mid 15.7 \pm 1.8$              & $20.9 \pm 2.9 \mid 19.5 \pm 1.4$               & $13.1 \pm 2.0  \mid 16.9 \pm 1.9$           & $12.6 \pm 2.2 \mid 18.5 \pm 6.3$ \\ \hline

    \hline \hline
                      
                      & \multicolumn{2}{l|}{\textbf{E}}                 &  $91.0 \pm 6.7 \mid 90.8 \pm 5.8$              & $89.8 \pm 7.7 \mid 89.1 \pm 5.9$               &  $91.8 \pm 5.9 \mid 88.0 \pm 8.4$           & $90.7 \pm 5.5 \mid 90.4 \pm 6.4$ \\ 
    \textbf{All}      & \multicolumn{2}{l|}{\textbf{M}}                 &  $69.3 \pm 5.3 \mid 68.1 \pm 4.4$              & $70.5 \pm 6.4 \mid 67.6 \pm 6.7$               &  $70.0 \pm 5.8 \mid 66.8 \pm 6.8$           & $69.4 \pm 5.2 \mid 69.9 \pm 7.6$ \\ 
                      & \multicolumn{2}{l|}{\textbf{H}}                 &  $42.3 \pm 4.2 \mid 42.4 \pm 2.5$              & $46.5 \pm 5.1 \mid 45.3 \pm 3.7$               &  $43.6 \pm 4.8 \mid 44.1 \pm 4.9$           & $42.6 \pm 3.9 \mid 43.4 \pm 5.3$ \\ 
                      & \multicolumn{2}{l|}{\textbf{I}}                 &  $14.4 \pm 3.3 \mid 15.3 \pm 2.2$              & $17.7 \pm 3.8 \mid 18.8 \pm 3.4$               &  $14.6 \pm 4.1 \mid 17.1 \pm 4.1$           & $14.8 \pm 3.2 \mid 16.4 \pm 4.2$ \\ \hline
                       
\end{tabular}}
\caption{\textbf{Multi-source} in-domain strategy comparison for difficulty-stratified test outcomes between original training set and training set with hard-to-learn examples excluded. Values denote test accuracies and standard deviations. We consider performance on test examples belonging to the Easy (\textbf{E}), Medium (\textbf{M}), Hard (\textbf{H}) and Impossible (\textbf{I}) sets. \\Cell scheme to be read as \textit{ablated} $\mid$ \textit{original}.}
\label{tab:in_domain_stratified_test}
\end{table*}


\newpage


\begin{figure*}\centering
\includegraphics[width=1\linewidth]{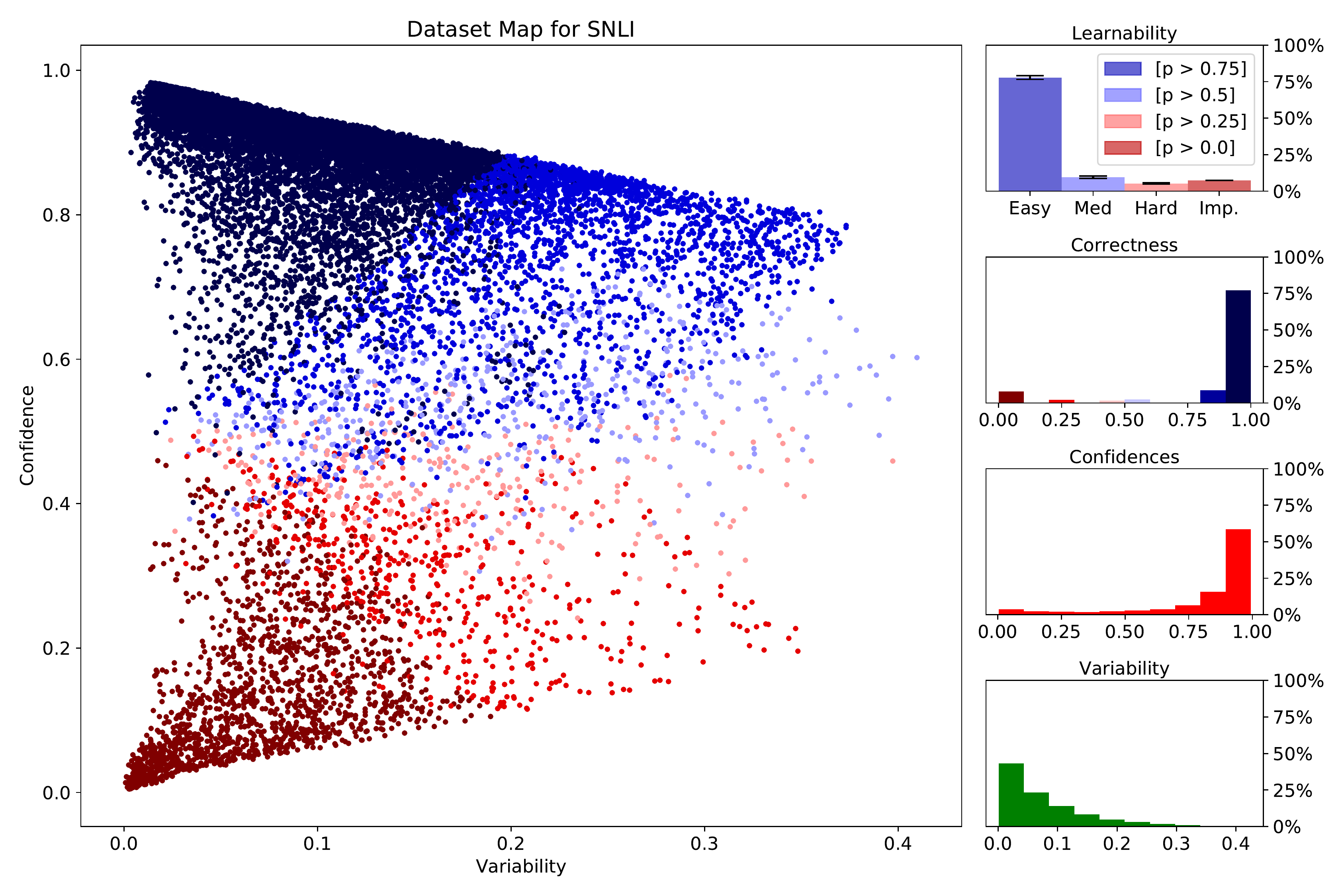}
\vfill
\includegraphics[width=1\linewidth]{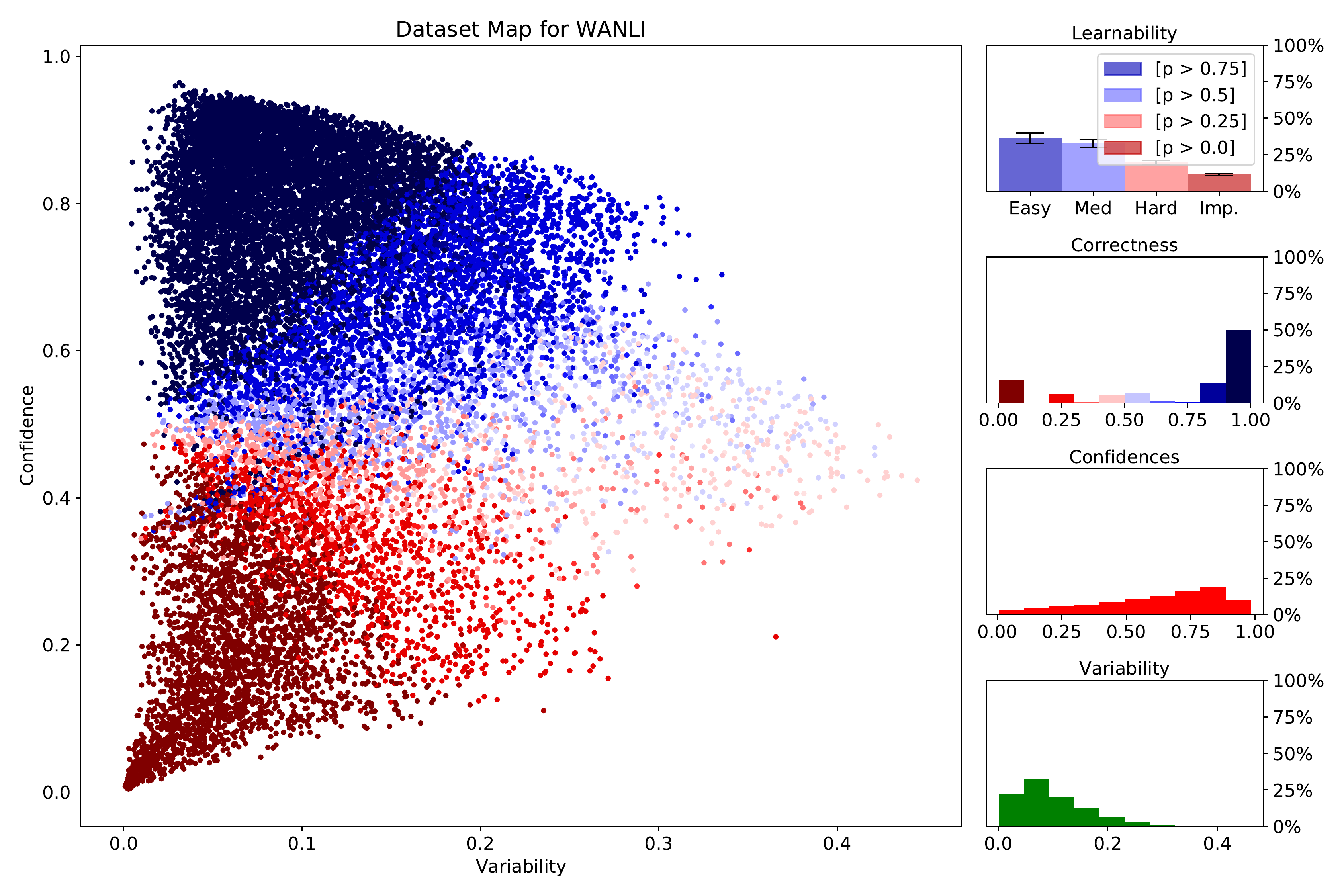}
\caption{Dataset Maps for SNLI and WANLI training data. The majority of SNLI examples lie in the easy-to-learn region. For WANLI, datapoints are more evenly distributed across the learnability spectrum and exhibit greater variability.}
\label{fig:cartography_sources}
\end{figure*}

\begin{figure*}\centering
\includegraphics[width=1\linewidth]{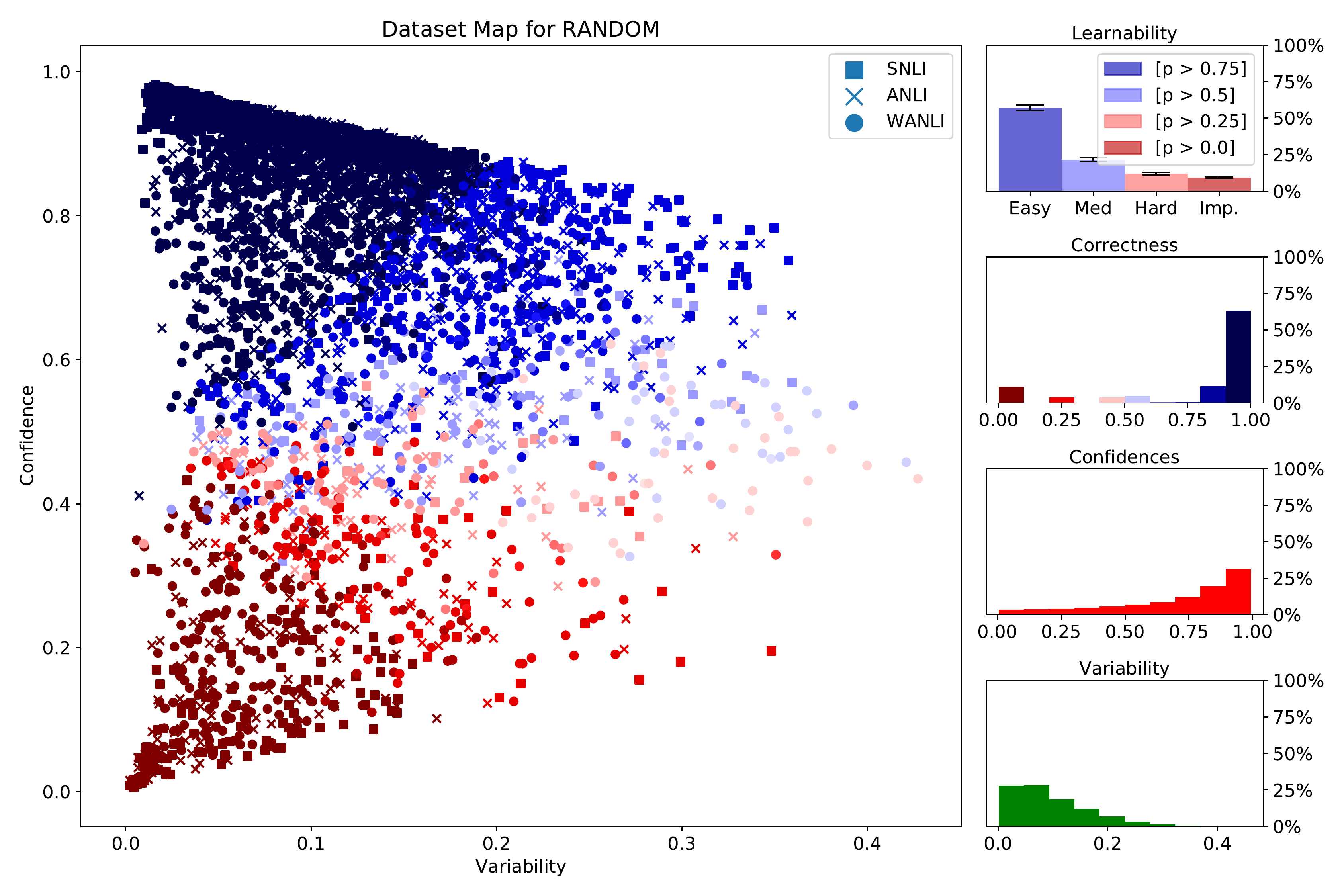}
\vfill
\includegraphics[width=1\linewidth]{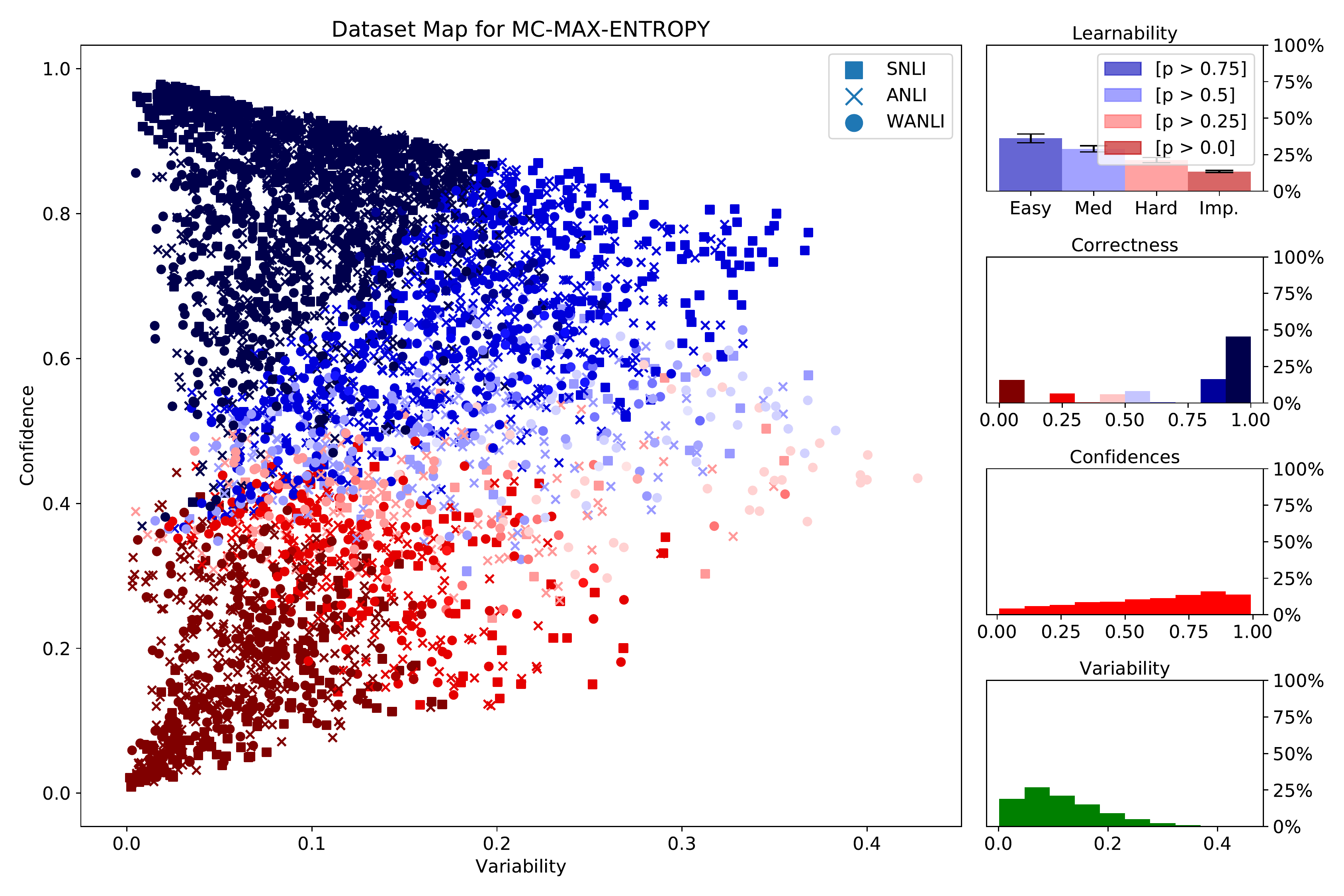}

\caption{Strategy maps for the Random and MC-Max-entropy strategies for \textbf{multi-source} active learning. The latter acquires considerably more hard-to-learn instances.}
\label{fig:cartography_strategies}
\end{figure*}


\newpage
\begin{figure*}\centering
\minipage{0.33\textwidth}
\includegraphics[width=\linewidth]{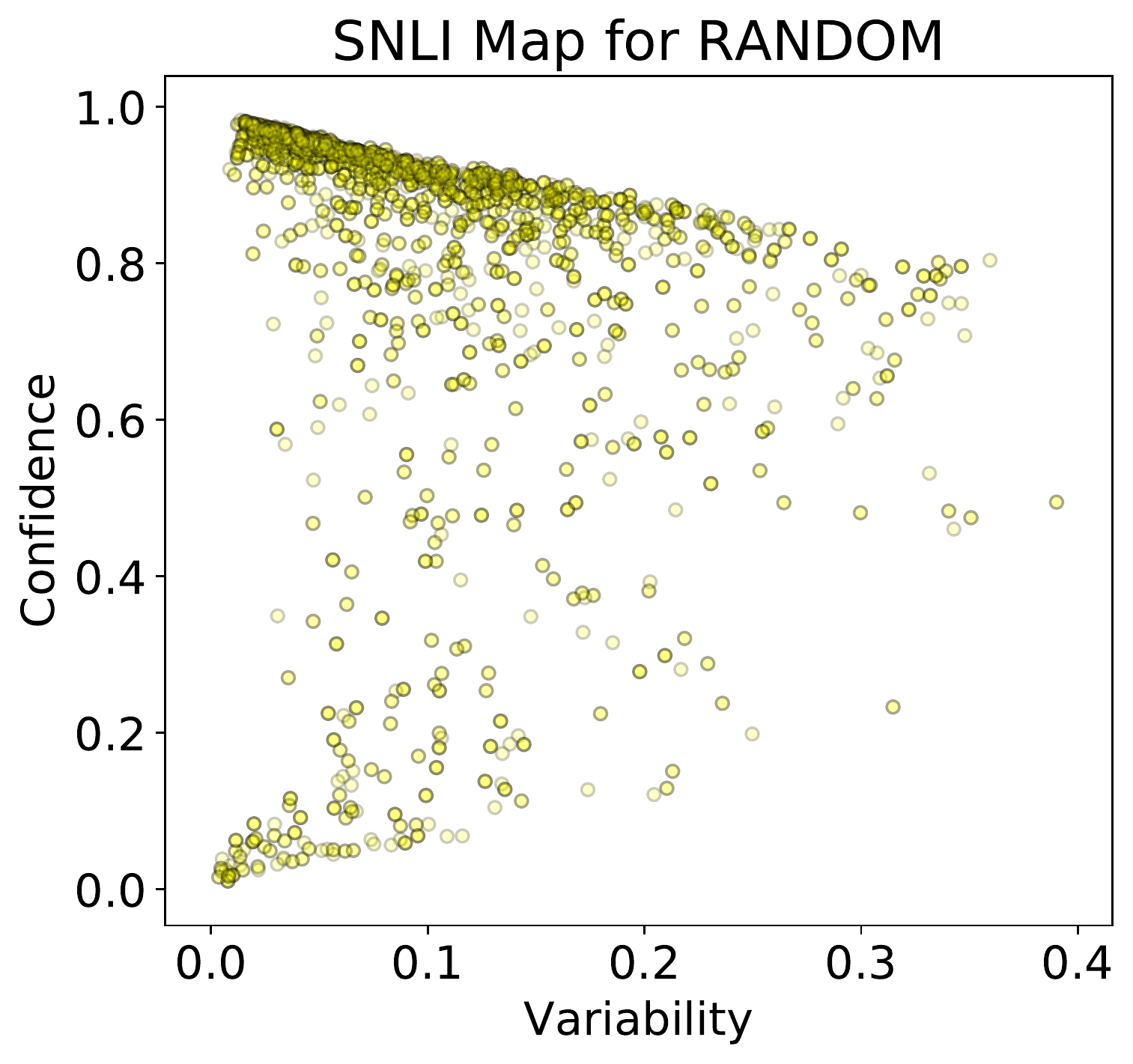}
\endminipage
\minipage{0.33\textwidth}
\includegraphics[width=\linewidth]{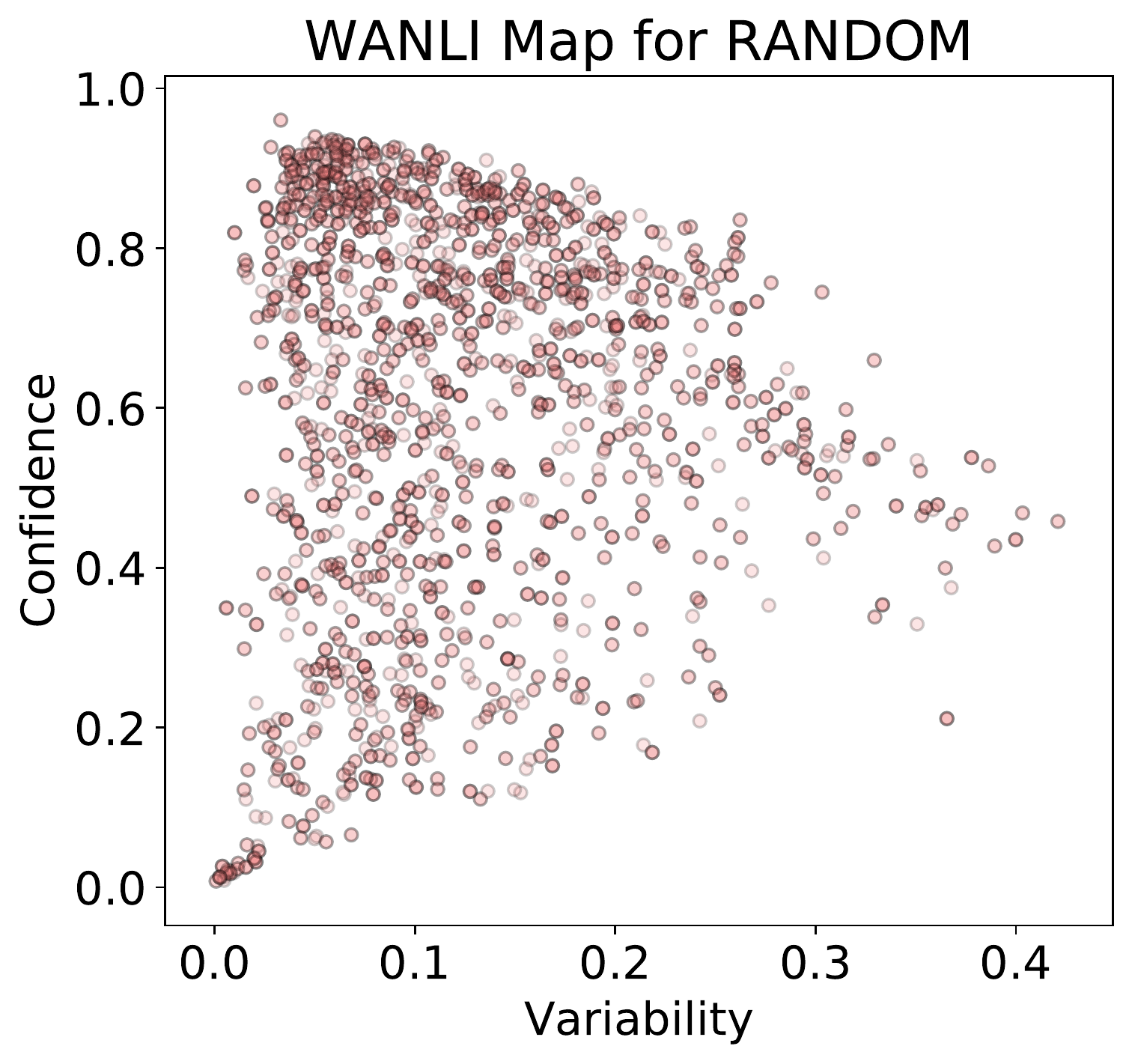}
\endminipage
\minipage{0.33\textwidth}
\includegraphics[width=\linewidth]{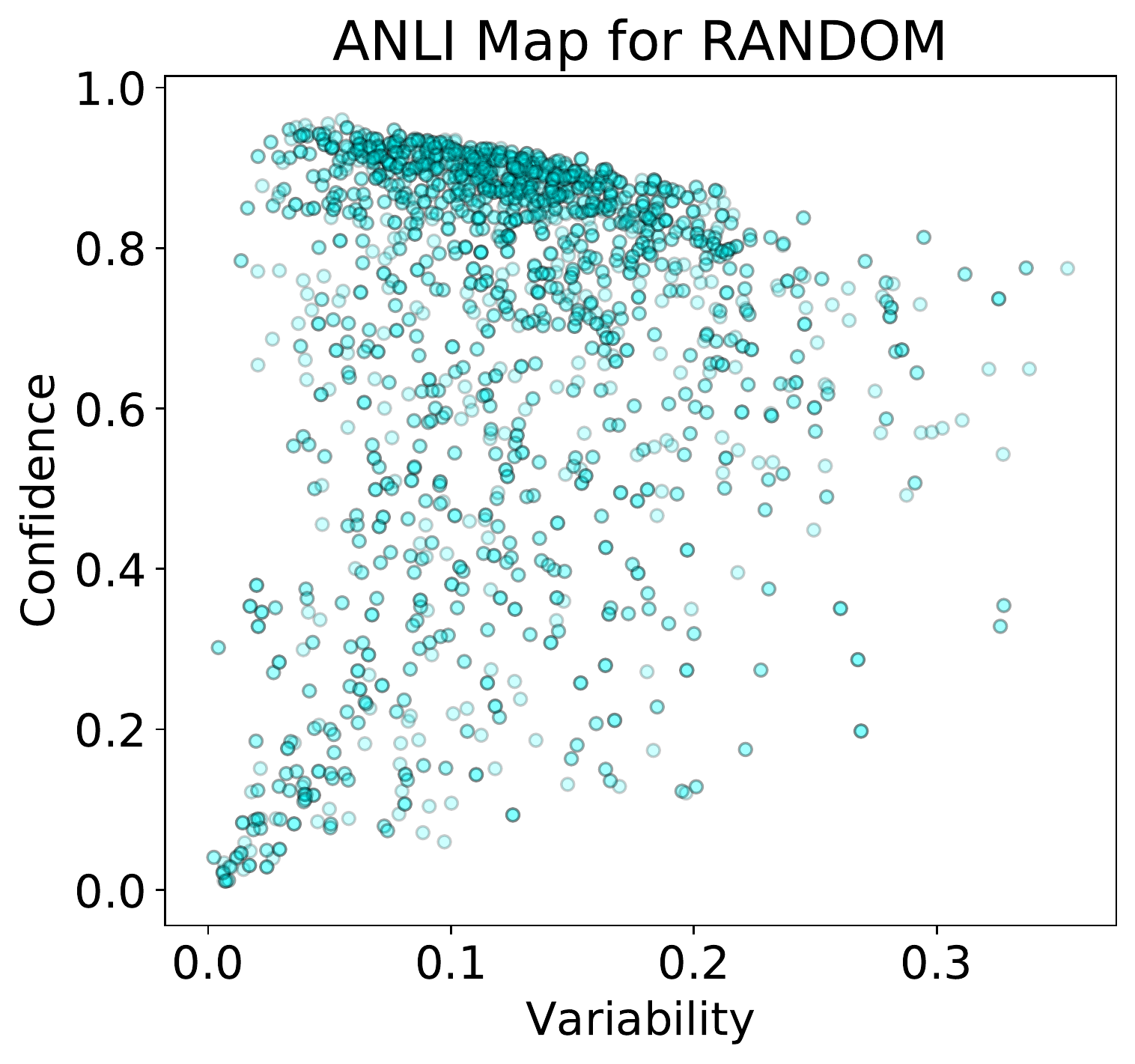}
\endminipage

\hfill

\minipage{0.33\textwidth}
\includegraphics[width=\linewidth]{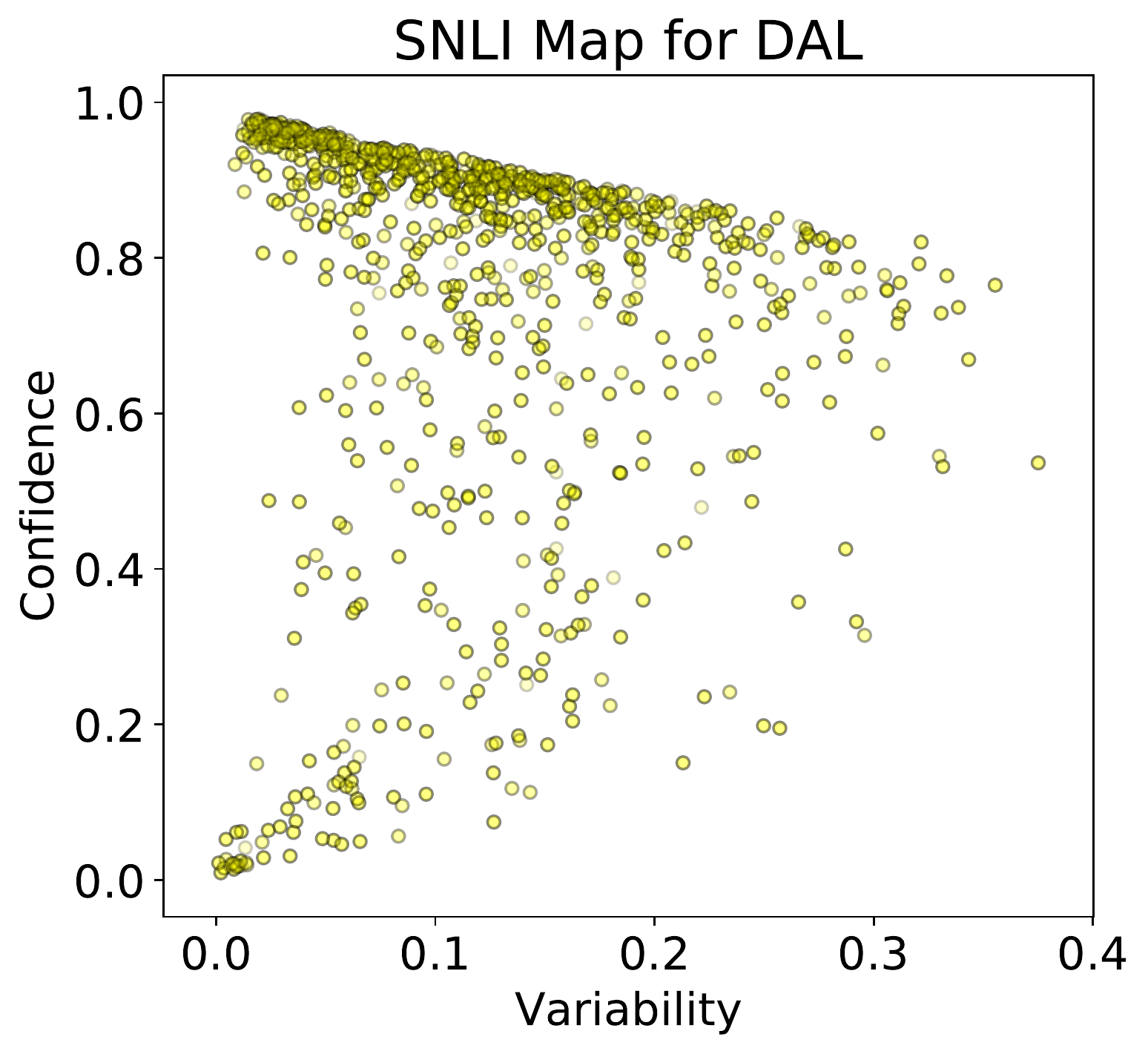}
\endminipage
\minipage{0.33\textwidth}
\includegraphics[width=\linewidth]{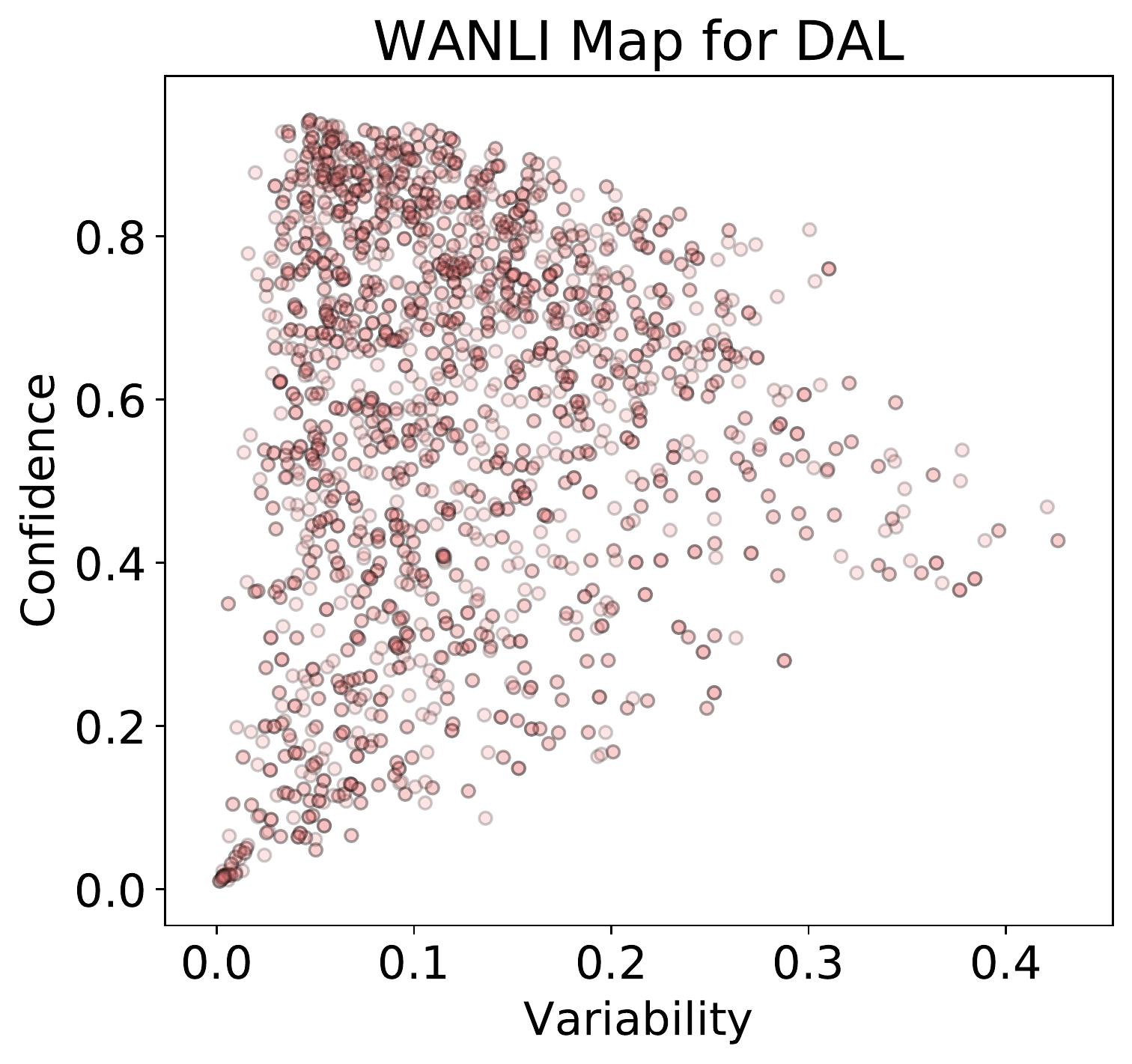}
\endminipage
\minipage{0.33\textwidth}
\includegraphics[width=\linewidth]{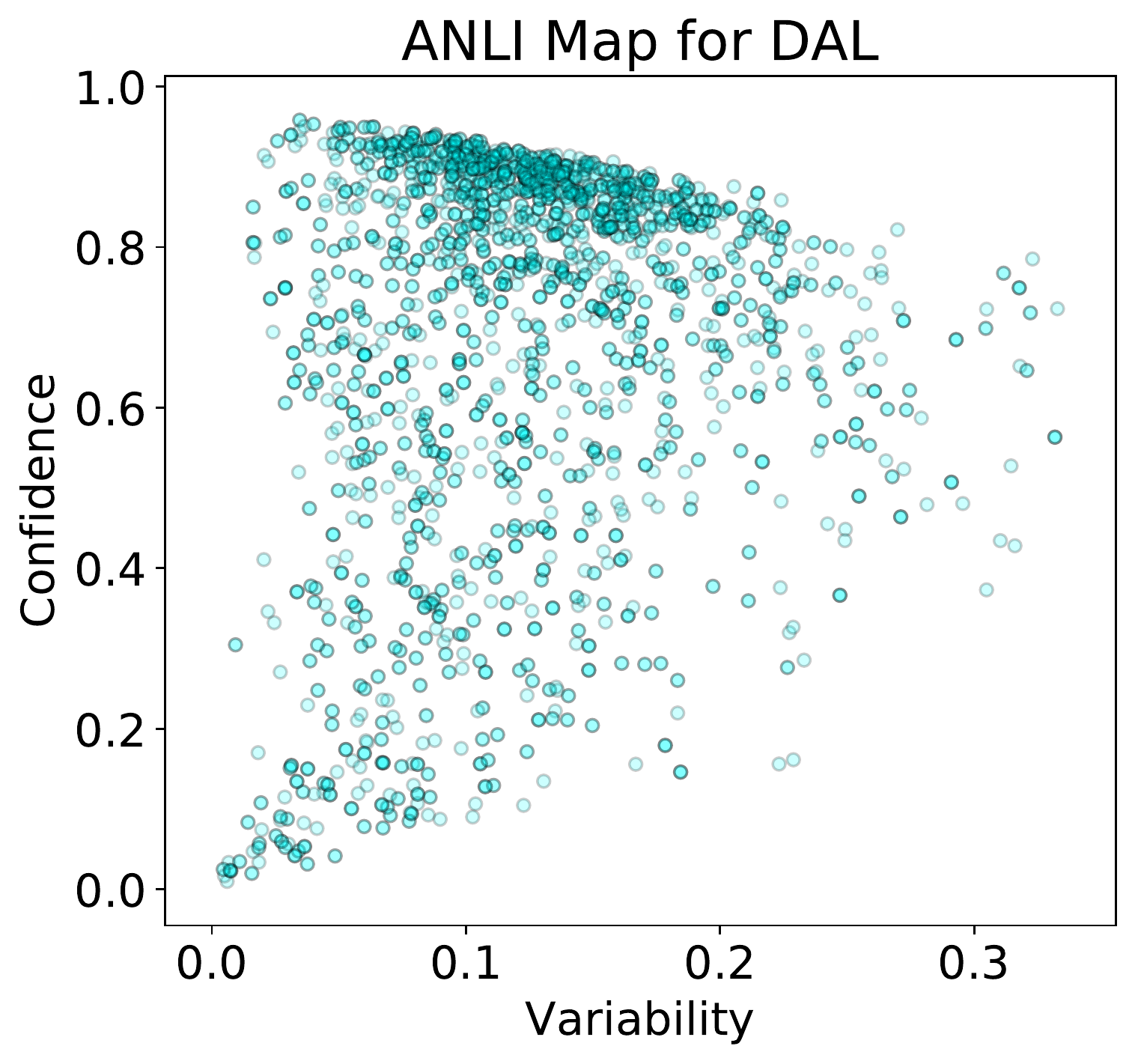}
\endminipage

\hfill

\minipage{0.33\textwidth}
\includegraphics[width=\linewidth]{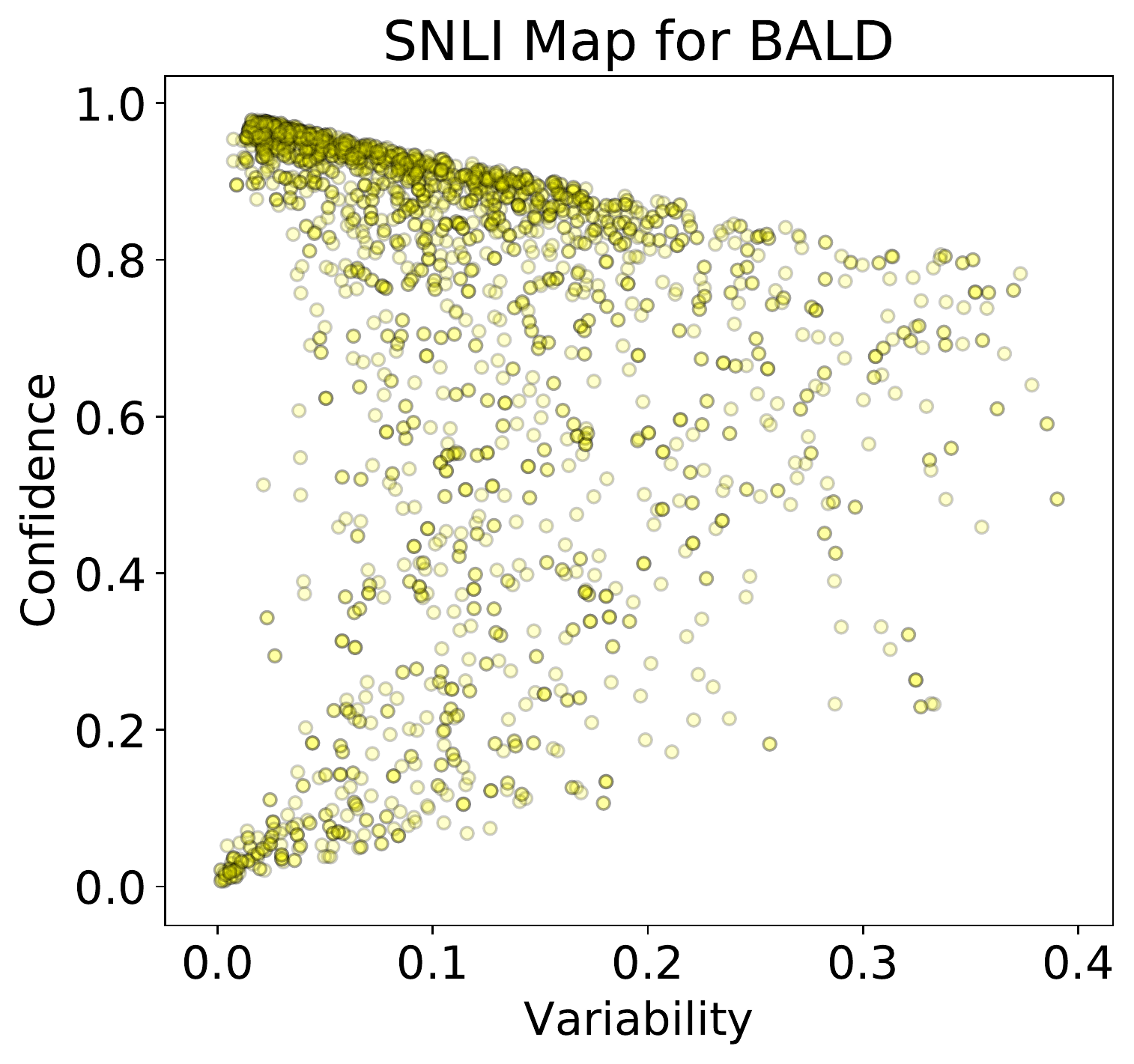}
\endminipage
\minipage{0.33\textwidth}
\includegraphics[width=\linewidth]{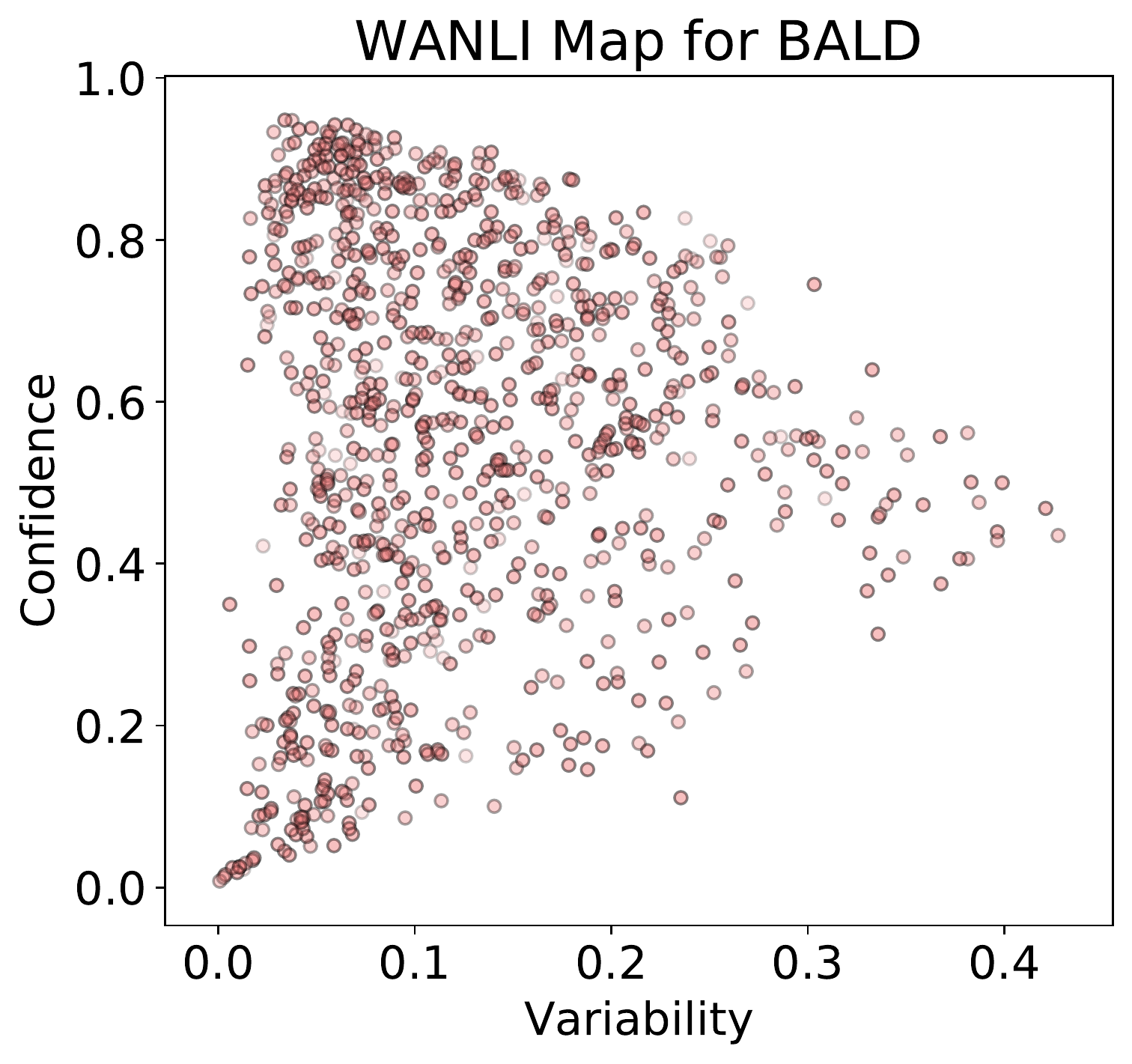}
\endminipage
\minipage{0.33\textwidth}
\includegraphics[width=\linewidth]{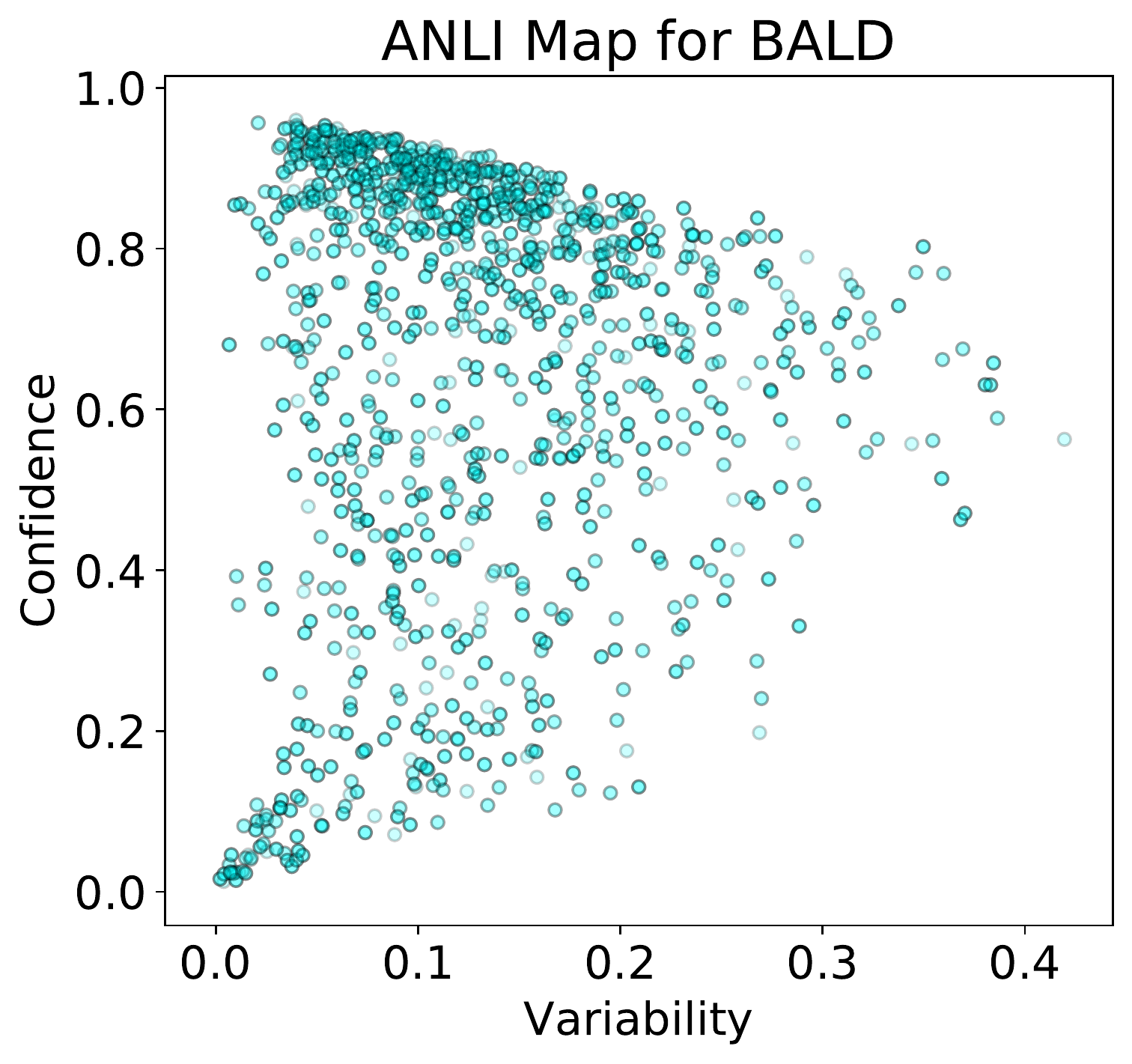}
\endminipage

\hfill

\minipage{0.33\textwidth}
\includegraphics[width=\linewidth]{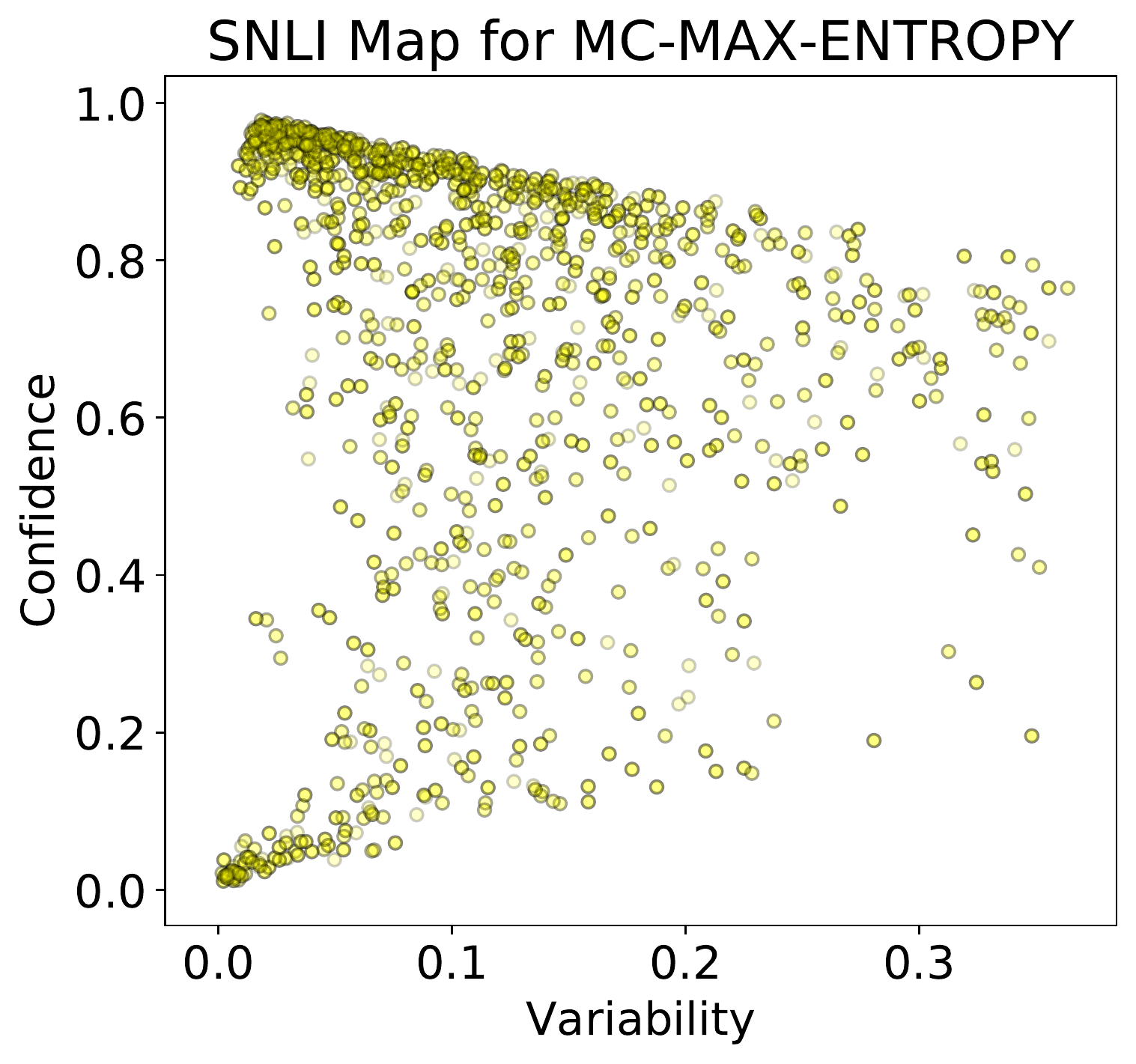}
\subcaption{Strategy maps for \snli{}}
\endminipage
\minipage{0.33\textwidth}
\includegraphics[width=\linewidth]{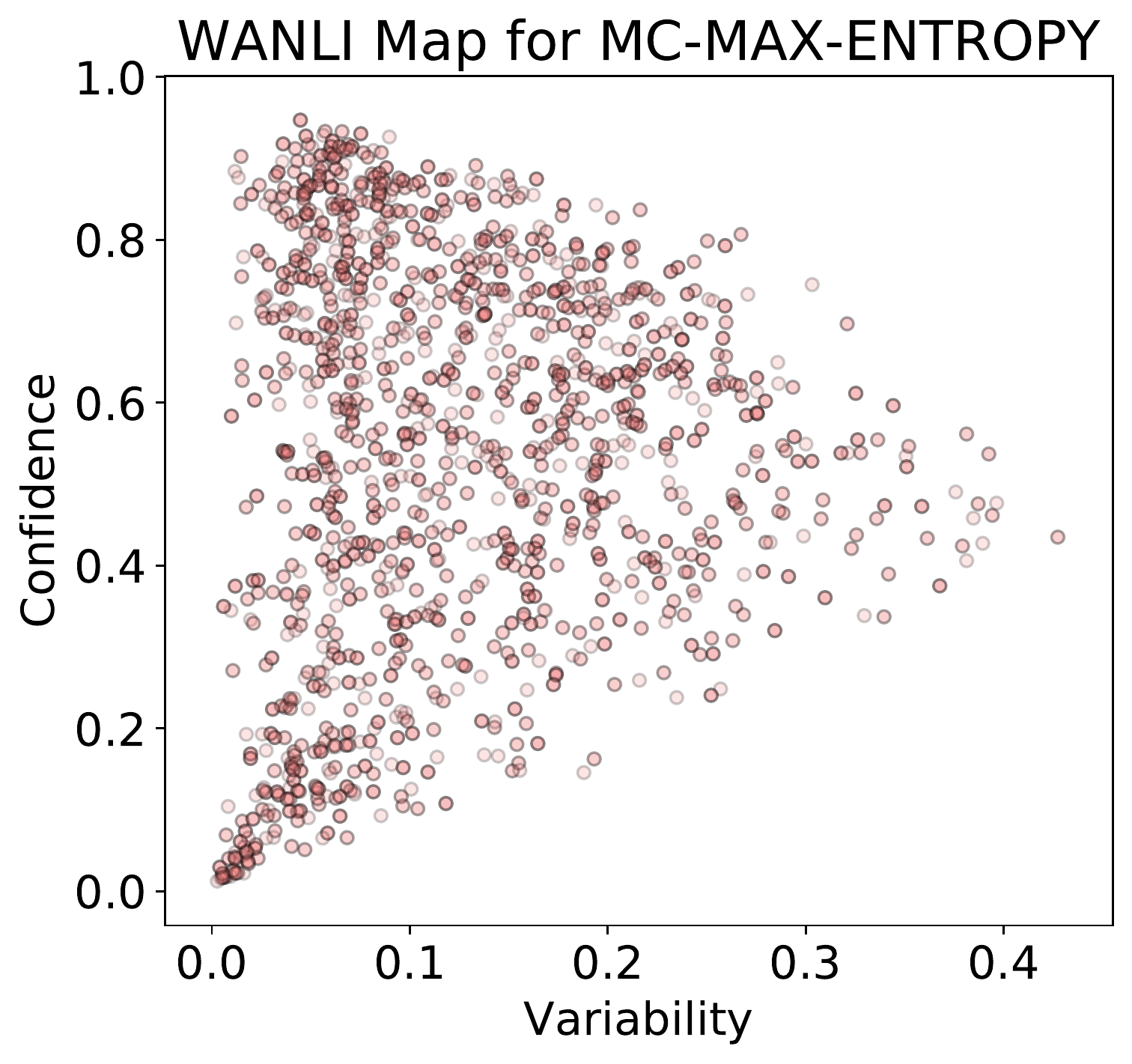}
\subcaption{Strategy maps for \wanli{}}
\endminipage
\minipage{0.33\textwidth}
\includegraphics[width=\linewidth]{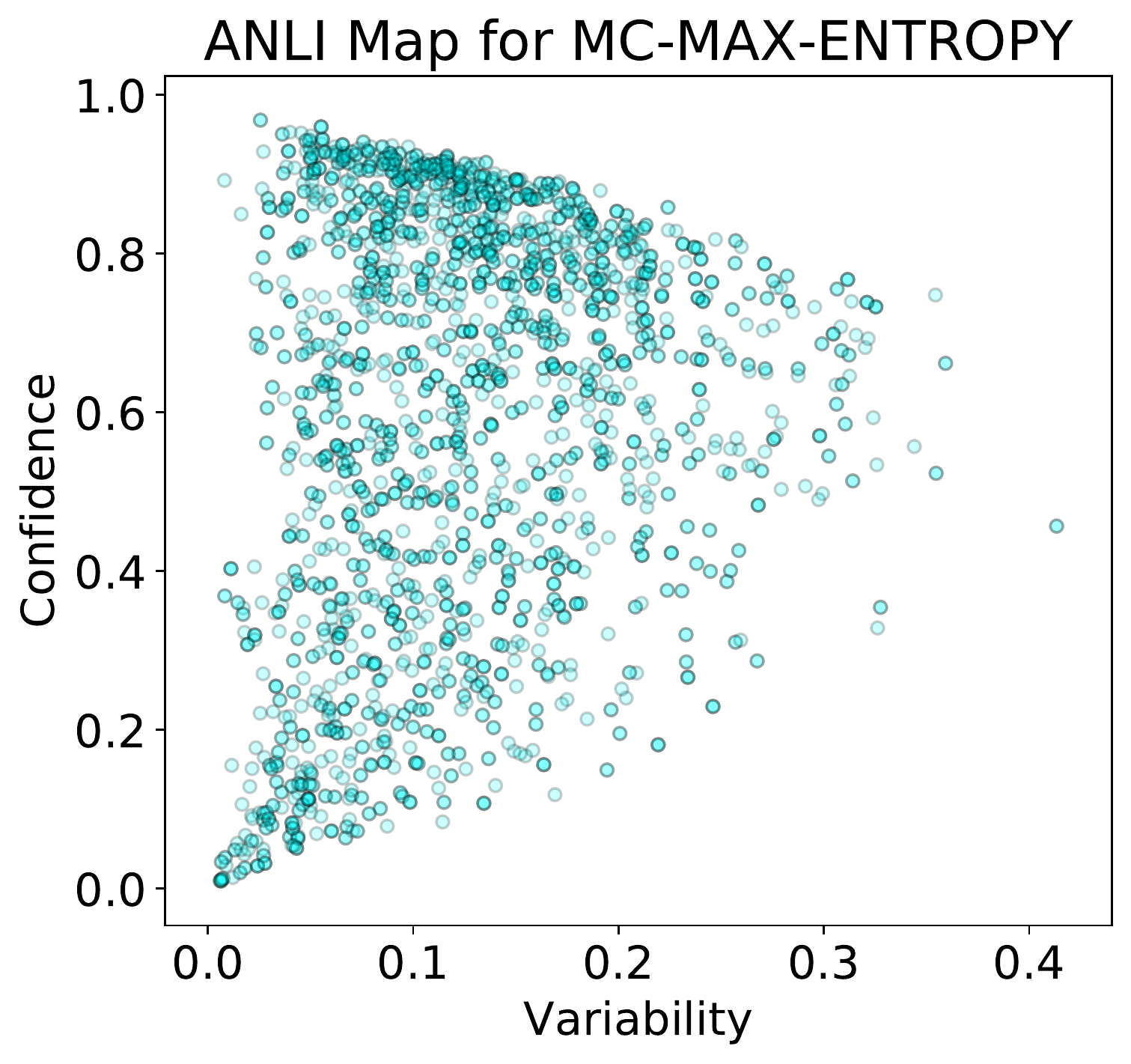}
\subcaption{Strategy maps for \anli{}}
\endminipage

\caption{Strategy maps for \textbf{multi-source} active learning, plotted per source.}
\label{fig:acquisitions_by_difficulty_per_source}
\end{figure*}

\newpage

\begin{figure*}\centering
\minipage{0.33\textwidth}
\includegraphics[width=\linewidth]{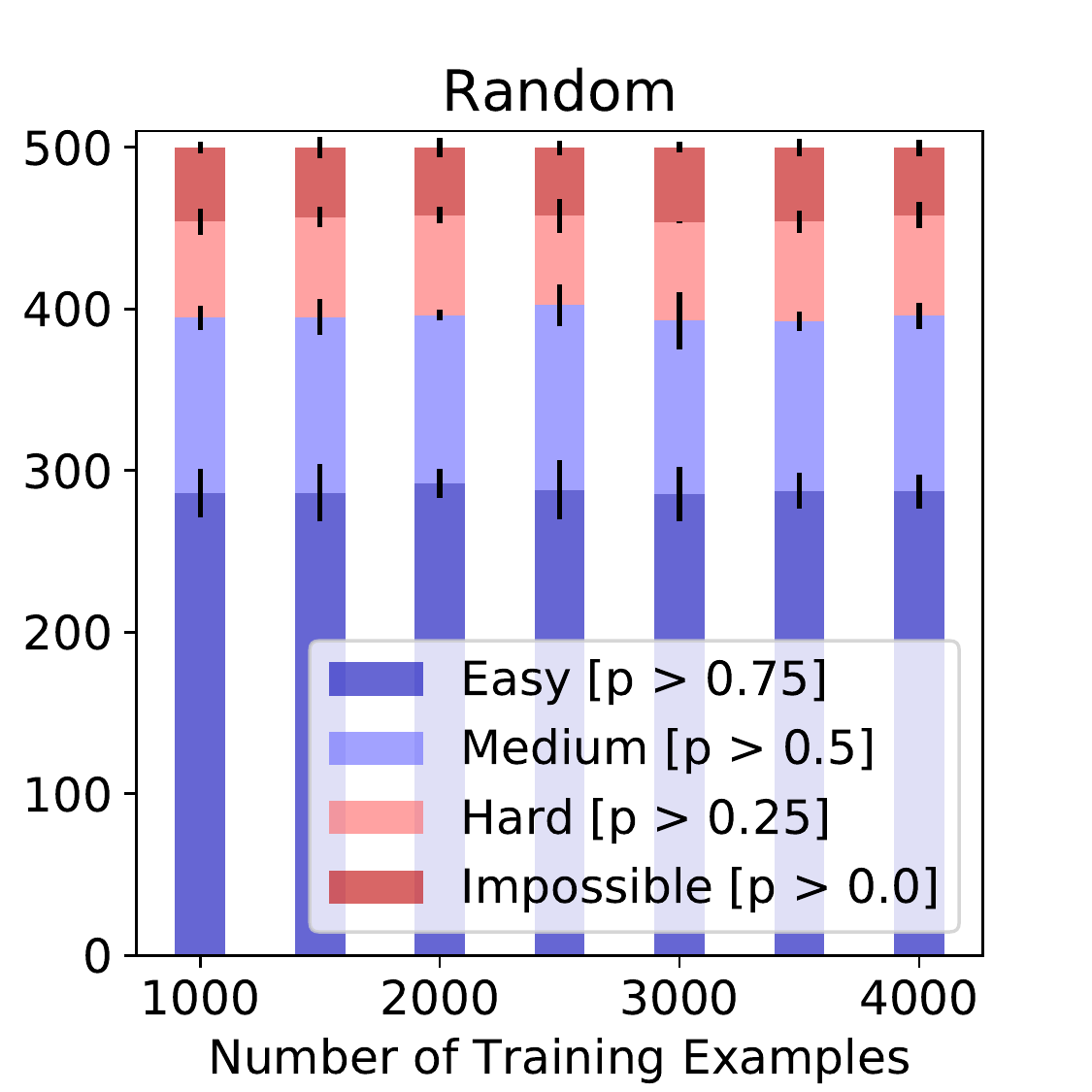}
\endminipage
\minipage{0.33\textwidth}
\includegraphics[width=\linewidth]{src/imgs/acquisition_difficulty/acquisitions_by_data_Random.pdf}
\endminipage
\minipage{0.33\textwidth}
\includegraphics[width=\linewidth]{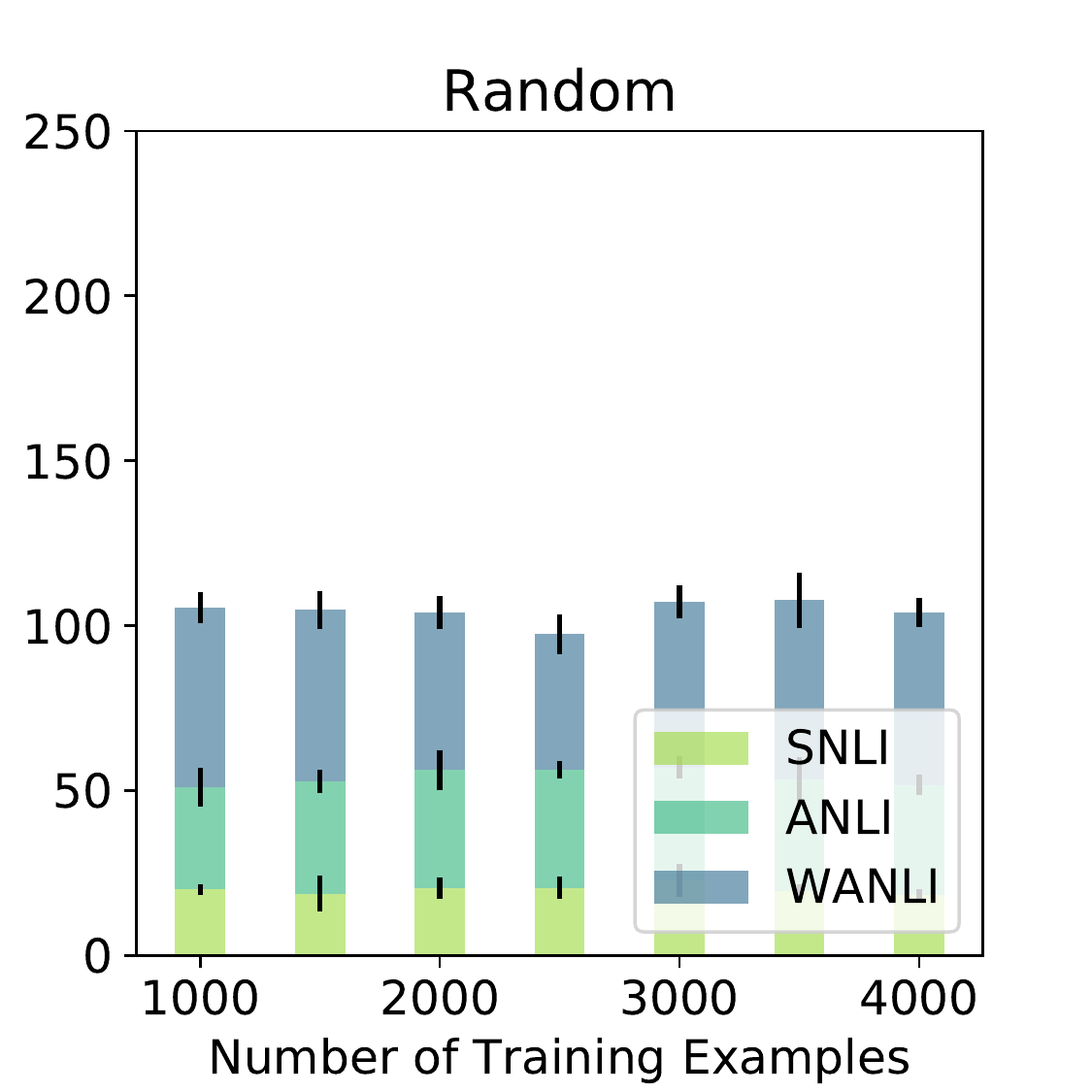}
\endminipage

\hfill

\minipage{0.33\textwidth}
\includegraphics[width=\linewidth]{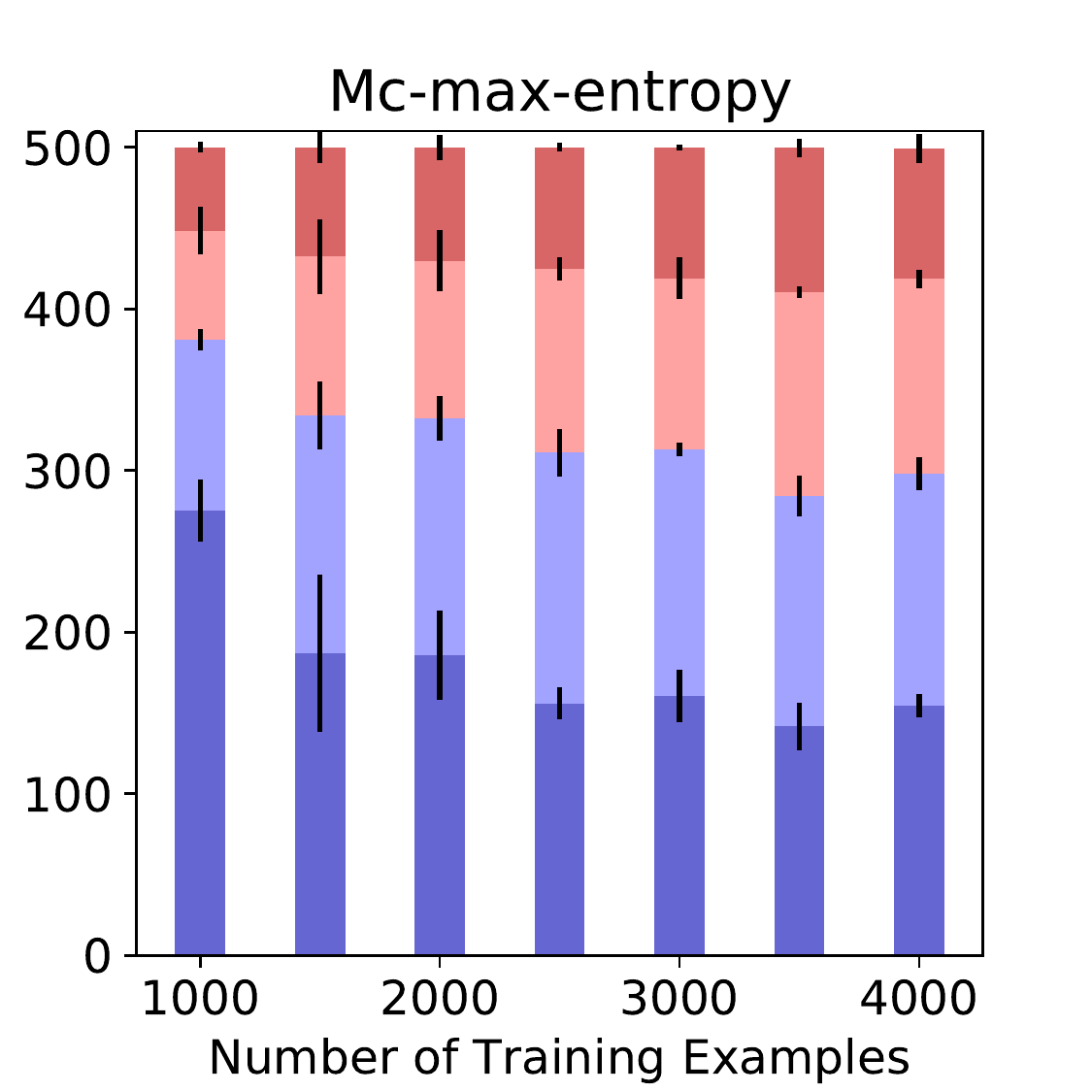}
\endminipage
\minipage{0.33\textwidth}
\includegraphics[width=\linewidth]{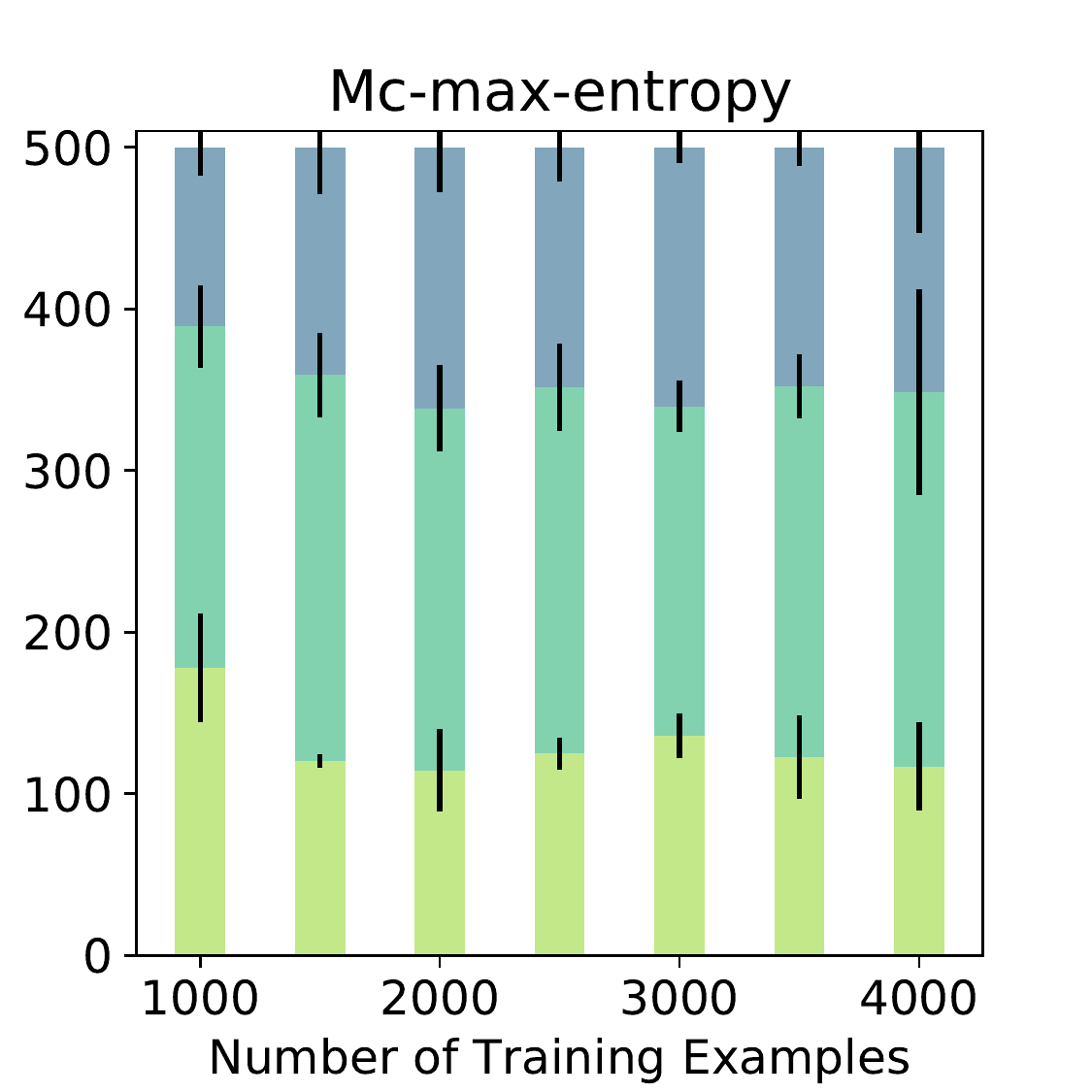}
\endminipage
\minipage{0.33\textwidth}
\includegraphics[width=\linewidth]{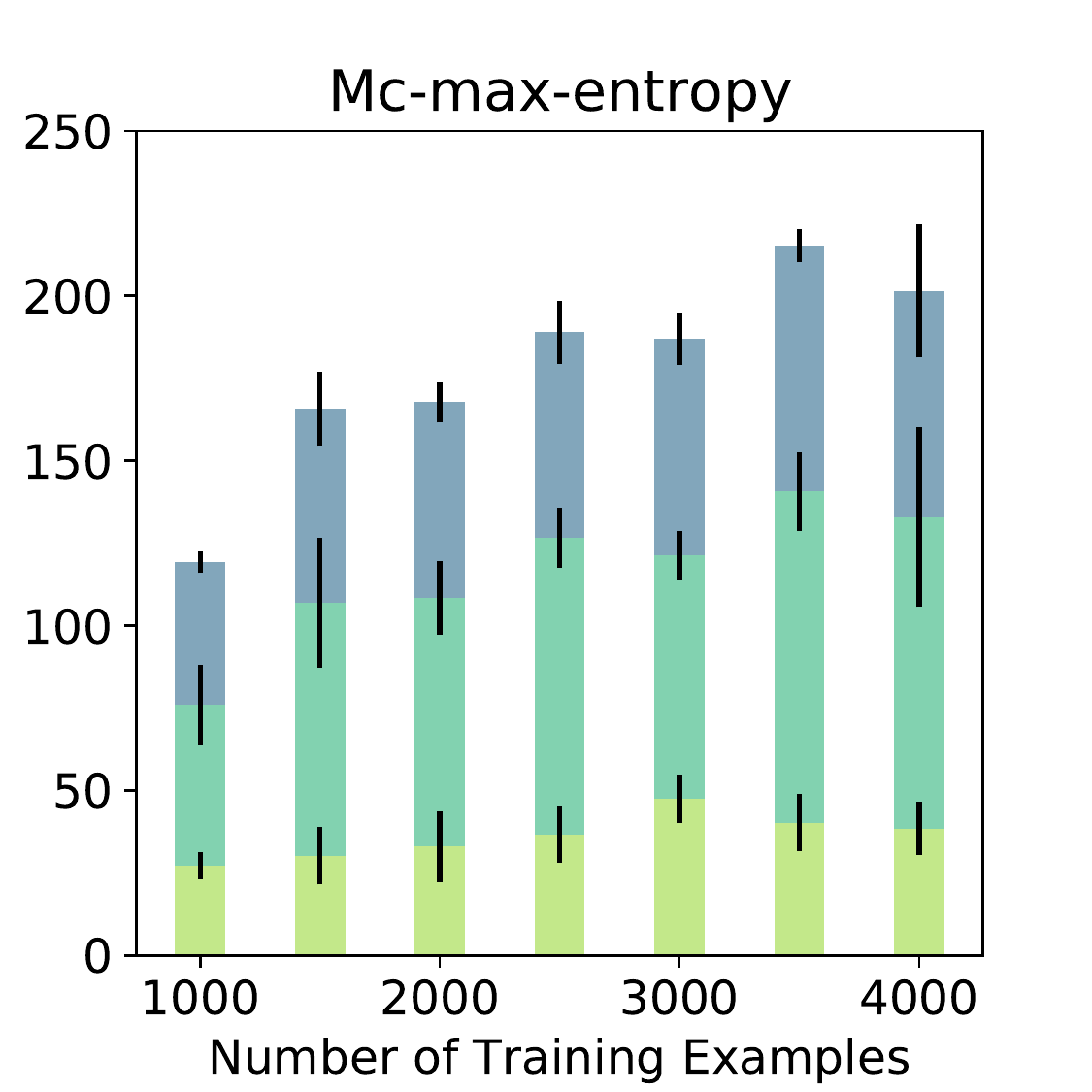}
\endminipage

\hfill

\minipage{0.33\textwidth}
\includegraphics[width=\linewidth]{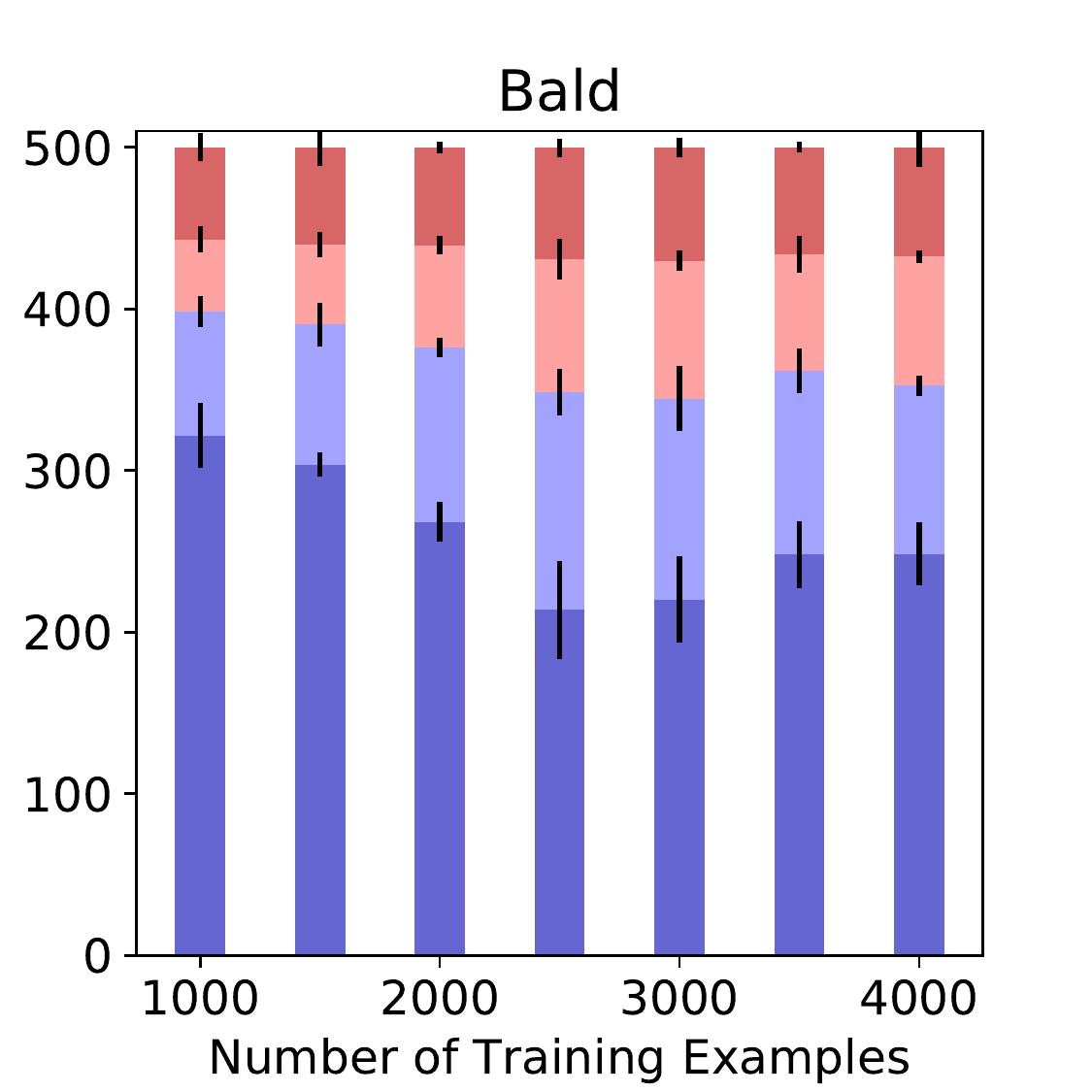}
\endminipage
\minipage{0.33\textwidth}
\includegraphics[width=\linewidth]{src/imgs/acquisition_difficulty/acquisitions_by_data_Bald.pdf}
\endminipage
\minipage{0.33\textwidth}
\includegraphics[width=\linewidth]{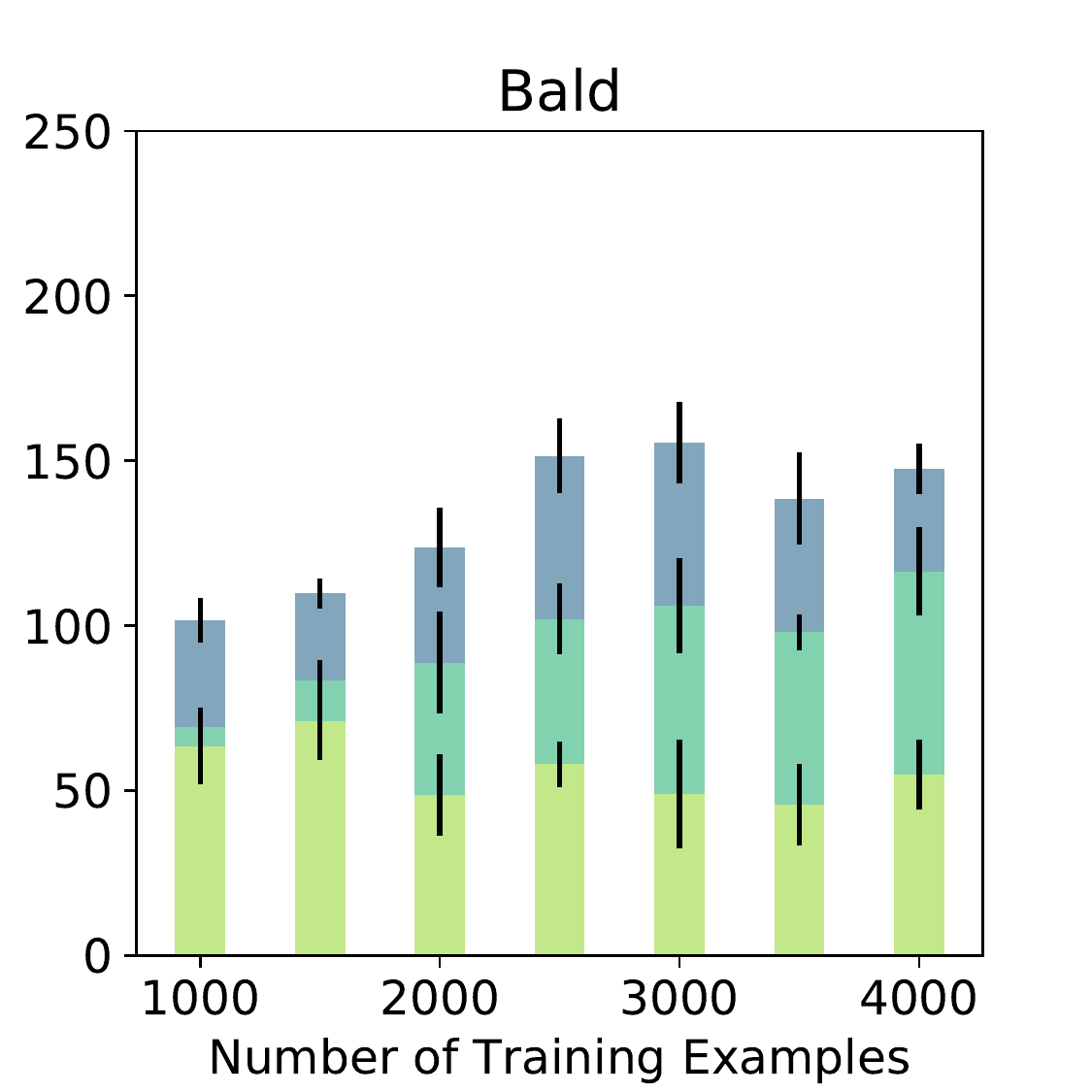}
\endminipage

\minipage{0.33\textwidth}
\includegraphics[width=\linewidth]{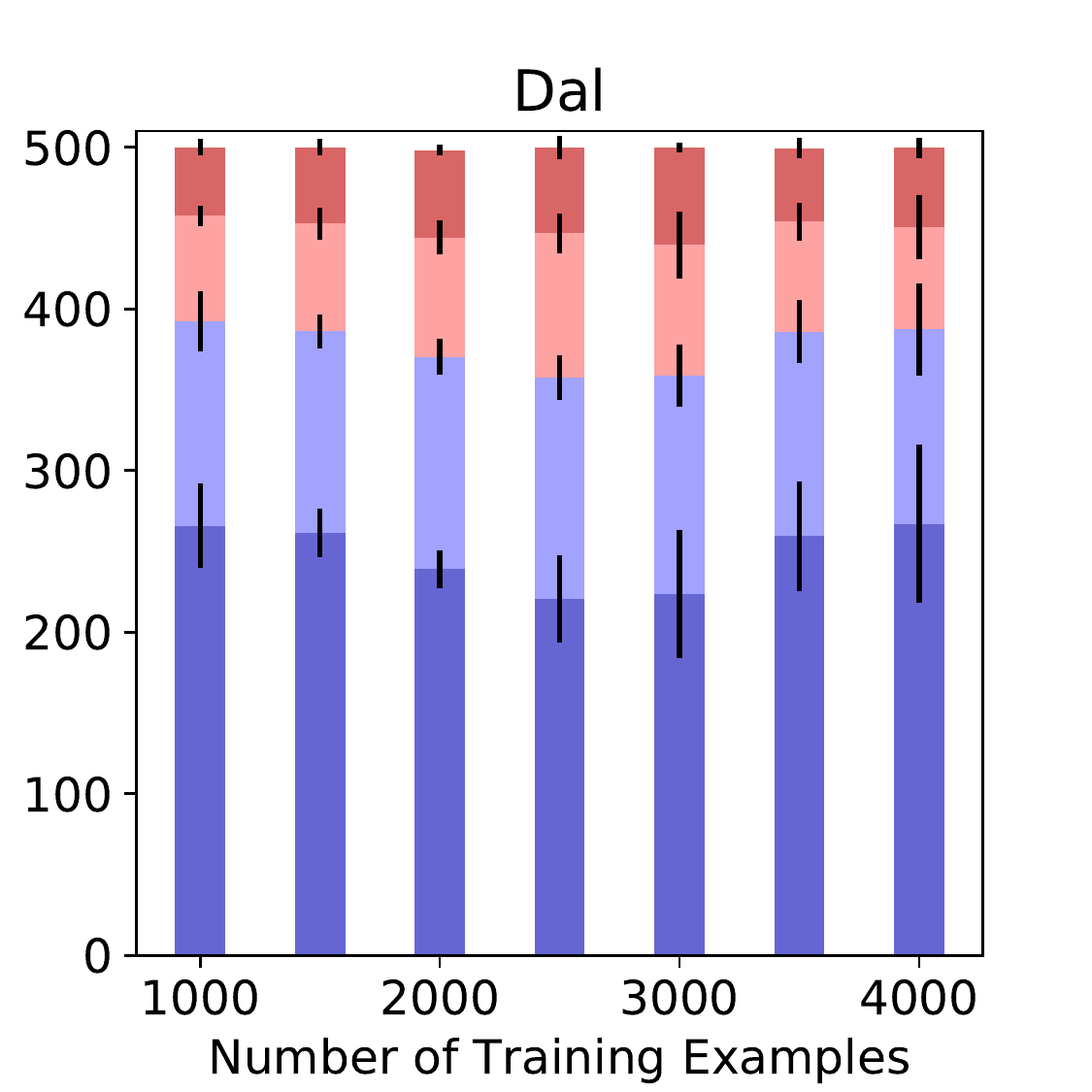}
\subcaption{Acquisition by learnability}
\endminipage
\minipage{0.33\textwidth}
\includegraphics[width=\linewidth]{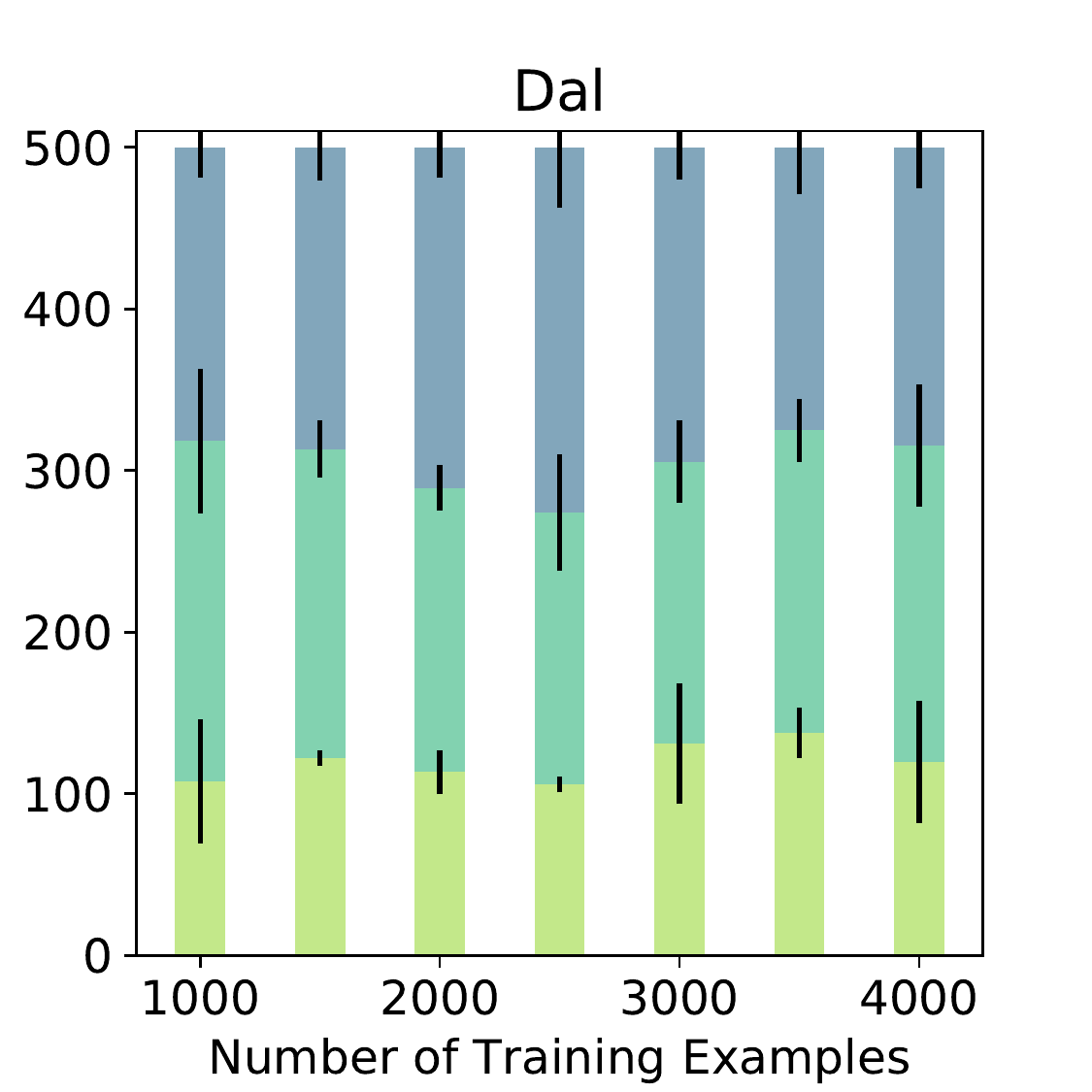}
\subcaption{Acquisition per source}
\endminipage
\minipage{0.33\textwidth}
\includegraphics[width=\linewidth]{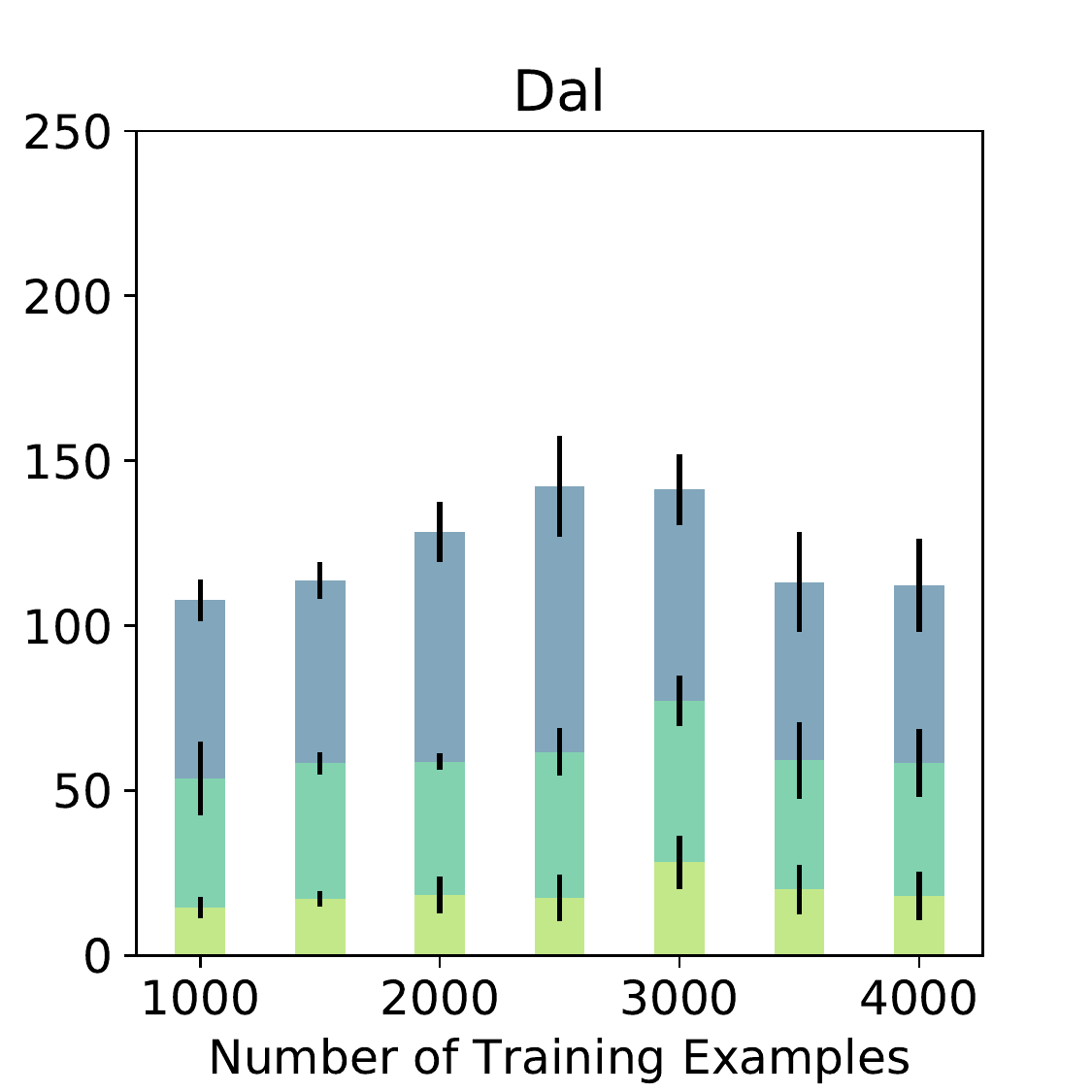}
\subcaption{Hard \& Imp. per source}
\endminipage

\caption{Profiling acquisitions for strategies over time. Y-axis denotes amount of acquired examples per round. The right graph plots the acquisition of hard and impossible examples. Dark bars indicate standard deviations across seeds. MCM-Entropy and BALD acquire notably more hard and impossible instances relative to random sampling.}
\label{fig:acquisitions_by_difficulty}
\end{figure*}


\newpage
\subsection{Impact of data difficulty on training time}
We find that models quickly learn simple examples while requiring more iterations for complex ones (See Figure \ref{fig:agg_difficulty_splits}). A similar outcome was observed by \citet{Lalor2018} who employ item response theory for data difficulty estimation - interestingly however, their difficulty parameters are estimated via a human population, while our difficulty estimations follow directly from training dynamics. This echoes a similar finding by \citep{Zhang}, who show that cartography-based model confidence scores strongly correlate with human agreement on SNLI validation data. Similarly, it may be of interest to see whether examples which \textit{many different} models identify as ambiguous correlate with human judgements of ambiguity, e.g. with respect to human label distributions \citep{Nie2020, Chen2020}, or annotator disagreement scores.

\subsection{Hard-to-learn Data}\label{section:impossible_data}
In the tables below, we present various cases of hard-to-learn examples (i.e., collective outliers) for the NLI datasets we examine.
\begin{table*}[h!]\caption{Impossible ($0 \leq p \leq 0.25$) training examples from \snli{}. \textbf{GT} denotes the gold truth; \textbf{P.} denotes the model prediction and \textbf{Conf.} the confidence associated with the predicted label. }
\label{imp_snli}
\small
\begin{tabularx}{\linewidth}{
    >{\hsize=0.1\hsize}X
    >{\hsize=1.9\hsize}X
    >{\hsize=1.88\hsize}X
    >{\hsize=0.06\hsize}X
    >{\hsize=0.06\hsize}X
    >{\hsize=0.12\hsize}X
  }
\toprule
\textbf{Ex.} & \textbf{Premise} & \textbf{Hypothesis} & \textbf{GT}  & \textbf{P.} & \textbf{Conf.} \\
\midrule
(1)
&
\texttt{A skier in electric green on the edge of a ramp made of metal bars.}	
&
\texttt{The skier was on the edge of the ramp.}	
&
N
&
E
&
0.976
\\
\addlinespace
($2$)
&
\texttt{A skier in electric green on the edge of a ramp made of metal bars.	}
&
\texttt{The brightly dressed skier slid down the race course.	}
&
E
&
C
&
0.770
\\
\addlinespace
($3$)
&
\texttt{A man wearing a red sweater is sitting on a car bumper watching another person work.}
&
\texttt{people make speed fast at speed breaker.	}
&
C	
&
N
&
0.799
\\
\addlinespace
(4)
& 
\texttt{Middle-aged female wearing a white sunhat and white jacket, slips her hand inside a man's pants pocket.}	
&
\texttt{The man and woman are playing together.}
&
C	
&
N
&
0.869
\\
\addlinespace
(5)
&
\texttt{A young girl jumps off of a couch and high into the air}
&
\texttt{the young lady knows how to fly in sky}	
&
C	
&
N
&
0.765
\\
\addlinespace
(6)
&
\texttt{A young boy jumps into the oncoming wave.}
&
\texttt{The boy is at a lake.}	
&
C	
&
N
&
0.951
\\
\addlinespace
(7)
&
\texttt{A woman taking her wallet out of her purse at a vendor stand}
&
\texttt{A woman buying something from a vendor.}	
&
E	
&
N
&
0.892
\\
\addlinespace
(8)
&
\texttt{A group of people standing on a rock path.}
&
\texttt{A group of people are hiking.}	
&
E	
&
N
&
0.972
\\
\addlinespace
(9)
&
\texttt{Woman sitting in tree with dove.}
&
\texttt{The lady is touching a dove.}	
&
N	
&
E
&
0.920
\\
\addlinespace
(10)
&
\texttt{A lady standing on the corner using her phone.}
&
\texttt{The lady has a smartphone.}	
&
N	
&
E
&
0.753
\\
\bottomrule
\end{tabularx} 
\end{table*}

\begin{table*}[h!]\caption{Impossible ($0 \leq p \leq 0.25$) training examples from \wanli{}. \textbf{GT} denotes the gold truth; \textbf{P.} denotes the model prediction and \textbf{Conf.} the confidence associated with the predicted label. 
}
\label{tab2}
\small
\begin{tabularx}{\linewidth}{
    >{\hsize=0.10\hsize}X
    >{\hsize=2.1\hsize}X
    >{\hsize=1.5\hsize}X
    >{\hsize=0.06\hsize}X
    >{\hsize=0.06\hsize}X
    >{\hsize=0.12\hsize}X
  }
\toprule
\textbf{Ex.} & \textbf{Premise} & \textbf{Hypothesis} & \textbf{GT}  & \textbf{P.} & \textbf{Conf.} \\
\midrule
(1)
&
\texttt{The first principle of art is that art is not a way of life, but a means of life.}  
& 
\texttt{Art is a way of life.}
& 
E
& 
C 
&
0.944
\\
\addlinespace
(2)
&
\texttt{"Must be right good stock," Fenner observed.}
&
\texttt{"Must be pretty good stock," Fenner said.}
&
C
&
E
&
0.646
\\
\addlinespace
(3)
&
\texttt{He believes that the best thing to do is to buy the firm at a reasonable price.}
&
\texttt{If the firm is cheap, it is best to buy it.}
&
N
&
E
&
0.964
\\
\addlinespace
(4)
&
\texttt{A piece of paper with a stamp on it is worth less than a piece of paper without a stamp.}
&
\texttt{A piece of paper without a stamp is worth more than a piece of paper with a stamp.}
&
E
&
C
&
0.818
\\
\addlinespace
(5)
&
\texttt{It was the only time he had seen her laugh.}
&
\texttt{He had never seen her laugh before.}
&
C
&
N
&
0.502
\\
\addlinespace
(6)
&
\texttt{The musician shook his head.}
&
\texttt{The musician moved his head up and down.}
&
C
&
E
&
0.510
\\
\addlinespace
(7)
&
\texttt{Some students believe that to achieve their goals they must take the lead.}
&
\texttt{Some students believe that to achieve their goals they must follow the lead.}
&
E
&
C
&
0.630
\\
\addlinespace
(8)
&
\texttt{The very nature of the "American Dream" is that it is not always attainable}
&
\texttt{The American Dream is attainable.}
&
N
&
C
&
0.817
\\
\addlinespace
(9)
&
\texttt{This might be an issue for a company that is in the process of introducing a new product.}
&
\texttt{Every company is always in the process of introducing a new product.}
&
E
&
N
&
0.894
\\
\addlinespace
(10)
&
\texttt{Would Higher Interest Rates Stimulate Saving?}
&
\texttt{Higher interest rates would not stimulate saving.}
&
E
&
N
&
0.929
\\
\bottomrule
\end{tabularx} 
\label{imp_wanli}
\end{table*}

\begin{table*}[h!]\caption{Impossible ($0 \leq p \leq 0.25$) training examples from \anli{}. \textbf{GT} denotes the gold truth; \textbf{P.} denotes the model prediction and \textbf{Conf.} the confidence associated with the predicted label. Instances are comparatively less characterized by noisy labels or ambiguity; rather, their difficulty arises through requirement of more advanced inference types, e.g.\ identifying relevant information from long passages, multi-hop reasoning to connect subjects to events and numerical reasoning about dates of birth and events.}\label{tab3}
\small
\begin{tabularx}{\linewidth}{
    >{\hsize=0.10\hsize}X
    >{\hsize=2.9\hsize}X
    >{\hsize=0.8\hsize}X
    >{\hsize=0.06\hsize}X
    >{\hsize=0.06\hsize}X
    >{\hsize=0.12\hsize}X
  }
\toprule
\textbf{Ex.} & \textbf{Premise} & \textbf{Hypothesis} & \textbf{GT}  & \textbf{P.} & \textbf{Conf.} \\
\midrule
(1)
&
\texttt{Glaiza Herradura-Agullo (born February 24, 1978) is a Filipino former child actress. She was the first-ever grand winner of the Little Miss Philippines segment of "Eat Bulaga!" in 1984. She starred in RPN-9's television series "Heredero" with Manilyn Reynes and Richard Arellano. She won the 1988 FAMAS Best Child Actress award for her role in "Batas Sa Aking Kamay" starring Fernando Poe, Jr. .}	
&
\texttt{Herradura-Agullo was born in the 80's}
&
C	
&
E
&
0.941
\\
\addlinespace
(2)
&
\texttt{The Whitechapel murders were committed in or near the impoverished Whitechapel district in the East End of London between 3 April 1888 and 13 February 1891. At various points some or all of these eleven unsolved murders of women have been ascribed to the notorious unidentified serial killer known as Jack the Ripper.}	
&
\texttt{The women killed in the Whitechapel murders were impoverished.}
&
N	
&
E
&
0.832
\\
\addlinespace 
(3)
&
\texttt{Departure of a Grand Old Man is a 1912 Russian silent film about the last days of author Leo Tolstoy. The film was directed by Yakov Protazanov and Elizaveta Thiman, and was actress Olga Petrova's first film.}
&
\texttt{Olga performed in many films before This one}	
&
C	
&
N
&
0.922
\\ 
\addlinespace
(4)
&
\texttt{Gay Sex in the 70s is a 2005 American documentary film about gay sexual culture in New York City in the 1970s. The film was directed by Joseph Lovett and encompasses the twelve years of sexual freedom bookended by the Stonewall riots of 1969 and the recognition of AIDS in 1981, and features interviews with Larry Kramer, Tom Bianchi, Barton Lidice Beneš, Rodger McFarlane, and many others.}	
&
\texttt{Gay Sex in the 70s was directed by a gay man.}	
&
N	
&
E
&
0.907
\\
\addlinespace
(5)
&
\texttt{Héctor Canziani was an Argentine poet, screenwriter and film director who worked in Argentine cinema in the 1940s and 1950s. Although his work was most abundant in screenwriting and poetry after his brief film career, he is best known for his directorship and production of the 1950 tango dancing film Al Compás de tu Mentira based on a play by Oscar Wilde.}	&
\texttt{He did direct a movie after 1950}	
&
N
&
E
&
0.944
\\
\bottomrule
\end{tabularx} 
\label{imp_anli}
\end{table*}



\end{document}